\documentclass[3p,11pt]{elsarticle}

\journal{Information Sciences}

\usepackage{amsmath,amssymb,amsthm,mathtools}
\usepackage{stmaryrd}
\usepackage{latexsym}
\usepackage{MnSymbol} 
\usepackage{bbm}
\usepackage{subfig}
\usepackage{multirow}
\usepackage{rotating}
\usepackage{url}
\usepackage{color}
\usepackage{lineno}
\usepackage{hyperref}

\usepackage{ulem}
\newcommand{\new}{}

\usepackage{algorithmic}
\usepackage{algorithm}

\newcommand{\reels}{\mathbb{R}}

\newcommand{\calG}{{\cal G}}
\newcommand{\calL}{{\cal L}}

\newcommand{\calM}{{\cal M}}

\newcommand{\calO}{{\cal O}}

\newcommand{\calP}{{\cal P}}
\newcommand{\calF}{{\cal F}}

\newcommand{\calX}{{\cal X}}

\newcommand{\calI}{{\cal I}}

\def\mh{\widehat{m}}

\def\bthetah{{\widehat{\boldsymbol{\theta}}}}

\newcommand{\deriv}[2]{\frac{\partial #1}{\partial #2}}

\def\bdelta{\boldsymbol{\delta}}
\def\btheta{{\boldsymbol{\theta}}}

\def\bx{{\boldsymbol{x}}}

\def\bm{{\boldsymbol{m}}}

\def\bD{{\boldsymbol{D}}}
\def\bC{{\boldsymbol{C}}}

\def\bS{{\boldsymbol{S}}}

\def\bQ{{\boldsymbol{Q}}}
\def\bS{{\boldsymbol{S}}}
\def\bE{{\boldsymbol{E}}}

\def\bone{{\boldsymbol{1}}}

\newcommand{\bi}{\begin{itemize}}
\newcommand{\ei}{\end{itemize}}
\newcommand{\be}{\begin{enumerate}}
\newcommand{\ee}{\end{enumerate}}
\newcommand{\bd}{\begin{description}}
\newcommand{\ed}{\end{description}}

\newtheorem{Ex}{Example}

\begin{document}

\begin{frontmatter}

\title{NN-EVCLUS: Neural Network-based Evidential Clustering}

\author[utc,iuf,shu]{Thierry Den{\oe}ux}
\ead{Thierry.Denoeux@utc.fr}
\address[utc]{Universit\'e de technologie de Compi\`egne, CNRS, Heudiasyc, Compi\`egne, France}
\address[iuf]{Institut universitaire de France, Paris, France}
\address[shu]{Shanghai University, UTSEUS, Shanghai, China}

\begin{abstract}
Evidential clustering is an approach to clustering based on the use of Dempster-Shafer mass functions to represent cluster-membership uncertainty.  In this paper, we introduce a neural-network based evidential clustering algorithm, called NN-EVCLUS, which learns a mapping from  attribute vectors to mass functions, in such a way that more similar inputs are mapped to output mass functions with a lower degree of conflict. The neural network can be paired with a one-class support vector machine to make it robust to outliers and capable of \new{detecting previously unseen clusters when applied to new data}. The network is trained to minimize the discrepancy between dissimilarities and degrees of conflict for all or some object pairs. Additional terms can be added to the loss function to account for pairwise constraints or labeled data, which can also be used to adapt the metric. Comparative experiments show the superiority of NN-EVCLUS over state-of-the-art evidential clustering algorithms for a range of unsupervised and constrained clustering tasks involving both attribute and dissimilarity data. 
\end{abstract}


\end{frontmatter}



\section{Introduction}
\label{sec:intro}

One of the most important tasks in machine learning and exploratory data analysis is finding groups of similar objects in a dataset, in such a way that the dissimilarity between groups is maximized. This problem, referred to as \emph{clustering}, has been addressed using a wide range of techniques and from a variety of perspectives (see, e.g. \cite{jain88,kaufman90,xu09}). While the earlier methods such as the hard $c$-means algorithm do not consider group-membership uncertainty, quantifying this uncertainty has been a major issue in the last 40 years \cite{peters13,durso17}. Among the most widely-used formalisms, we can mention fuzzy sets \cite{bezdek99,durso19}, possibility theory \cite{krishnapuram93,pal05}, and rough sets \cite{peters14,ferone19,ubukata21}. These approaches are, to some extent, subsumed by a  relatively new approach, referred to as \emph{evidential clustering}, which is based on the Dempster-Shafer (DS) theory of belief functions \cite{denoeux04b,masson08,denoeux16b}. Evidential clustering algorithms quantify clustering uncertainty using DS mass functions assigning masses to  sets of clusters, called \emph{focal sets}, in such a way that the masses sum to one. The collection of mass functions  related to the $n$ objects is called an \emph{evidential} (or \emph{credal}) \emph{partition}. An evidential partition boils down to a fuzzy partition when the focal sets are singletons, and it is equivalent to a rough partition when each mass function has a single focal set \cite{denoeux16b}.

Among  the earliest evidential clustering procedures are the evidential $c$-means (ECM) algorithm \cite{masson08} and its relational version \cite{masson09a}. The ECM algorithm maximizes an objective function based on distances to prototypes associated not only to individual clusters but also to sets of clusters or``meta-clusters''. The prototype representing a meta-cluster is the barycenter of the prototypes representing the clusters it contains. The method was shown to provide meaningful evidential partitions, but it can provide undesirable results when the prototype of a meta-cluster is close to that of an individual cluster. The Belief $c$-means \cite{liu12} and Credal $c$-means (CCM) \cite{liu15} algorithms are alternative procedures designed to address this problem. The Belief Peak Evidential Clustering (BPEC) method \cite{su19} combines ideas from density peak clustering \cite{rodriguez14,xu20,he21} and ECM. The Median Evidential $c$-means (MECM)  \cite{zhou15} is an evidential version of the median $c$-means for relational data, while the Evidential $c$-medoid (ECMdd) with either a single prototype per class or multiple weighted prototypes \cite{zhou16} are inspired by the $c$-medoids algorithm. 

In this paper, we revisit an earlier approach to evidential clustering that is not based on prototypes, but that takes inspiration from multidimensional scaling (MDS) \cite{borg97}. The EVCLUS algorithm \cite{denoeux04b} constructs an evidential partition in such a way that the degree of conflict between mass functions related to any two objects match the similarity between these two objects. The method has been shown to outperform most other  algorithms for handling nonmetric dissimilarity data  \cite{denoeux04b}, but it can obviously also be applied to attribute data after a distance matrix has been computed.  Whereas the initial algorithm was initially only applicable to small datasets of a few hundred objects, algorithmic improvements   introduced in \cite{denoeux16a} have made it applicable to much larger datasets containing tens of thousands of objects.

Whereas the EVCLUS has good performances in clustering tasks, it has some limitations. First, as it directly constructs mass functions describing the cluster membership of each object as the solution of an optimization problem, it does not allow us to classify new objects, other than by including these objects in the dataset and solving the new optimization problem globally, which may be time-consuming. Also, as EVCLUS does not represent clusters by parametric models such as prototypes, the number of parameters grows linearly with the number of objects, which becomes problematic when the number of objects is very large (say, hundreds of thousands). Finally, a third limitation of EVCLUS  is that it does not easily incorporate side information in the clustering process. Semi-supervised versions of EVCLUS using pairwise constraints have been proposed \cite{antoine14,li18}, but the performances of these algorithms are limited because constraints imposed on some pairs of objects do not naturally propagate to neighboring objects, which is another consequence of the absence of a parametric model of clusters.

In this paper, we address the above limitations by proposing a new version of EVCLUS, called NN-EVCLUS, in which attributes are mapped to mass functions using a feedforward neural network. Continuing the analogy with MDS, this approach bears some resemblance with the SAMANN model  for Sammon's mapping \cite{mao95} or Webb's radial basis function network implementation of MDS \cite{webb95}. It is also related to ``Siamese'' networks for distance learning \cite{bromley93,yi14}. In NN-EVCLUS, the network weights are learnt by minimizing the discrepancy between degrees of conflict and dissimilarities over all pairs of objects, as in EVCLUS. However, the number of model parameters, equal to the number of weights in the network, is usually much less than that of EVCLUS.  The model can be used to predict the class of new objects, and it can be trained from very large datasets by stochastic gradient descent. It can also easily incorporate side information in the form of labeled data or pairwise constraints. 

The rest of this paper is organized as follows. Background knowledge on DS theory and evidential clustering is first recalled in Section \ref{sec:background}. The new model is then described in Section \ref{sec:approach} and numerical experiments are reported in Section \ref{sec:exper}. Finally, conclusions are presented in Section \ref{sec:concl}.

\section{Background}
\label{sec:background}

The purpose of this section is to make this paper self-contained. Necessary definitions and results related to DS mass functions will first be recalled in Section \ref{subsec:DS}. The notion of evidential partition and its relation with other notions of hard and ``soft'' partition will then be exposed in Section \ref{subsec:evidential_clustering}. Finally the EVCLUS algorithm will be summarized in Section \ref{subsec:evclus}.

\subsection{Mass functions}
\label{subsec:DS}

Let $\Omega=\{\omega_1,\ldots,\omega_c\}$ be a finite set. A \emph{mass function} on $\Omega$ is a mapping from the power set $2^\Omega$ to the interval $[0,1]$, such that
\[
\sum_{A\subseteq \Omega} m(A)=1.
\]
Each subset $A$ of $\Omega$ such that $m(A)>0$ is called a \emph{focal set} of $m$. In DS theory \cite{shafer76,denoeux20b},  $\Omega$ is the set of possible answers to some question (called the \emph{frame of discernment}), and a mass function $m$ describes a piece of evidence pertaining to that question. Each mass $m(A)$ represents  a  share  of  a  unit  mass  of  belief  allocated  to focal set $A$, and  which  cannot  be  allocated  to  any  strict  subset  of $A$.  A mass function $m$ is said to be \emph{logical} if it has only one focal set, \emph{consonant} of its focal sets are nested (i.e., for any two focal sets $A$ and $B$, we have either $A\subseteq B$ or $B\subseteq A$), and \emph{Bayesian} if its focal sets are singletons. A mass function that is both logical and Bayesian has only one singleton focal set: it is said to be \emph{certain}.

Just as a probability mass function induces a probability measure, a DS mass function induces two nonadditive measures:  a \emph{belief function}, defined as
\begin{equation}
Bel(A)=\sum_{\emptyset \neq B \subseteq A} m(B)
\end{equation}
for all $A \subseteq \Omega$ and a \emph{plausibility function} defined as
\begin{equation}
Pl(A)=\sum_{B \cap A \neq \emptyset} m(B).
\end{equation}
These two functions are linked by the relation $Pl(A)=Bel(\Omega)-Bel(\overline{A})$, for all $A \subseteq \Omega$.  The quantity $Bel(A)$ is a measure of how much subset $A$ is supported by the available evidence, while $Bel(\Omega)-Pl(A)=Bel(\overline{A})$ is a measure of how much the complement $\overline{A}$ is supported, so that $Pl(A)$ can be seen as a measure of lack of support for $\overline{A}$. When $m$ is consonant, the following equality holds for all subsets $A$ and $B$ of $\Omega$:
\[
Pl(A\cap B)=\max(Pl(A),Pl(B)).
\]
Function $Pl$ is, thus, a possibility measure \cite{zadeh78}. The function $pl: \Omega \rightarrow [0,1]$ that maps each element $\omega$ of $\Omega$ to its plausibility $pl(\omega)=Pl(\{\omega\})$  is called the \emph{contour function} associated to $m$. When $m$ is consonant, it is a possibility distribution.

\new{Given two mass functions $m_1$ and $m_2$ on the same frame of discernment $\Omega$, their conjunctive sum \cite{smets90a} is the following mass function:
\[
(m_1\cap m_2)(C)=\sum_{A\cap B=C} m_1(A)m_2(B)
\] 
for all $C\subseteq \Omega$. The \emph{degree of conflict} between $m_1$ and $m_2$ is then defined as the mass assigned to the empty set,}
\begin{equation}
\label{eq:conflict}
\kappa \new{=(m_1\cap m_2)(\emptyset)}=\sum_{A \cap B=\emptyset} m_1(A)m_2(B).
\end{equation}
It ranges in the interval [0,1]. When $m_1$ and $m_2$ represent two independent pieces of evidence pertaining to \emph{the same question}, $\kappa$ can be interpreted as a \emph{measure of conflict} between these two pieces of evidence \cite{shafer76}. In contrast, when $m_1$ and $m_2$ represent independent pieces of evidence about \emph{two distinct questions} $Q_1$ and $Q_2$ with the same  frame of discernment $\Omega$,  $\kappa$  can be given a different interpretation as one minus the plausibility that the true answers to $Q_1$ and $Q_2$ are identical   \cite{denoeux04b}. 

\begin{Ex}
Consider a population of objects partitioned in three classes, and let $\Omega=\{\omega_1,\omega_2,\omega_3\}$ denote the set of classes. Assume that a sensor provides information about the class of three objects $o_1$, $o_2$ and $o_3$  as the following three mass functions on $\Omega$:
\[
m_1(\{\omega_1\})= 0.6, \quad m_1(\{\omega_1,\omega_2\})= 0.3, \quad m_1(\Omega)=0.1
\]
\[
m_2(\{\omega_1,\omega_2\})= 0.5, \quad m_2(\{\omega_3\})=0.2, \quad m_2(\Omega)=0.3
\]
\[
m_3(\{\omega_1\})= 0.1, \quad m_3(\{\omega_2\})= 0.1, \quad m_3(\{\omega_3\})=0.8
\]
The degree of conflict between $m_1$ and $m_2$ is
\[
\kappa_{12}= 0.6\times 0.2+0.3\times 0.2=0.18,
\]
while the degree of conflict between $m_1$ and $m_3$ is
\[
\kappa_{13}= 0.6\times 0.1+0.6\times 0.8 + 0.3\times 0.8=0.06+0.54=0.78,
\]
Consequently, the plausibility that objects $o_1$ and $o_2$ belong to the same class is $1-\kappa_{12}=0.82$, whereas this plausibility is only 0.22 for objects $o_1$ and $o_3$.
\end{Ex}

\subsection{Evidential clustering}
\label{subsec:evidential_clustering}

Let $\calO=\{o_1,\ldots,o_n\}$ be a set of $n$ objects, such that each object is assumed to belong to at most one cluster in a set $\Omega=\{\omega_1,\ldots,\omega_c\}$. Using the formalism recalled in Section \ref{subsec:DS}, partial knowledge about the cluster membership of  object $o_i$ can be described by a mass function $m_i$ on $\Omega$. The $n$-tuple $\calM=(m_1,\ldots,m_n)$ is called an \emph{evidential} (or \emph{credal}) \emph{partition} of $\calO$.

The notion of evidential partition encompasses several classical clustering structures \citep{denoeux16b}:
\bi
\item When  all mass functions $m_i$ are certain, then $\calM$ is equivalent to a hard partition; this case corresponds to full certainty about the group of each object.
\item When mass functions are Bayesian, then $\calM$ boils down to a fuzzy partition; the degree of membership $u_{ik}$ of object $i$ to group $k$ is  then 
\[
u_{ik}=Bel_i(\{\omega_k\})=Pl_i(\{\omega_k\}) \in [0,1]
\]
and we have $\sum_{k=1}^c u_{ik}=1$.
\item When all mass functions $m_i$ are consonant, the corresponding contour functions $pl_i$ are possibility distributions and they define a possibilistic partition.
\item When each mass function $m_i$ is logical with focal set $A_i\subseteq \Omega$, $\calM$ is equivalent to a rough partition \citep{peters06}. The lower and upper  approximations of cluster $\omega_k$ are then defined, respectively, as the set of objects that \emph{surely} belong to group $\omega_k$, and the set of objects that \emph{possibly} belong to group $\omega_k$ \cite{masson08}; they are formally given by
\begin{equation}
\label{eq:lower_upper}
\omega^l_k:=\{i \in \calO \mid A_i=\{\omega_k\}\} \quad \textrm{and} \quad \omega^u_k:=\{i \in \calO\mid  \omega_k \in A_i\}.
\end{equation}
We then have $Bel_i(\{\omega_k\})=I(i \in \omega_k^l)$ and $Pl_i(\{\omega_k\})=I(i \in \omega_k^u)$, where $I(\cdot)$ denotes the indicator function. 
\ei

\begin{Ex}
\label{ex:butterfly}
Consider the \textsf{Butterfly} data displayed in Figure \ref{fig:butterfly}, consisting in 12 objects described by two attributes. Table \ref{tab:butterfly}  shows an evidential partition of these data obtained by EVCLUS, with $c=2$ clusters.  This evidential partition is represented graphically in Figure  \ref{fig:butterfly_credalpart}. We can see that  object 6, which is located between clusters $\omega_1$ and $\omega_2$, has the largest mass assigned to $\Omega=\{\omega_1,\omega_2\}$. In contrast, object 12, which is an outlier, has the largest mass assigned to the empty set. A convenient way to summarize an evidential partition is to approximate each mass function $m_i$ by a logical mass function $\mh_i$ such that $\mh_i(A_i)=1$ with $A_i=\arg \max_A m_i(A)$. We can then compute the lower and approximations of each cluster using \eqref{eq:lower_upper}. Here, we have $\omega^l_1=\{7,8,9,10,11\}$, $\omega^u_1=\{6,7,8,9,10,11\}$, $\omega^l_2=\{1,2,3,4,5\}$ and $\omega^u_2=\{1,2,3,4,5,6\}$. The objects that do not belong to the upper approximation of any cluster are outliers, which is the case for object 12 in this example.

\begin{figure}
\centering  
\subfloat[\label{fig:butterfly}]{\includegraphics[width=0.35\textwidth]{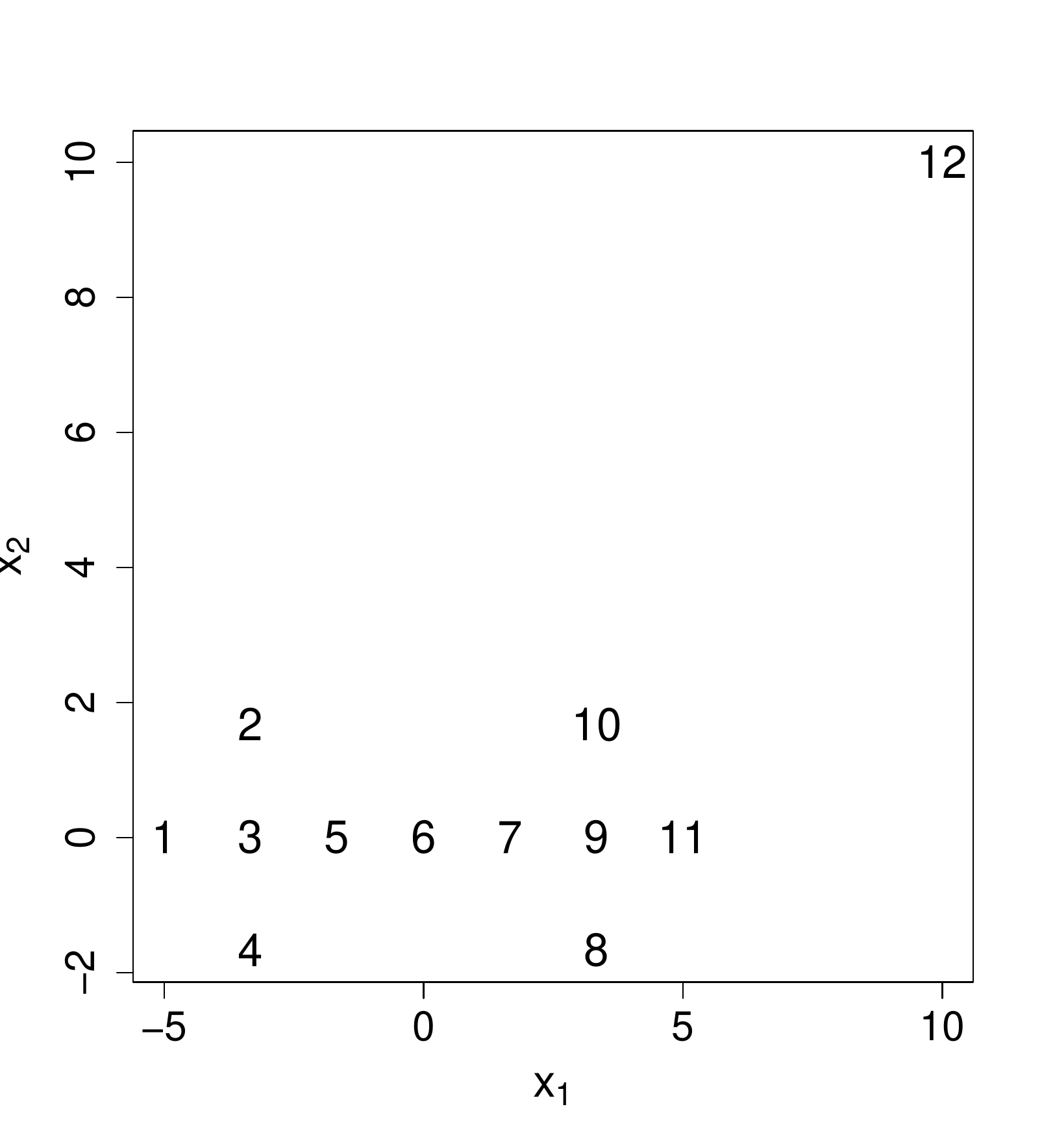}}
\subfloat[\label{fig:butterfly_credalpart}]{\includegraphics[width=0.35\textwidth]{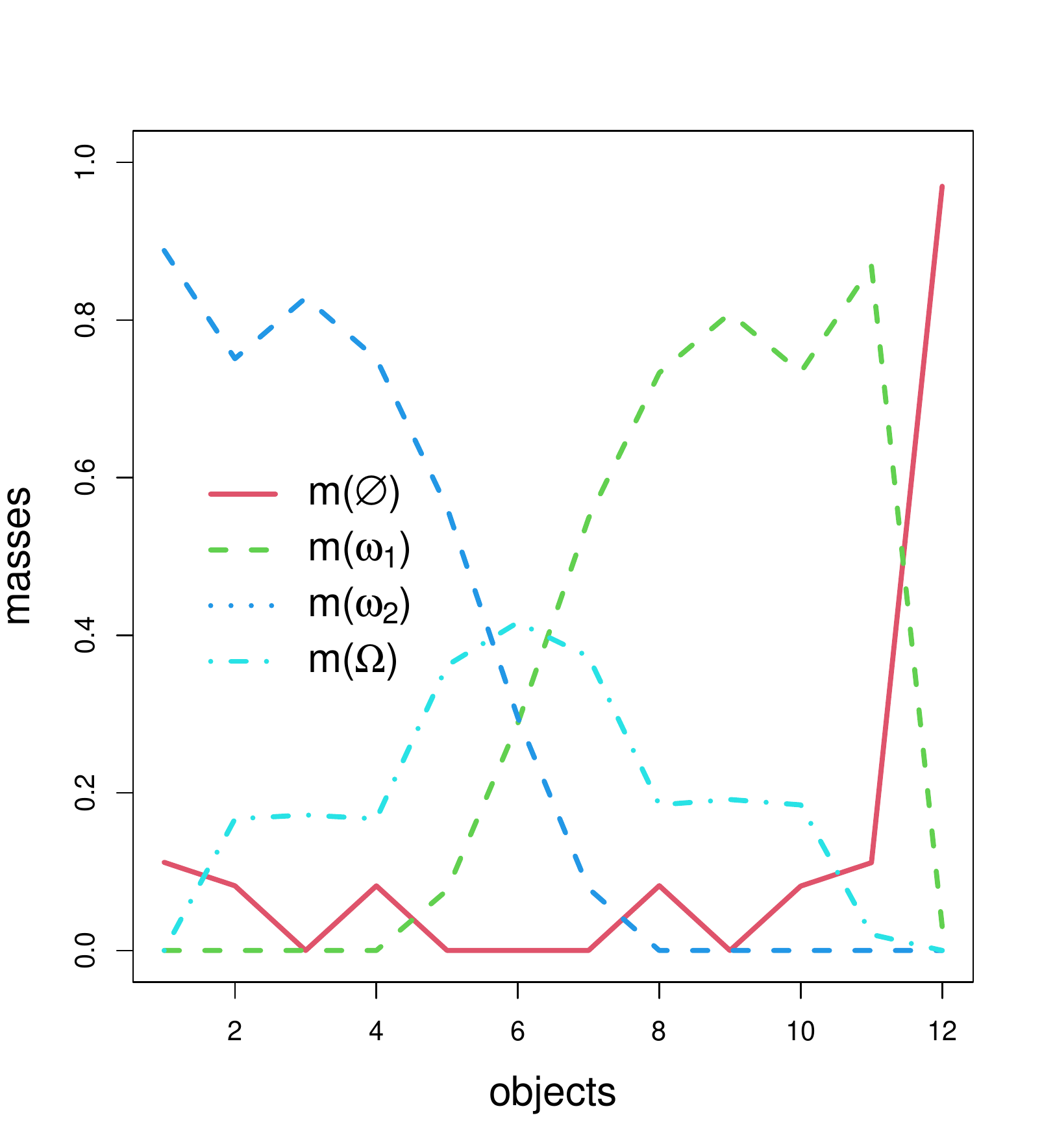}}
\caption{Butterfly dataset (a) and evidential partition with $c=2$ obtained by ECM (b).   \label{fig:ex1}}
\end{figure}

\begin{table}
\caption{Evidential partition of the \textsf{Butterfly} data. The largest mass for each object is printed in bold. \label{tab:butterfly}}
\begin{center}
\begin{tabular}{ccccc}
\hline
object \# & $m(\emptyset)$ & $m(\{\omega_1\})$ & $m(\{\omega_2\})$ & $m(\Omega)$\\
\hline
1 & 0.11  & $\cdot$ &\bf{0.89} &$\cdot$\\
2 & 0.082 & $\cdot$ &\bf{0.75} &0.17\\
3  &0.000 &$\cdot$ &\bf{0.83} &0.17\\
4 &0.082 &$\cdot$ &\bf{0.75} &0.17\\
5 &$\cdot$ &0.077 &\bf{0.56} &0.36\\
6 &$\cdot$ &0.29 &0.30 &\bf{0.42}\\
7 &$\cdot$ &\bf{0.55} &0.079 &0.37\\
8 &0.082 &\bf{0.73} &$\cdot$ &0.18\\
9 &0.000 &\bf{0.81}&$\cdot$ &0.19\\
10 &0.082 &\bf{0.73} &$\cdot$ &0.18\\
11 &0.11 &\bf{0.87} &$\cdot$ &0.02\\
12 &\bf{0.97} &0.030 &$\cdot$ &$\cdot$\\
\hline
\end{tabular}
\end{center}
\end{table}
\end{Ex}

\subsection{EVCLUS algorithm}
\label{subsec:evclus}

Evidential clustering aims at generating an optimal evidential partition from attribute or dissimilarity data, based on some optimality criterion. The earliest such procedure is  EVCLUS\footnote{EVCLUS is implemented with other evidential clustering algorithms in the R package {\tt evclust} \cite{denoeux21} available at \url{https://cran.r-project.org}.}, which was introduced in \cite{denoeux04b} and later improved in \cite{denoeux16a}. EVCLUS transposes some ideas from MDS \cite{borg97} to clustering. Let $\bD=(\delta_{ij})$ be a symmetric  $n\times n$ dissimilarity matrix, where $\delta_{ij}$ denotes the dissimilarity between objects $o_i$ and $o_j$. Dissimilarities may be  computed from attribute data, or they may be directly available. They \new{need not} satisfy the axioms of distances such as the triangular inequality, \new{i.e., we may have $\delta_{ik} > \delta_{ij} + \delta_{k,j}$ for some triple of objects $(i,j,k)$}. 

The fundamental assumption underlying EVCLUS is that the more similar are two objects, the more plausible it is that they belong to the same cluster. As recalled in Section \ref{subsec:DS}, the plausibility $pl_{ij}$ that two objects $o_i$ and $o_j$ belong to the same cluster is equal to $1-\kappa_{ij}$, where $\kappa_{ij}$ is the degree of conflict between $m_i$ and $m_j$. The credal partition $\calM$ should thus be determined in such a way that similar objects have mass functions $m_i$ and $m_j$ with low degree of conflict, whereas highly dissimilar objects are assigned highly conflicting mass functions.  To derive an evidential partition $\calM=(m_1,\ldots,m_n)$ from $\bD$, we can thus  minimize the discrepancy between the pairwise degrees of conflict and the dissimilarities, up to some increasing transformation. This problem bears some resemblance with the one addressed by  MDS, which aims to represent objects in some Euclidean space, in such a way that the distances in that space match the observed dissimilarities \cite{borg97}. Here, we want to find an evidential partition $\calM$ that minimizes the following loss function,
\begin{equation}
\label{eq:lossEVCLUS}
\calL(\calM)=\frac{2}{n(n-1)} \sum_{i<j} \left(\kappa_{ij}-\varphi(\delta_{ij})\right)^2,
\end{equation}
where $\varphi$ is a fixed nondecreasing mapping from $[0,+\infty)$ to $[0,1]$, such as
\begin{equation}
\label{eq:phi}
\varphi(\delta)=1-\exp(-\gamma \delta^2),
\end{equation}
for some  user-defined coefficient $\gamma$. In \cite{denoeux16a}, we proposed to set  $\gamma=-\log 0.05/\delta_0^2$, where $\delta_0$ is some quantile of the dissimilarities $\delta_{ij}$. This parametrization ensures that $\varphi(\delta)\ge 0.95$ whenever $\delta\ge \delta_0$, i.e., $\delta_0$ is the threshold such that two objects $o_i$ and $o_j$ with dissimilarity larger than  $\delta_0$ have a plausibility at least $0.95$ of belonging to different clusters. In \cite{denoeux16a}, we recommended to set $\delta_0$ to the  0.9-quantile as the default value, but this parameter sometimes needs to be fine-tuned to get optimal results. 

Computing the loss function \eqref{eq:lossEVCLUS} requires to store the whole dissimilarity matrix, which may not be feasible for large datasets. In \cite{denoeux16a}, it was shown that it is often sufficient to minimize the sum of squared errors for a subset of object pairs. We can thus replace  \eqref{eq:lossEVCLUS} by
\begin{equation}
\label{eq:lossEVCLUS1}
\calL(\calM;J)=\frac{1}{np} \sum_{i=1}^n\sum_{j\in J(i)} \left(\kappa_{ij}-\varphi(\delta_{ij})\right)^2,
\end{equation}
where $J(i)$ is \new{a randomly selected subset of $p<n-1$ indices  in $\{1,\ldots,i-1,i+1,\ldots,n\}$. The calculation of  $\calL(\calM;J)$ thus requires to store only $np$ dissimilarities $\delta_{ij}$  instead of $n(n-1)/2$ for the calculation of $\calL(\calM)$ in \eqref{eq:lossEVCLUS}.}
Experiments reported in \cite{denoeux16a} show that, for a number $n$ of objects between 1000 and 10,000, optimal results are obtained with $p$ in the range 100-500.

In \cite{denoeux04b}, it was originally proposed to minimize a loss function similar to \eqref{eq:lossEVCLUS} using a gradient-based algorithm. A much more efficient cyclic coordinate descent  procedure, called Iterative Row-wise Quadratic Programming (IRQP)  \cite{terbraak09} was proposed in \cite{denoeux16a}. This    procedure  consists in minimizing the loss with respect to one mass function $m_i$ at a time while keeping the other mass functions fixed, which can be shown to be a linearly constrained least-squares problem.  The IRQP algorithm together with the loss function \eqref{eq:lossEVCLUS1} allow EVCLUS  to cluster datasets containing up to  a few tens of thousands of objects.

\begin{Ex}
\label{ex:evclus}
The \textsf{fourclass} dataset\footnote{This dataset is part of the R package {\tt evclust}.} is composed of 400 two-dimensional vectors generated from four two-dimensional Student distributions.  Figure \ref{fig:fourclass_credpart} displays the lower and upper approximations of each of the four clusters computed from an evidential partition obtained by EVCLUS, with loss function \eqref{eq:lossEVCLUS1} and $p=100$. The focal sets were restricted to subsets of cardinality less than or equal to two, and $\Omega$. As shown in Figure \ref{fig:convergence_fourclass},  the algorithm converges in a few dozens of iterations. Figure \ref{fig:shepard_fourclass} shows the Shepard diagram, which displays the degrees of conflict $\kappa_{ij}$ versus the transformed dissimilarities $\varphi(d_{ij})$.

\begin{figure}
\centering  
\includegraphics[width=0.9\textwidth]{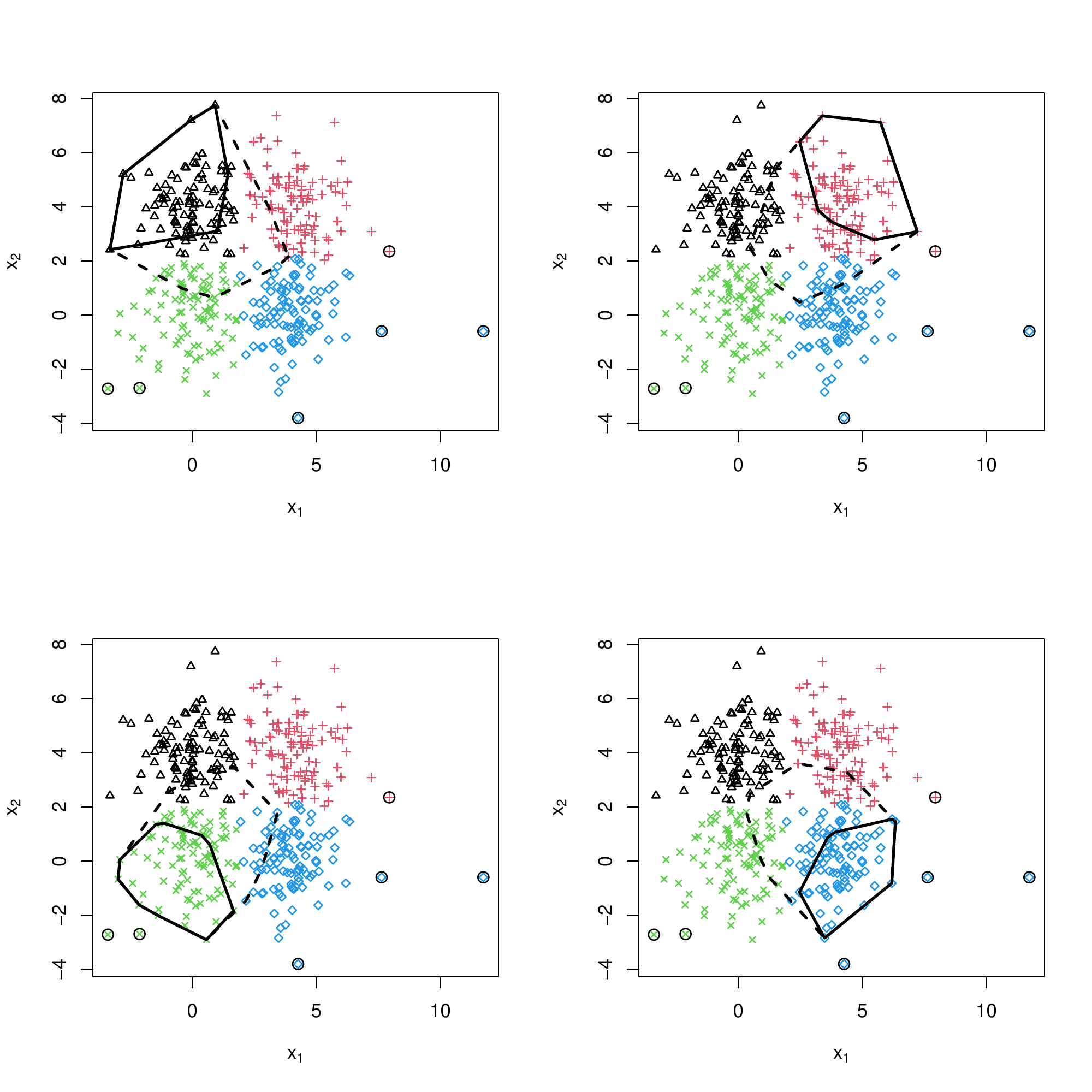}
\caption{Lower and upper approximations of the four clusters found by EVCLUS for  the \textsf{fourclass} dataset. The true classes are displayed with different colors. The identified clusters are plotted with different symbols. The convex hulls of the cluster lower and upper approximations  are displayed using solid and interrupted lines, respectively. The  outliers are indicated by circles.   \label{fig:fourclass_credpart}}
\end{figure}

\begin{figure}
\centering  
\subfloat[\label{fig:convergence_fourclass}]{\includegraphics[width=0.35\textwidth]{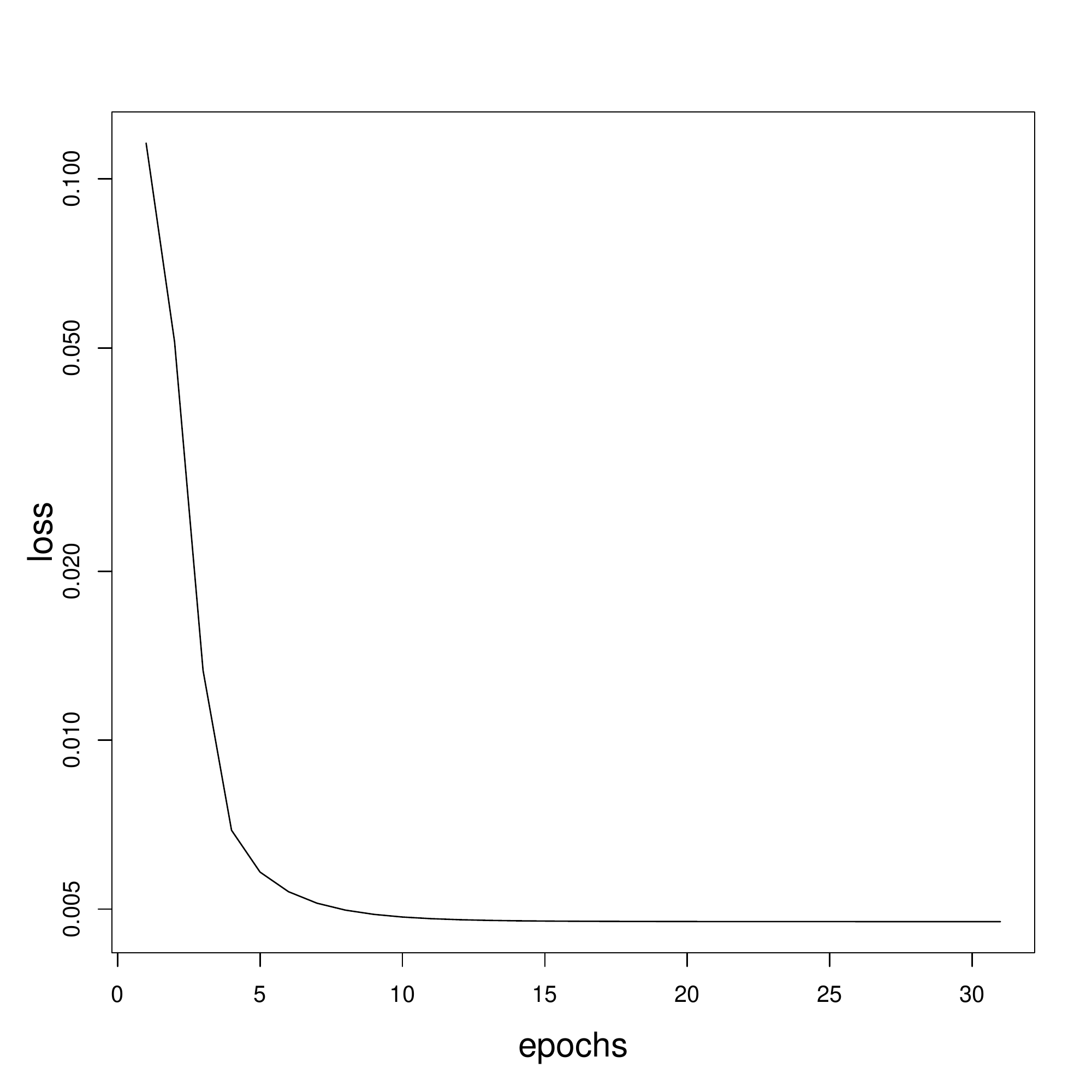}}
\subfloat[\label{fig:shepard_fourclass}]{\includegraphics[width=0.35\textwidth]{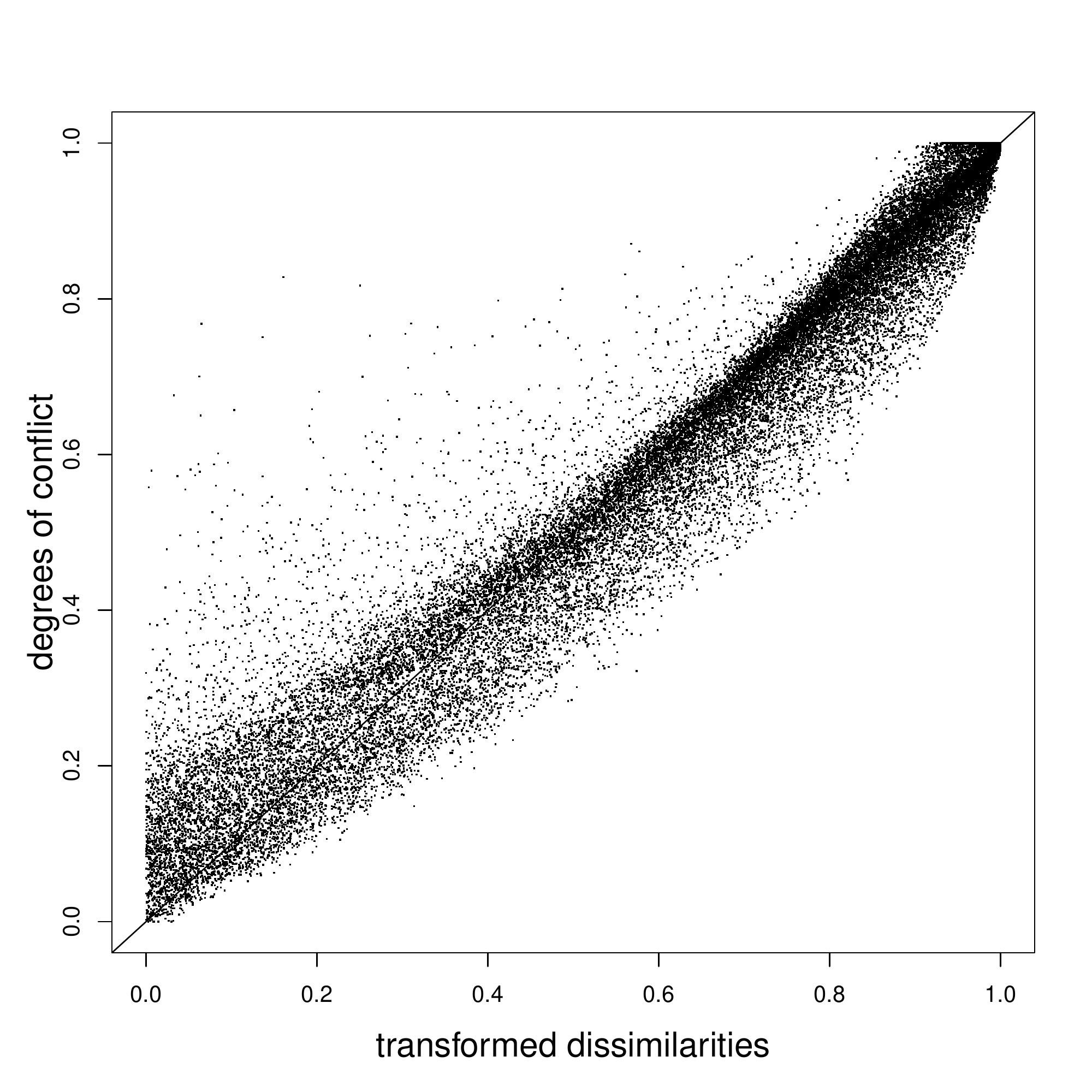}}
\caption{EVCLUS algorithm applied to the \textsf{fourclass} dataset: (a): loss vs. number of epochs (iterations thorugh the whole learning set);  (b): plot of the degrees of conflict $\kappa_{ij}$ vs. the transformed dissimilarities $\delta^*_{ij}$.   \label{fig:fourclass}}
\end{figure}

\end{Ex}

\section{NN-EVCLUS}
\label{sec:approach}

As opposed to prototype-based evidential clustering algorithms such as ECM \cite{masson08} or CCM \cite{liu15}, the EVCLUS algorithm summarized in Section \ref{subsec:evclus} does not build a compact representation of clusters as a collection of prototypes, but it learns an evidential partition of the $n$ objects directly. If each mass function is constrained to have $f$ focal sets, the number of free parameters is, thus, $n(f-1)$, i.e., it grows linearly with the number of objects. This characteristic makes EVCLUS impractical for clustering very large datasets. Also, the algorithm learns an evidential partition of a given dataset, but it does not allow us to extrapolate beyond the learning set and make predictions for new objects. In this section, we describe a neural network version of EVCLUS that addresses these issues. This new model will also be shown to outperform EVCLUS and ECM in semi-supervised clustering tasks. The model will first be introduced in Section \ref{subsec:model}, and learning algorithms will be described in Section \ref{subsec:learning}. Finally,  semi-supervised learning will be addressed  in Section \ref{subsec:semi}.

\subsection{Model}
\label{subsec:model}

\paragraph{Learning data} We assume the learning data to consist in
\bi
\item A dissimilarity matrix $\bD=(\delta_{ij})$;
\item A collection of $n$ vectors $\calX=(\bx_1,\ldots,\bx_n)$, each vector  $\bx_i$ being composed of $d$ attributes describing object $o_i$. 
\ei
Most of the time, we get the $n$ attribute vectors first and compute $\bD$ as, for instance, the matrix of Euclidean distances between vectors $\bx_i$:
\[
\delta_{ij}=\|\bx_i-\bx_j\|.
\]
Sometimes, it may be advantageous to compute the dissimilarities using not only the attribute vectors, but also side information such as must-link and cannot-link constraints. This case will be investigated in Section \ref{subsec:semi}. The most delicate situation is that of pure dissimilarity data, i.e.  data consisting only in the dissimilarity matrix $\bD$. Even then, we may  still be able to construct a collection of attribute vectors $(\bx_1,\ldots,\bx_n)$ by applying Principal Component Analysis (PCA) to the dissimilarity matrix. Examples of this approach will be presented in Section \ref{subsec:rel}.

\paragraph{Basic principle} Any mass function $m$ on $\Omega$ with $f$ focal sets $F_1,\ldots,F_f$ can be represented by  a mass vector $\bm\in[0,1]^f$. The  basic idea behind our approach is to learn a mapping from attribute vectors to mass vectors, so as to minimize a loss function such as \eqref{eq:lossEVCLUS} or \eqref{eq:lossEVCLUS1}. For this, we can define a parametrized family of mappings
\[
\calG=\{g(\cdot;\btheta): \reels^d \rightarrow [0,1]^f \mid \btheta\in \Theta\}.
\]
For each choice of $\btheta$, we then have an evidential partition $\calM(\btheta)=(\bm_1(\btheta),\ldots,\bm_n(\btheta))$, with $\bm_i(\btheta)=g(\bx_i;\btheta)$. Let $\kappa_{ij}(\btheta)$ be the conflict between $\bm_i(\btheta)$ and $\bm_j(\btheta)$; it can be written using matrix notations as
\begin{equation}
\label{eq:conflict1}
\kappa_{ij}(\btheta) = \bm_i(\btheta)^T \bC \bm_j(\btheta),
\end{equation}
where $\bC$ is the symmetric $f\times f$ matrix  with general term $C_{qr}=I(F_q\cap F_r=\emptyset)$. We can determine $\btheta$ so as to minimize a loss function such as
\begin{equation}
\label{eq:lossEVCLUS2}
\calL(\btheta)=\frac{2}{n(n-1)} \sum_{i<j} \left(\kappa_{ij}(\btheta)-\delta^*_{ij}\right)^2,
\end{equation}
where $\delta^*_{ij}=\varphi(\delta_{ij})$ denotes the transformed dissimilarities. Let 
\[
\bthetah=\arg \min_\btheta \calL(\btheta)
\]
be the solution to this optimization problem. This approach allows us to predict the cluster membership of a new object with attribute vector $\bx$ by the mass vector $\bm=g(\bx;\bthetah)$.  

\paragraph{Neural network model} A common choice for a parametrized family of function $\calG$ is a multi-layer feedforward neural network (NN) model \cite{goodfellow16}. Without loss of generality, let us consider a network with one hidden layer of $n_H$ neurons with rectified linear units and a softmax output layer. The activation $a_h$ and the output $z_h$ of hidden unit $h\in \{1,\ldots,n_H\}$ in such a network are defined, respectively, as
\begin{subequations}
\label{eq:propag1}
\begin{equation}
a_h=v_{h0}+\sum_{k=1}^d v_{hk}x_k 
\end{equation}
and
\begin{equation}
z_h=\max(0,a_h),
\end{equation}
\end{subequations}
where $v_{hk}$ is the weight of the connection between input $k$ and hidden unit $h$ and $v_{h0}$ is a bias term. Similarly, the activation of output unit $q\in \{1,\ldots,f\}$ is
\begin{subequations}
\label{eq:propag2}
\begin{equation}
\mu_q=w_{q0}+\sum_{h=1}^{n_H} w_{qh}z_h,
\end{equation}
where $w_{qh}$ is the weight of the connection between hidden unit $h$ and output unit $q$ and $w_{q0}$ is a bias term. The corresponding  output of unit $q$ is finally
\begin{equation}
m_q=\frac{\exp(\mu_q)}{\sum_{r=1}^f \exp(\mu_r)}.
\end{equation}
\end{subequations}

\paragraph{Outlier detection} NN models such as described above usually have very good performances, but they provide arbitrary predictions for inputs that lie in regions with no training data. A trained NN might thus not be able to detect outliers in a test set of previously unseen input vectors. (We recall that, in EVCLUS, outliers are identified by a large mass assigned to the empty set). To address this issue, we propose to optionally couple a feedforward NN with a one-class support vector machine (SVM) \cite{sholkopf02}. \new{A one-class SVM is an unsupervised learning method that constructs a ``simple'' region $R$ of the input space containing a large fraction of the data. This region is  described by a function $f$ that returns a value $f(\bx)>0$ when $\bx$ belongs to $R$, and  $f(\bx)<0$ when $\bx$ belongs to the complement of  $R$.} Function $f$ has the following expression
\[
f(\bx)=\alpha_0+\sum_{i\in \textsf{SV}} \alpha_i K(\bx,\bx_i) ,
\]
where $K(\cdot,\cdot)$ is a kernel function \new{fixed a priori}, $\textsf{SV}\subset \{1,\ldots,n\}$ is the set of indices of the support vectors, and the $\alpha_i$ are coefficients. \new{The coefficients $\alpha_0$ and $\alpha_i$ as well as the support vectors are learnt by minimizing a loss function.} Thanks to the kernel trick, one-class SVMs can adapt to arbitrarily complex input vector distributions and \new{make it possible to detect new input vectors generated from a different distribution (a problem often referred to as ``novelty detection'' \cite{sholkopf02}). }

\begin{figure}
\centering  
\includegraphics[width=0.6\textwidth]{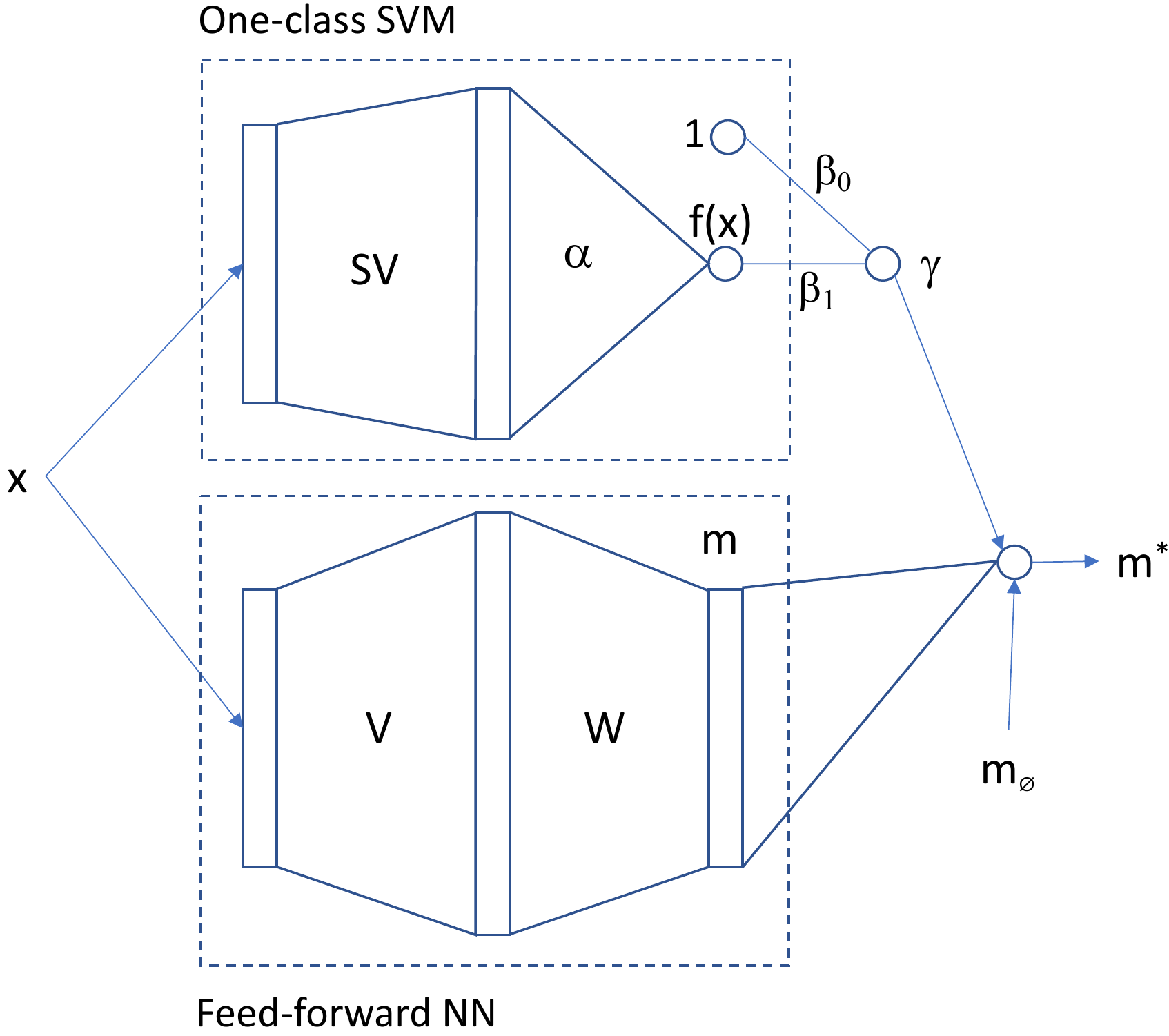}
\caption{Complete NN-EVCLUS model composed of a multi-layer NN and a one-class SVM. The output for input vector $\bx$ is a mass vector $\bm^*=\gamma \bm + (1-\gamma)\bm_\emptyset$, where   $\bm=g(\bx)$ is the NN output and $\gamma$ is a coefficient computed as an increasing function of the linearly transformed one-class SVM output $\beta_0+\beta_1 f(\bx)$. \label{fig:model}}
\end{figure}

Let $\bm=g(\bx)$ be the mass vector computed by the NN for input $\bx$, and $f(\bx)$ the output of the one-class SVM. We define a transformed mass function $\bm^*$ as
\begin{equation}
\label{eq:mstar}
\bm^*=  \gamma \bm + (1-\gamma) \bm_\emptyset,
\end{equation}
where $\bm_\emptyset$ is the mass vector corresponding to the mass function $m_\emptyset$ such that $m_\emptyset(\emptyset)=1$, and $\gamma \in [0,1]$ is a coefficient defined as an increasing function of $f(\bx)$ such that $\gamma\rightarrow 1$ when $f(\bx)\rightarrow +\infty$ and $\gamma\rightarrow 0$ when $f(\bx)\rightarrow -\infty$ . For instance, we can define $\gamma$ as
\begin{subequations}
\label{eq:gamma}
\begin{equation}
\gamma=\frac{\eta}{1+\eta}
\end{equation}
where $\eta$ is related to an affine function of $f(\bx)$ by the ``softplus'' mapping \cite{goodfellow16}
\begin{equation}
\eta = \log\left[1+\exp(\beta_0+\beta_1 f(\bx))\right].
\end{equation}
\end{subequations}
The complete model is illustrated in Figure \ref{fig:model}.

\subsection{Learning}
\label{subsec:learning}

\begin{figure}
\centering  
\includegraphics[width=0.5\textwidth]{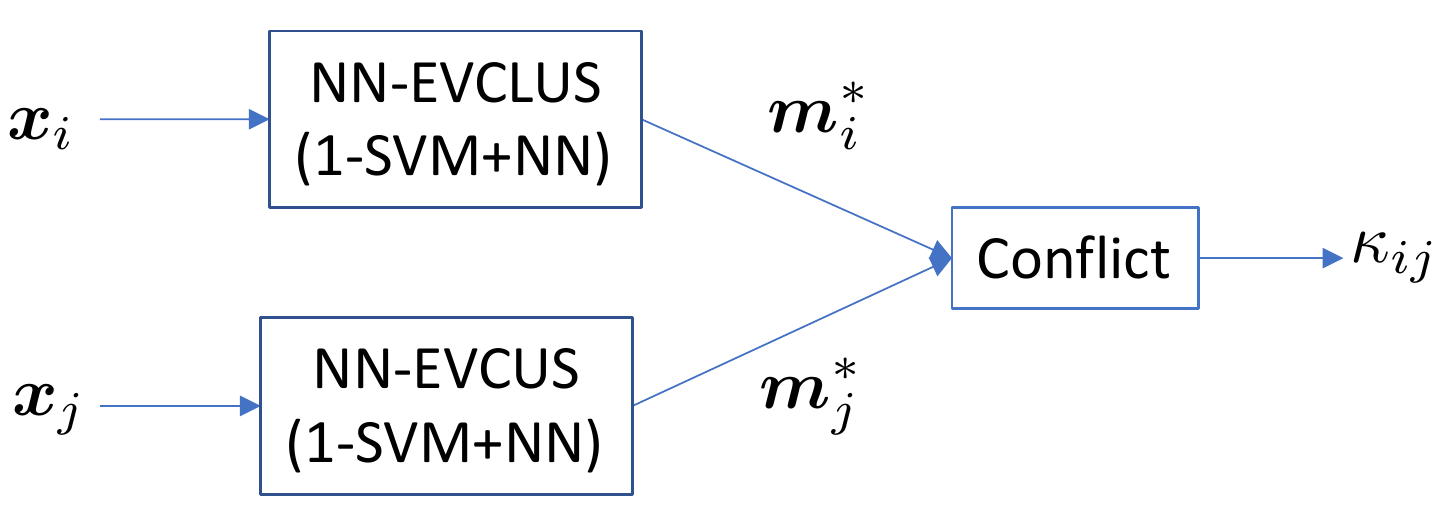}
\caption{The learning process of EVCLUS seen as two identical ``Siamese'' networks operating in parallel. \label{fig:learning}}
\end{figure}

The one-class SVM in the above model can be trained separately using  standard algorithms described in \cite{sholkopf02}. Here, we focus on the training of the NN component, which can be done by minimizing a loss function such as \eqref{eq:lossEVCLUS2}, which is the average over all objet pairs $(i,j)$ of  the error 
\begin{equation}
\label{eq:lossEVCLUS3}
\calL_{ij}(\btheta)= \left(\kappa_{ij}(\btheta)-\delta^*_{ij}\right)^2.
\end{equation}
We can remark that the error is not computed  for every instance as in standard NN training, but for every pair of instances. We can view the learning process as two identical (or ``Siamese''  \cite{bromley93}) networks operating in parallel, as illustrated in Figure \ref{fig:learning}. For each $(\bx_i,\bx_j)$, one input $\bx_i$ is presented to the first network and $\bx_j$ is presented to the second one. The degree of conflict $\kappa_{ij}$ between output mass functions $m_i^*$ and $m_j^*$ is then computed using \eqref{eq:conflict1}, and the error is defined as the squared difference between $\kappa_{ij}$ and the transformed dissimilarity $\delta^*_{ij}$.

\begin{figure}
\centering  
\includegraphics[width=0.5\textwidth]{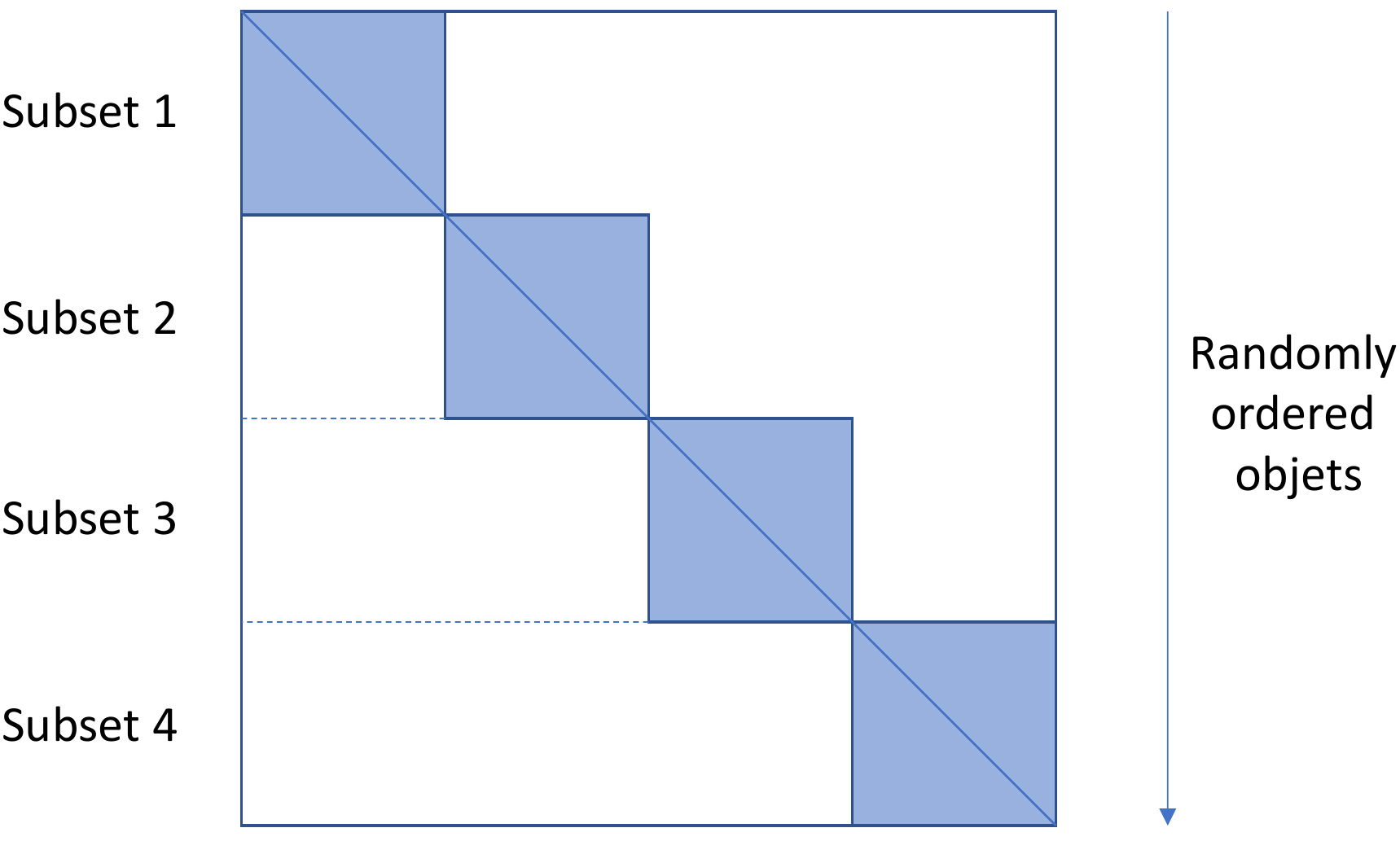}
\caption{Illustration of the sampling procedure. The objects are randomly ordered and partitioned into $s$ subset ($s=4$ in the figure). Distances within each subsets are computed. This procedure gives us $s$ mini-batches of approximately $(n/s)[(n/s)-1]/2$ pairs. (Only half of the distance sub-matrices need to be computed as they are assumed to be symmetric). \label{fig:minibatch}}
\end{figure}

The calculation of the error derivatives is detailed in  \ref{sec:error_grad}. The learning can be done in batch mode for small or medium-size datasets or using mini-batch stochastic gradient descent (SGD) for large datasets. In batch mode, we can directly minimize the average loss \eqref{eq:lossEVCLUS2} over the whole distance matrix, or over a subset of distances as done in EVCLUS (see Eq. \eqref{eq:lossEVCLUS1}). In the latter case, the average loss is
\begin{equation}
\label{eq:lossEVCLUS4}
\calL(\btheta)=\frac1{np} \sum_{i=1}^n \sum_{j\in J(i)} \calL_{ij}(\btheta),
\end{equation}
where $J(i)$ is a subset of $\{1,\ldots,n\}$ of cardinality $p$, randomly selected before the learning process. Minimizing \eqref{eq:lossEVCLUS4} instead of \eqref{eq:lossEVCLUS2} allows us to store only $np$ distances instead of $n(n-1)/2$ and to accelerate the calculations. To implement mini-batch SGD, we need to randomly sample $q$ pairs $(\bx_i,\bx_j)$ and average the gradient of $\calL_{ij}$ over these $q$ pairs before each  weight update. This sampling can be done in several ways. One approach is to randomly order the objects, partition them in $s$ subsets of approximately equal size $n/s$, and compute all the pairwise distances within each subset. This gives us $s$ mini-batches of approximately $(n/s)[(n/s)-1]/2$ pairs $(i,j)$. This sampling procedure is illustrated in Figure \ref{fig:minibatch}.

\paragraph{Architecture design and regularization} To implement the NN-EVCLUS method, we need to select the number $c$ of clusters,  the focal sets and the NN architecture (number of layers and number of units per layer). 

To determine the number of clusters, we can use automatic selection criteria such as the nonspecificity measure proposed in \cite{masson08}, or an interactive approach based on  visualization techniques, such as the $\delta-Bel$ graph proposed in \cite{su19}. The latter approach is less computationally intensive and often more reliable than the former; it will be adopted in this paper.  

The choice of the set $\calF$ of focal sets depends on the number $c$ of clusters. For small values of $c$ (say, $c\le 5$), all $2^c$ focal sets can be considered. For small and medium values of $c$ (typically, $c \le 10$), we can  consider only the  subsets of cardinality less than or equal to 2, and the frame $\Omega$. For large $c$, a common choice is to keep only the empty set, the singletons, and $\Omega$. Alternatively, we can adopt the two-step strategy proposed in \cite{denoeux16a}, in which we first fit the model with singletons (as well as $\emptyset$ and $\Omega$), and then include selected pairs of neighboring clusters in a second step.

As far as the network architecture is concerned, we have  found one hidden layer to be sufficient for most datasets. However, it is possible that very complex datasets would require two hidden layers or more. Assuming one hidden layer, the next decision to make is on the number  $n_H$ of hidden units. If the emphasis is on clustering a single data set, we need not concern ourselves with generalization and we can choose $n_H$ large enough to reach a small enough discrepancy between degrees of conflict and transformed distances. If we can run EVCLUS, then the loss reached by EVCLUS can be considered as a lower bound of the loss achievable by NN-EVCLUS (as the latter model has fewer free parameters). If we intend to use the model for prediction, then some complexity control technique should be used. A common approach of regularization: for instance, $\ell_2$ regularization combined with \eqref{eq:lossEVCLUS4} gives us the following regularized  loss function:
\[
\calL_R(\btheta)=\frac1{np} \sum_{i=1}^n \sum_{j\in J(i)} \calL_{ij}(\btheta) +  \frac{\lambda}2 \left(\frac{1}{n_H(d+1)} \sum_{h,k} v_{hk}^2 + \frac{1}{f(n_H+1)} \sum_{q,h} w_{qh}^2\right),
\]
where hyperparameter $\lambda$ can be tuned by cross-validation or using the hold-out method. When using mini-batch SGD, regularization can, alternatively, be obtained by the early stopping technique (interrupting the learning process when the loss computed on a validation set starts increasing).

\new{
\paragraph{Complexity} From Eqs. \eqref{eq:conflict1} and \eqref{eq:lossEVCLUS3}-\eqref{eq:lossEVCLUS4}, the complexity of computing the loss function in batch mode is $O(nkf^2)$, and from Eqs \eqref{eq:dLijdwqh}-\eqref{eq:Deltapijh}, the gradient of the loss function with respect to the weights (for a network with one hidden layer) can be computed in $O(nkn_Hf(f+ d))$ operations. 
The complexity is, thus,  proportional to the number $n$ of objects and to $f^2$,  which confirms the necessity of limiting the number $f$ of focal sets when the number $c$ of clusters is large. Keeping only the singletons, the empty set and $\Omega$, we have $f=c+2$, and the number of operations for the gradient calculation becomes proportional to $c^2$. 
}
\begin{Ex}
\label{ex:fourclass_nnevclus}
As an example, we consider again the \textsf{fourclass} dataset of Example \ref{ex:evclus}. As this dataset has only two features it is easy to determine the number of clusters by just displaying the data. The four clusters are also very clearly visible in the $\delta-Bel$ graph shown in Figure \ref{fig:deltaBel}. In this graph, $Bel$ denotes the degree of belief that an attribute vector is a cluster center, and $\delta$ is the minimum distance to vectors with a higher value of $Bel$ \cite{su19}. Cluster centers are typically located in the  upper-right corner of the graph.  

\begin{figure}
\centering  
\includegraphics[width=0.4\textwidth]{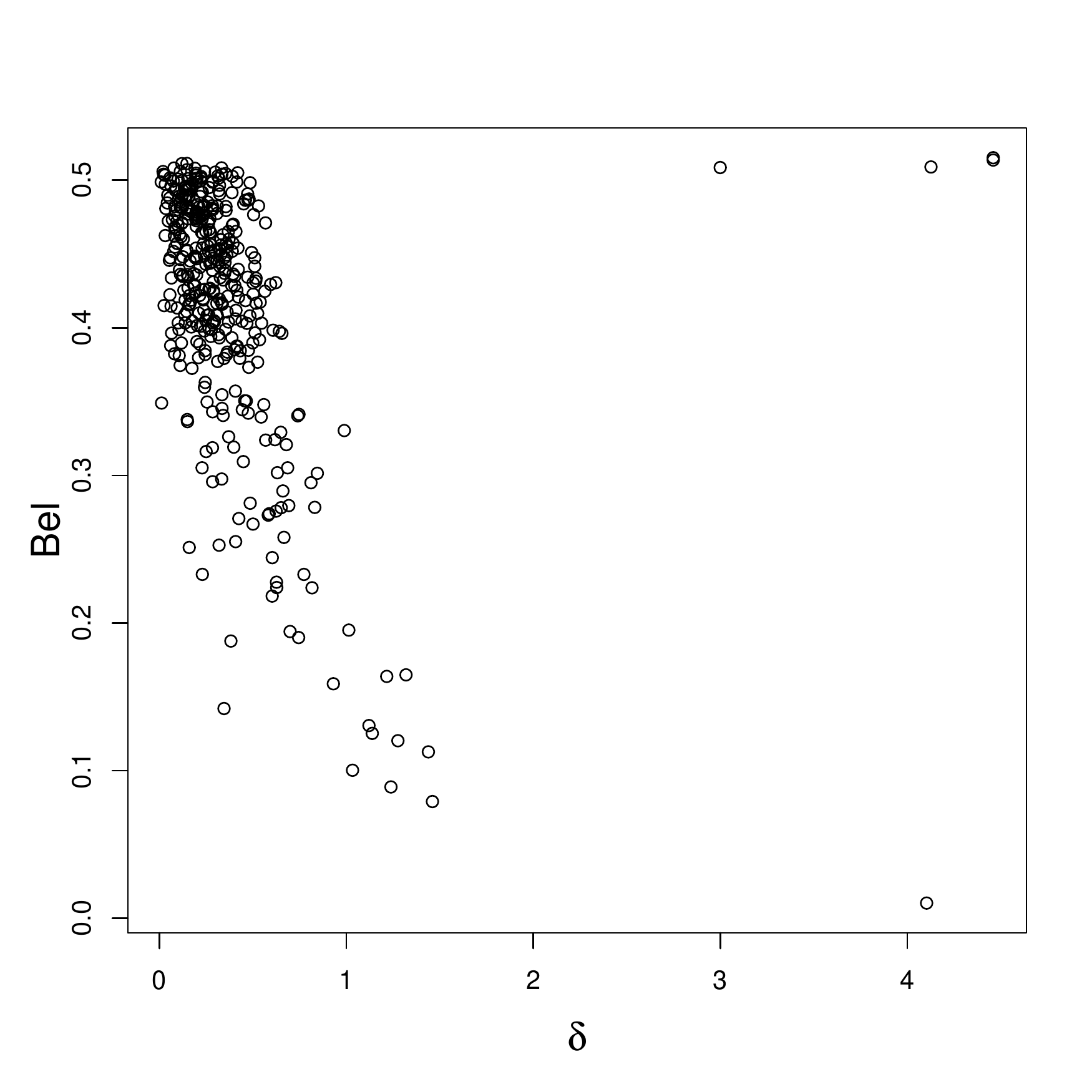}
\caption{$\delta-Bel$ plot of the \textsf{fourclass} dataset, with $K=50$ neighbors and $q=0.9$. The four cluster centers correspond to the four points in the upper-right corner of the graph. (Note that two points are very close and almost indiscernable). \label{fig:deltaBel}}
\end{figure}

For the one-class SVM part we used the $\nu$-SVM algorithm in R package {\tt kernlab} \cite{karatzoglou04} with a Gaussian kernel. This method has two hyperparameters: $\nu$,  which is an upper bound on the fraction of outliers, and the kernel width $\sigma$. We set $\nu=0.2$ and $\sigma=0.2$. Contours of the SVM output $f(\bx)$ are shown in Figure \ref{fig:fhat}.

\begin{figure}
\centering  
\subfloat[\label{fig:fhat}]{\includegraphics[width=0.4\textwidth]{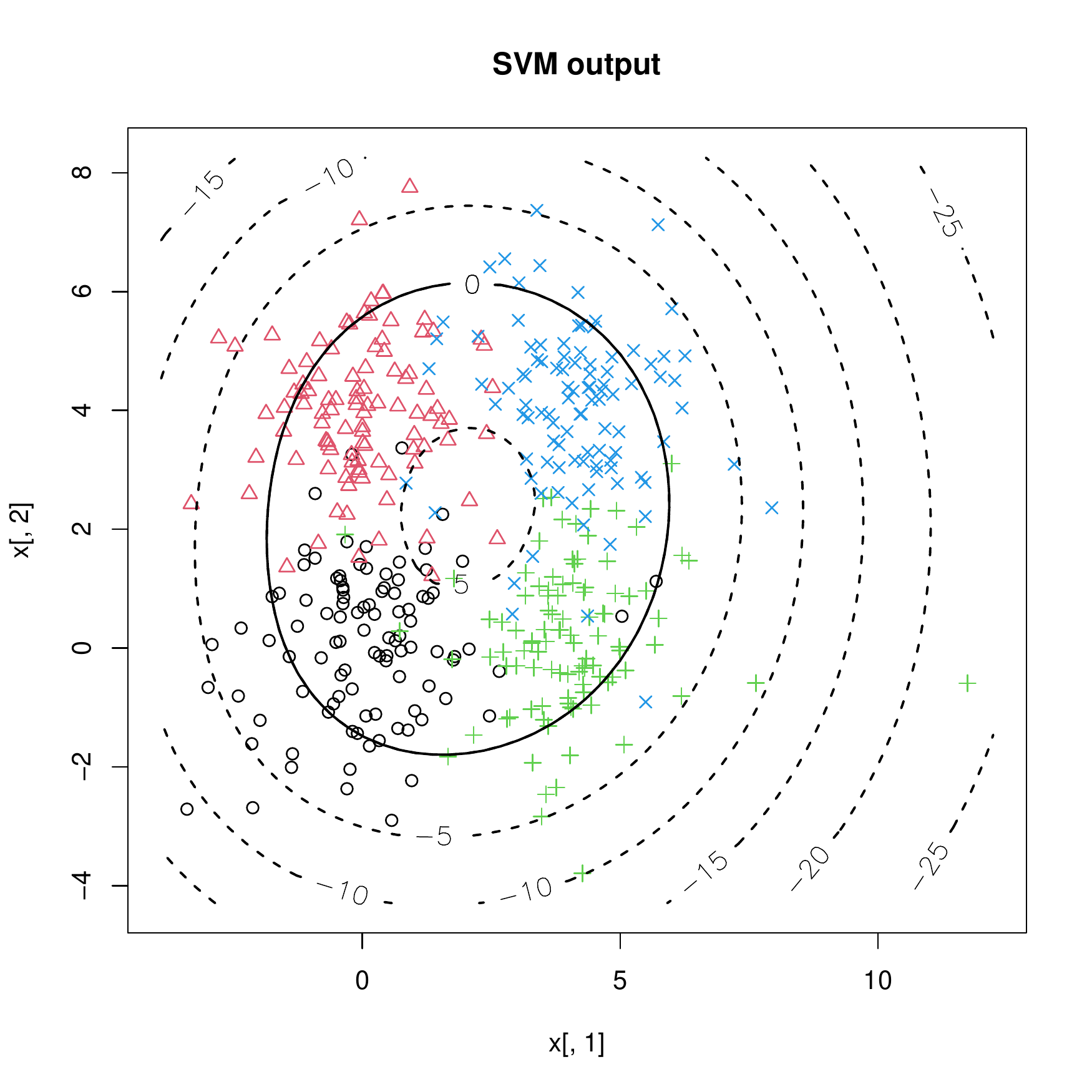}}
\subfloat[\label{fig:fourclass_emptyset}]{\includegraphics[width=0.4\textwidth]{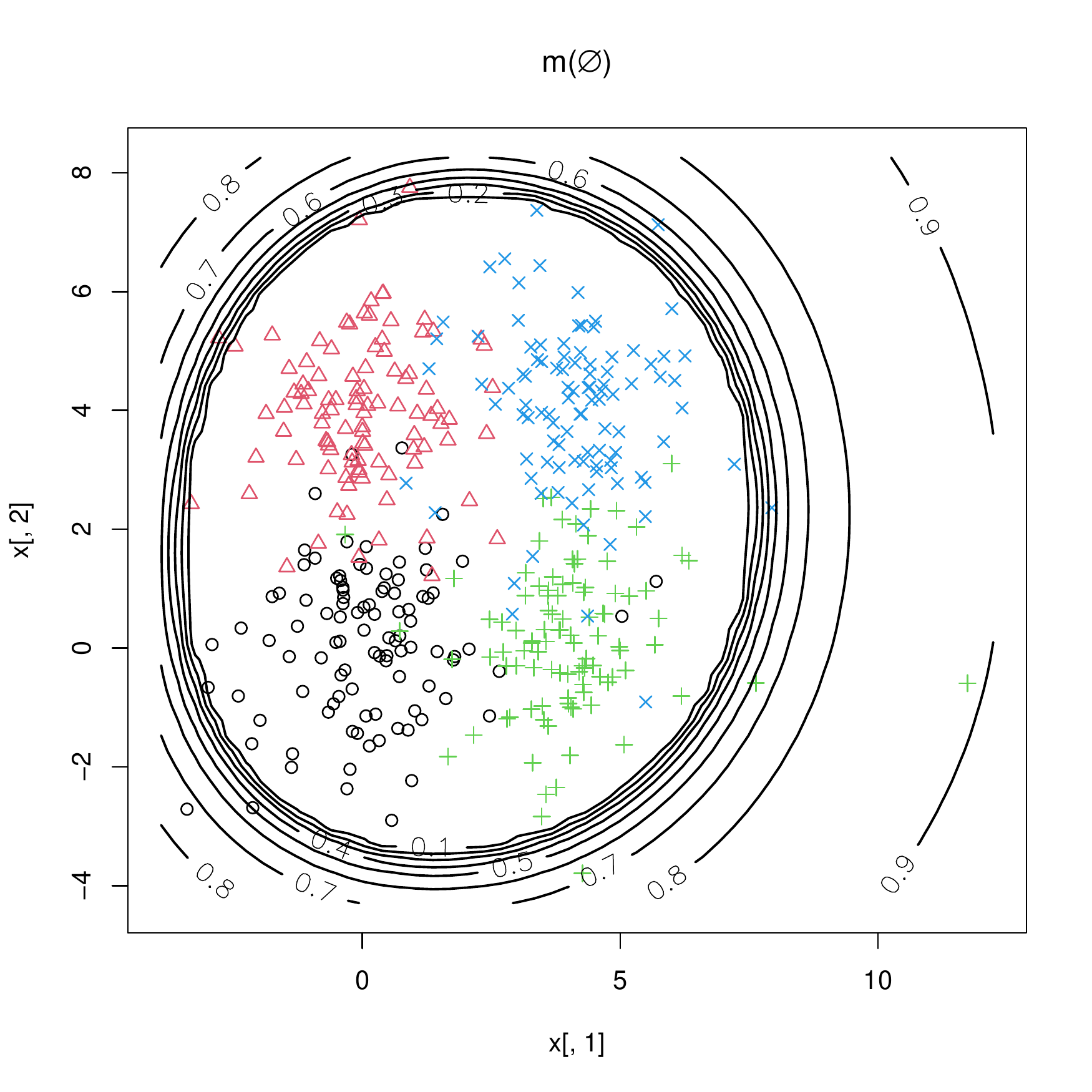}}
\caption{Output of the one-class SVM  (a) and mass on the empty set (b) for the \textsf{fourclass} dataset. \label{fig:fourclass_svm}}
\end{figure}

For the NN part, we set $n_H=20$ and $\lambda=0$. The focal sets were restricted to the  subsets of cardinality less than or equal to 2, and the frame $\Omega$. The network was trained in batch mode using a gradient-based procedure  quite similar to the method described in \cite{silva90}. We minimized loss function \eqref{eq:lossEVCLUS4} with $p=100$. We started the algorithm from five independent random conditions and we kept the best result. The learning curve is shown in Figure \ref{fig:learning_curve_fourclass} with the loss of EVCLUS as a comparison, and the Shepard diagram is displayed in Figure \ref{fig:shepard_NN_fourclass}. Comparing Figures \ref{fig:shepard_fourclass} and \ref{fig:shepard_NN_fourclass}, we can see that the quality of the approximation is similar for EVCLUS and NN-EVCLUS. The former algorithm reaches a loss of $4.75\times 10^{-3}$, while the latter yields $5.64\times 10^{-3}$. 

The obtained evidential partition,  shown in Figure \ref{fig:fourclass_credpart_NN}, is similar to that obtained by EVCLUS, which is displayed in Figure \ref{fig:fourclass_credpart}. However, NN-EVCLUS also allows us to predict the cluster membership of new objects.  Figure \ref{fig:fourclass_credpart_test} shows the predicted evidential  partition of a data set of 1000 vectors drawn from the same distribution as \textsf{fourclass} (a mixture of four multidimensional Student distributions). We can see that the four clusters and the outliers are correctly identified. Figure \ref{fig:NN_fourclass_contour} shows contours of the masses assigned to the singletons  (Figures \ref{fig:fourclass_omega1}-\ref{fig:fourclass_omega4}) and some pairs of clusters (Figures \ref{fig:fourclass_omega13}-\ref{fig:fourclass_omega24}) as functions of $\bx$, and Figure \ref{fig:NN_fourclass_contour_pl} displays contour plots of the plausibility of each of the four clusters as functions of $\bx$. A contour plot of the mass on the empty set is shown in Figure \ref{fig:fourclass_emptyset}.

Figure \ref{fig:Loss_nH_Lamda} shows the influence on the training and test performance of the number $n_H$ of hidden units (Figure \ref{fig:Loss_nH}) and of the regularization coefficient $\lambda$ (Figure \ref{fig:Loss_lambda}) with $n_H=50$. The test loss was computed using a dataset represented in Figure \ref{fig:fourclass_credpart_test}. As expected, the training error decreases slowly with $n_H$ while the generalization reaches a plateau at $n_H=45$. Similarly, the training error increases with $\lambda$ for fixed $n_H$, but the generalization error reaches a minimum for $\lambda=10^{-5}$.

\begin{figure}
\centering  
\subfloat[\label{fig:learning_curve_fourclass}]{\includegraphics[width=0.35\textwidth]{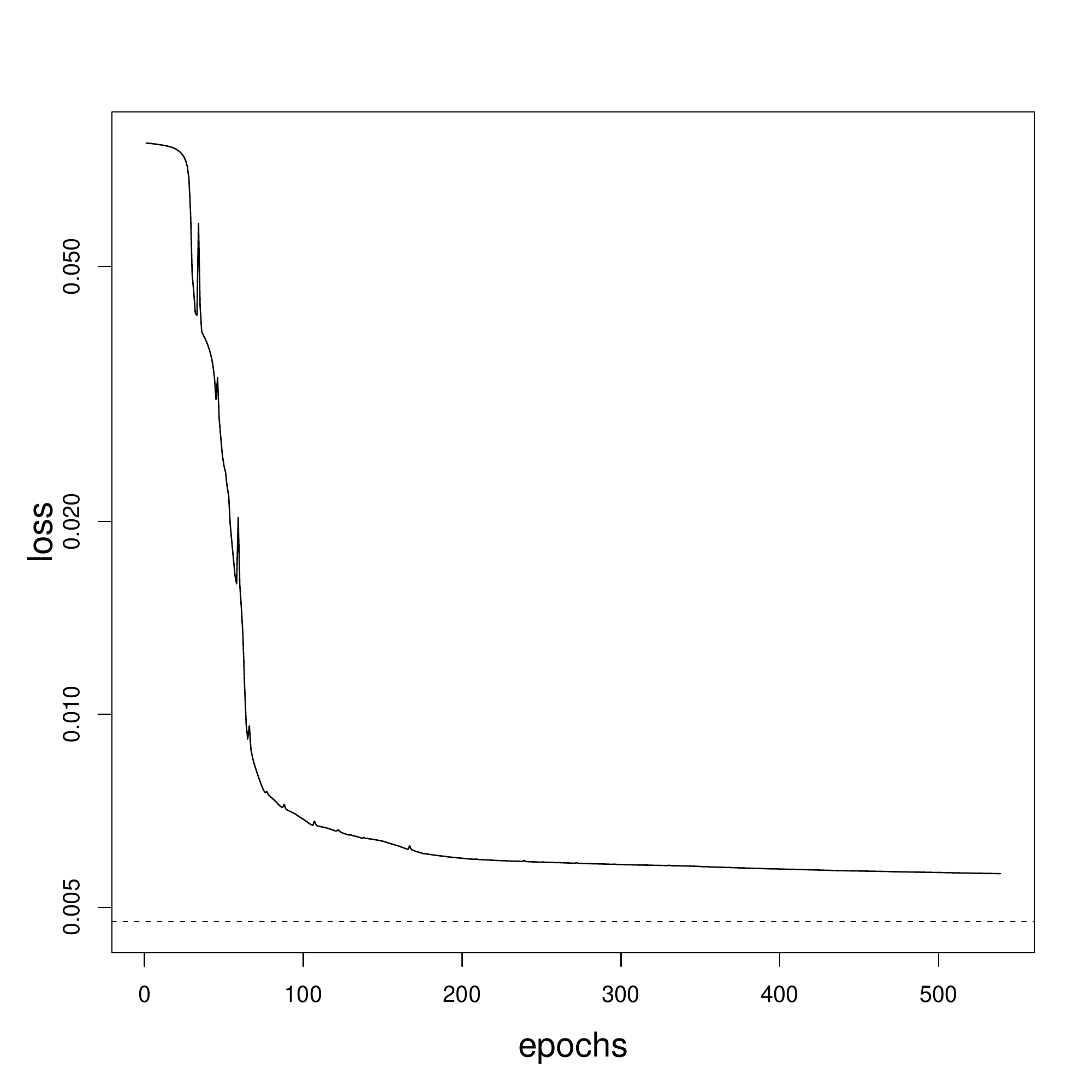}}
\subfloat[\label{fig:shepard_NN_fourclass}]{\includegraphics[width=0.35\textwidth]{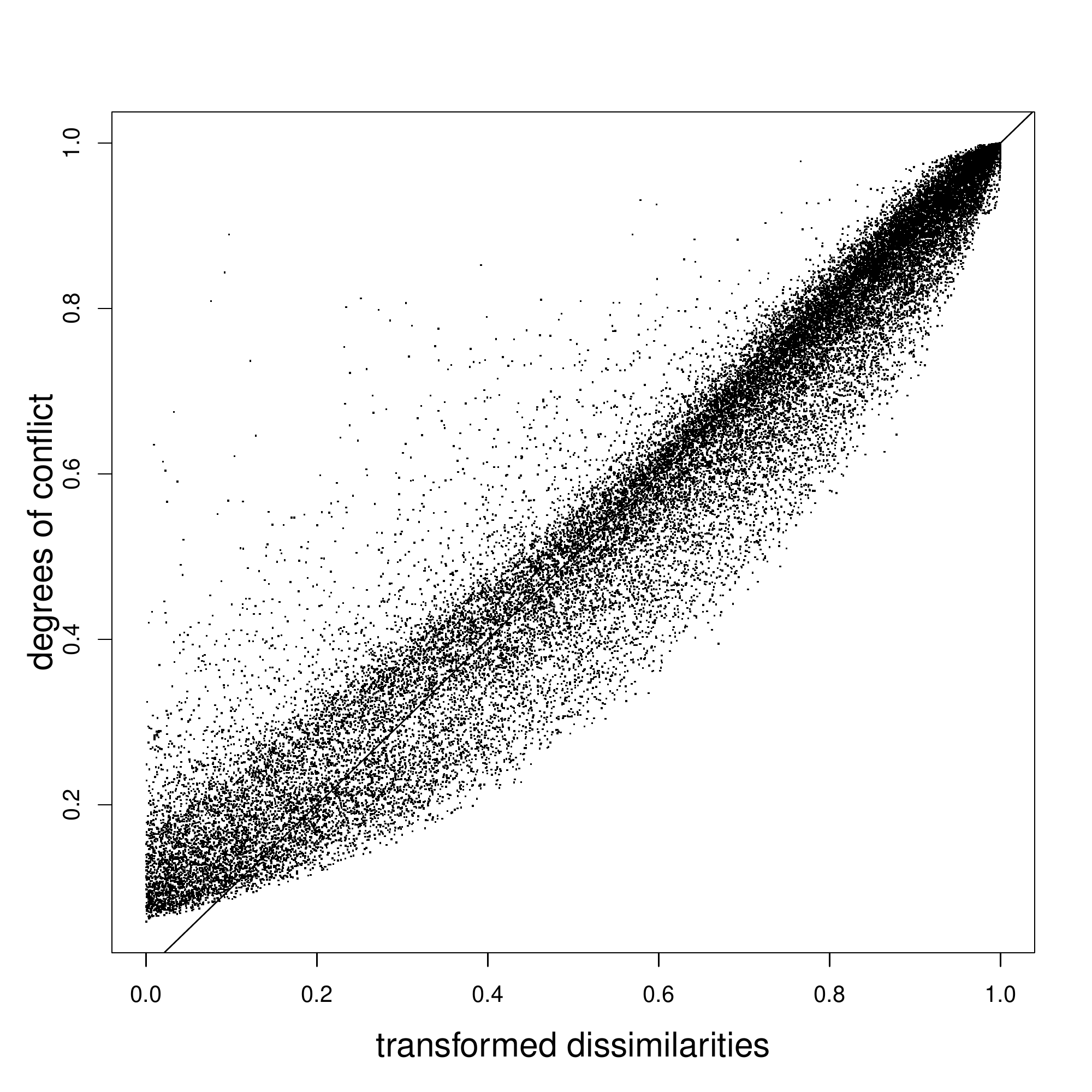}}
\caption{(a): Learning curve  of a NN trained in batch mode on the \textsf{fourclass} dataset (solid line), and value of the loss reached by EVCLUS (horizontal broken line); (b):  plot of the degrees of conflict $\kappa_{ij}$ vs. the transformed dissimilarities $\delta^*_{ij}$.   \label{fig:NN_fourclass}}
\end{figure}

\begin{figure}
\centering  
\includegraphics[width=0.9\textwidth]{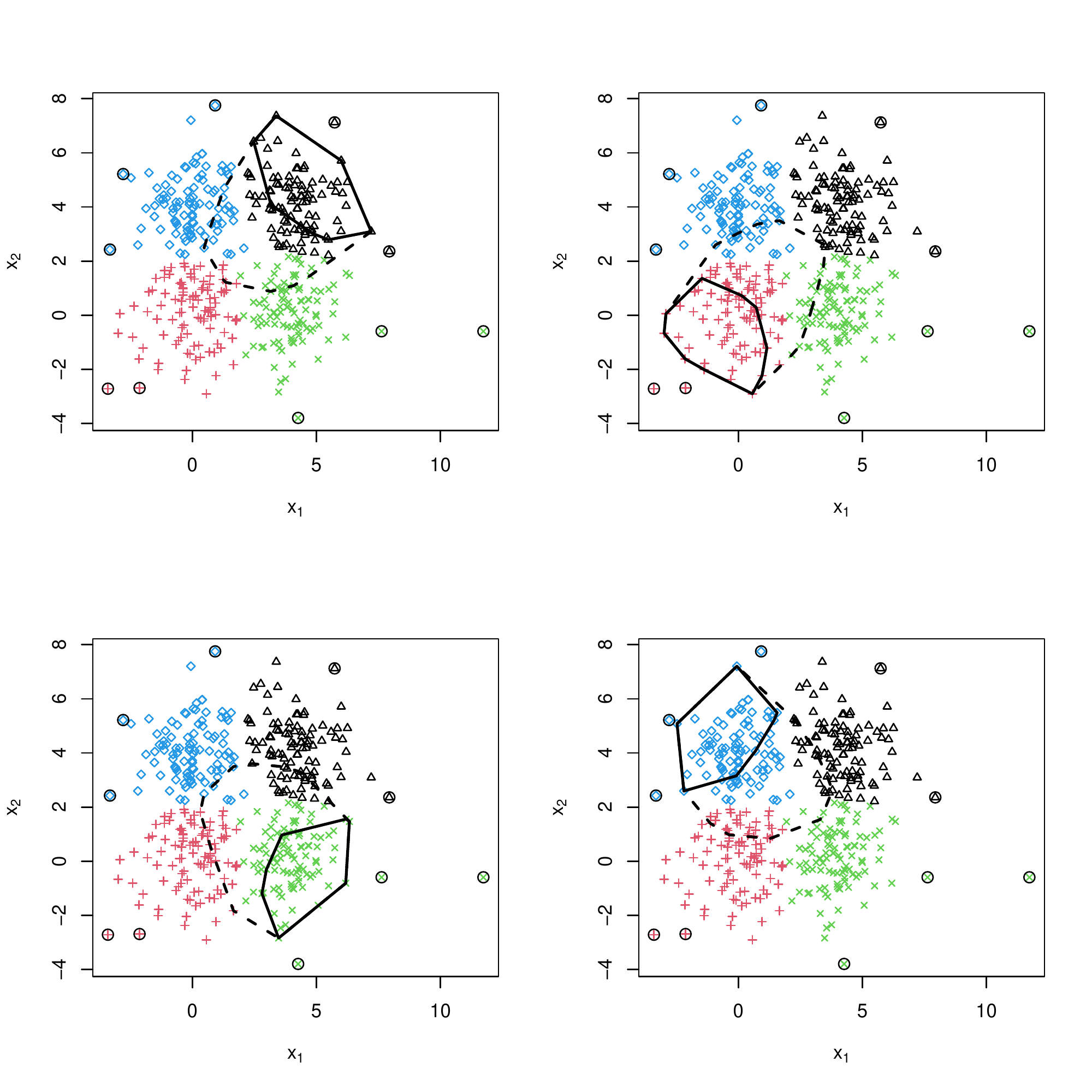}
\caption{Lower and upper approximations of the four clusters for  the \textsf{fourclass} dataset found by NN-EVCLUS. The true classes are displayed with different colors. The identified clusters  are plotted with different symbols. The convex hulls of the cluster lower and upper approximations  are displayed using solid and interrupted lines, respectively. The  outliers are indicated by circles.   \label{fig:fourclass_credpart_NN}}
\end{figure}

\begin{figure}
\centering  
\includegraphics[width=0.9\textwidth]{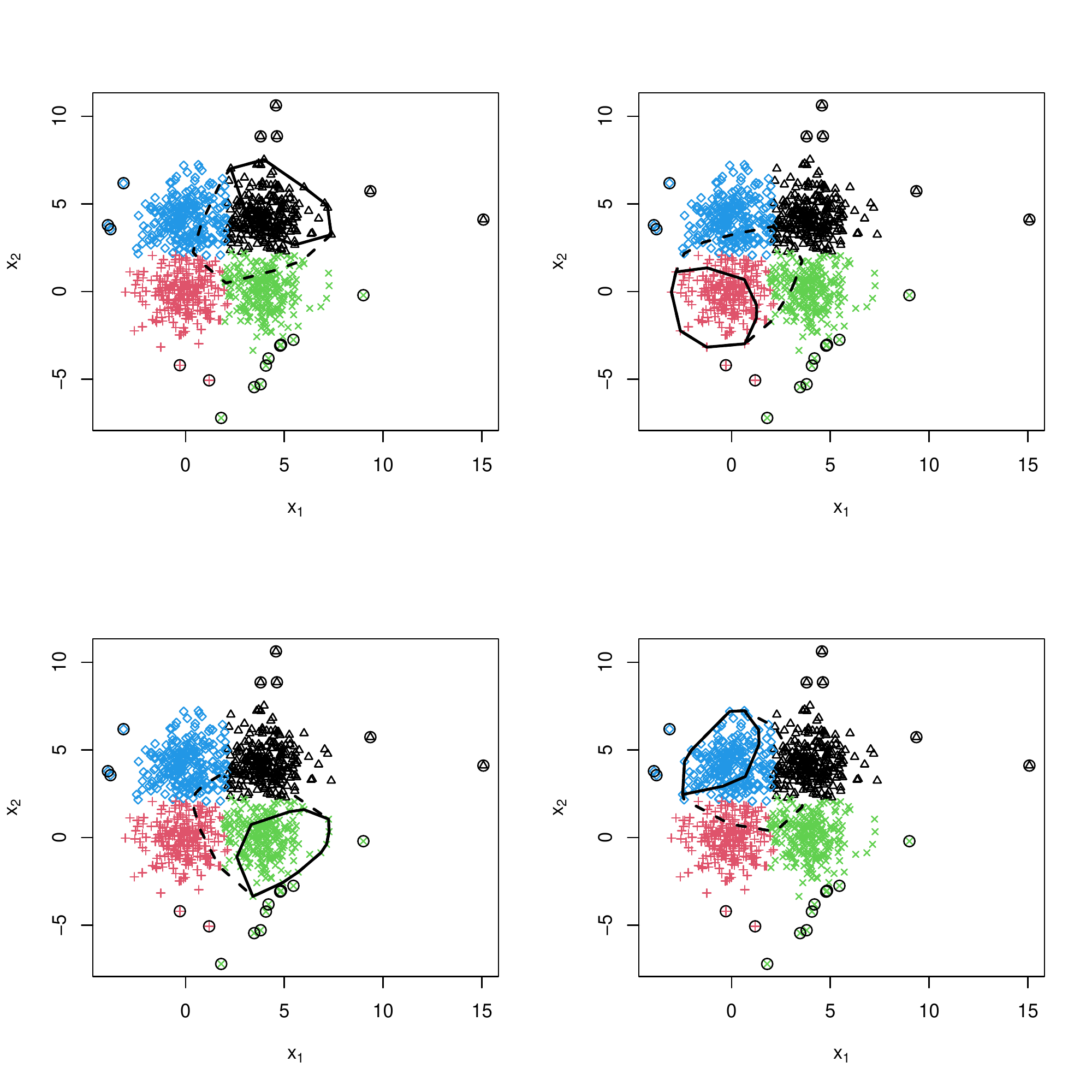}
\caption{Lower and upper approximations of the four clusters for  a test dataset of size 1000 drawn from the same distribution as the \textsf{fourclass} dataset. The true classes are displayed with different colors. The identified clusters  are plotted with different symbols. The convex hulls of the cluster lower and upper approximations  are displayed using solid and interrupted lines, respectively. The  outliers are indicated by circles.   \label{fig:fourclass_credpart_test}}
\end{figure}

\begin{figure}
\centering  
\subfloat[\label{fig:fourclass_omega1}]{\includegraphics[width=0.25\textwidth]{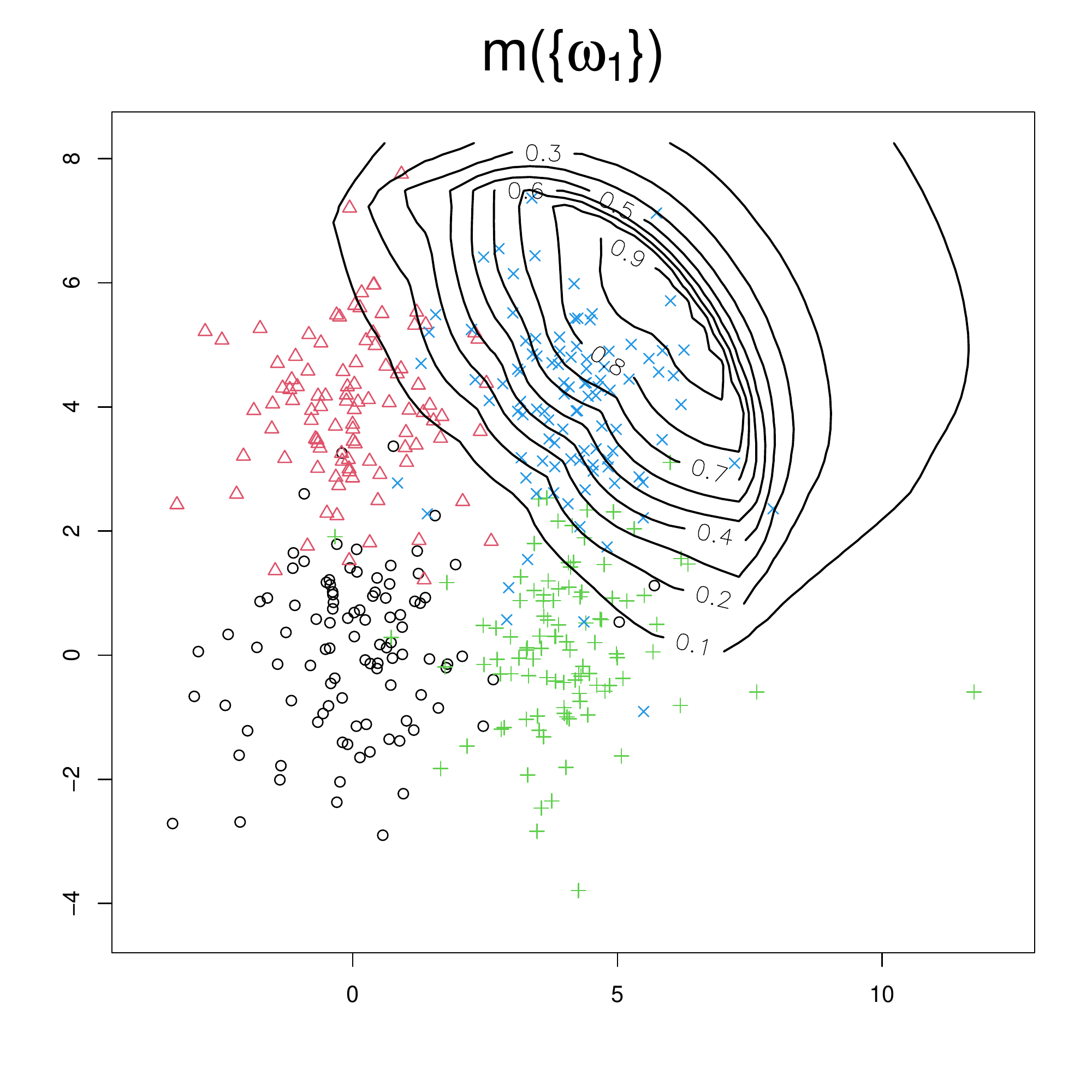}}
\subfloat[\label{fig:fourclass_omega2}]{\includegraphics[width=0.25\textwidth]{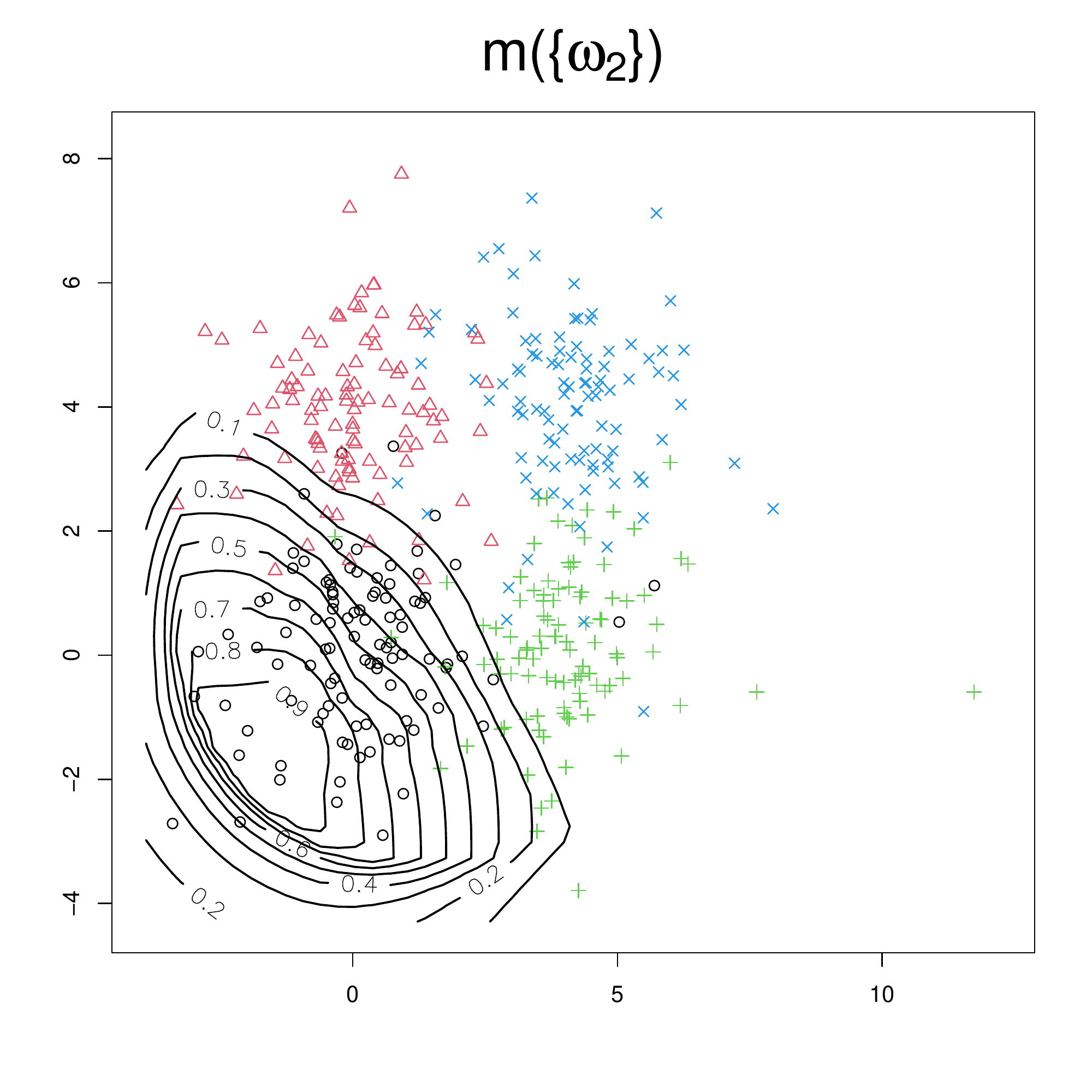}}
\subfloat[\label{fig:fourclass_omega3}]{\includegraphics[width=0.25\textwidth]{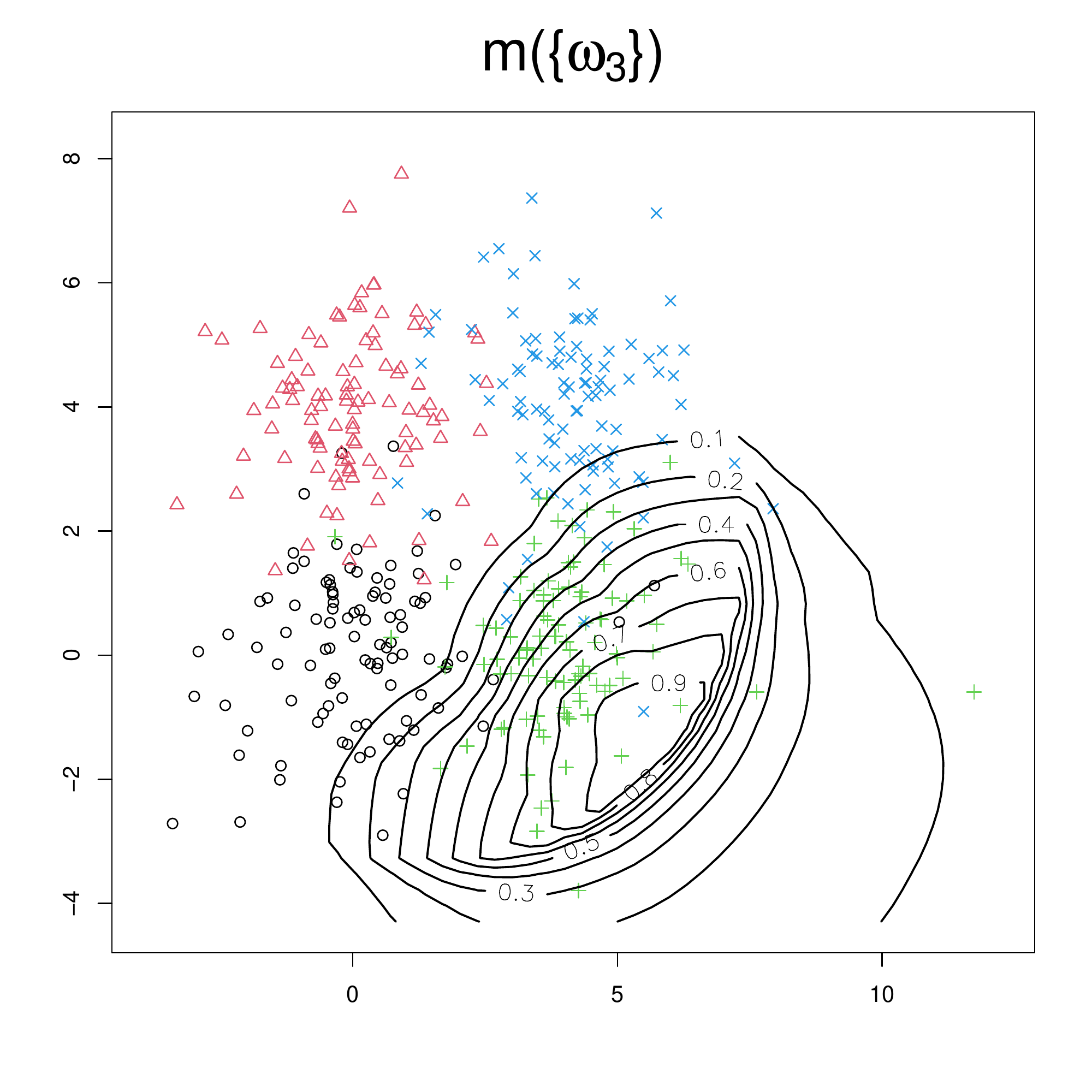}}
\subfloat[\label{fig:fourclass_omega4}]{\includegraphics[width=0.25\textwidth]{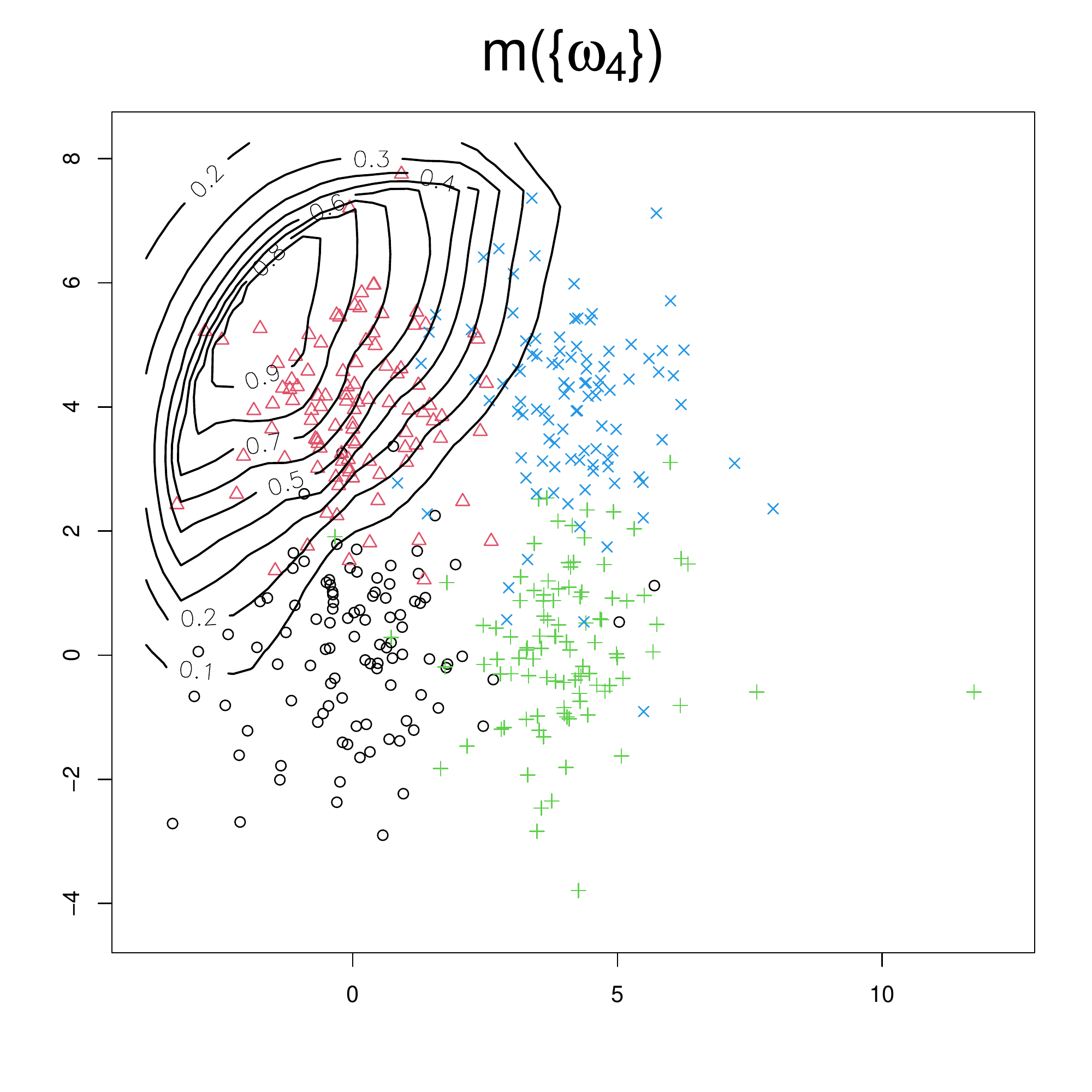}}\\
\subfloat[\label{fig:fourclass_omega13}]{\includegraphics[width=0.25\textwidth]{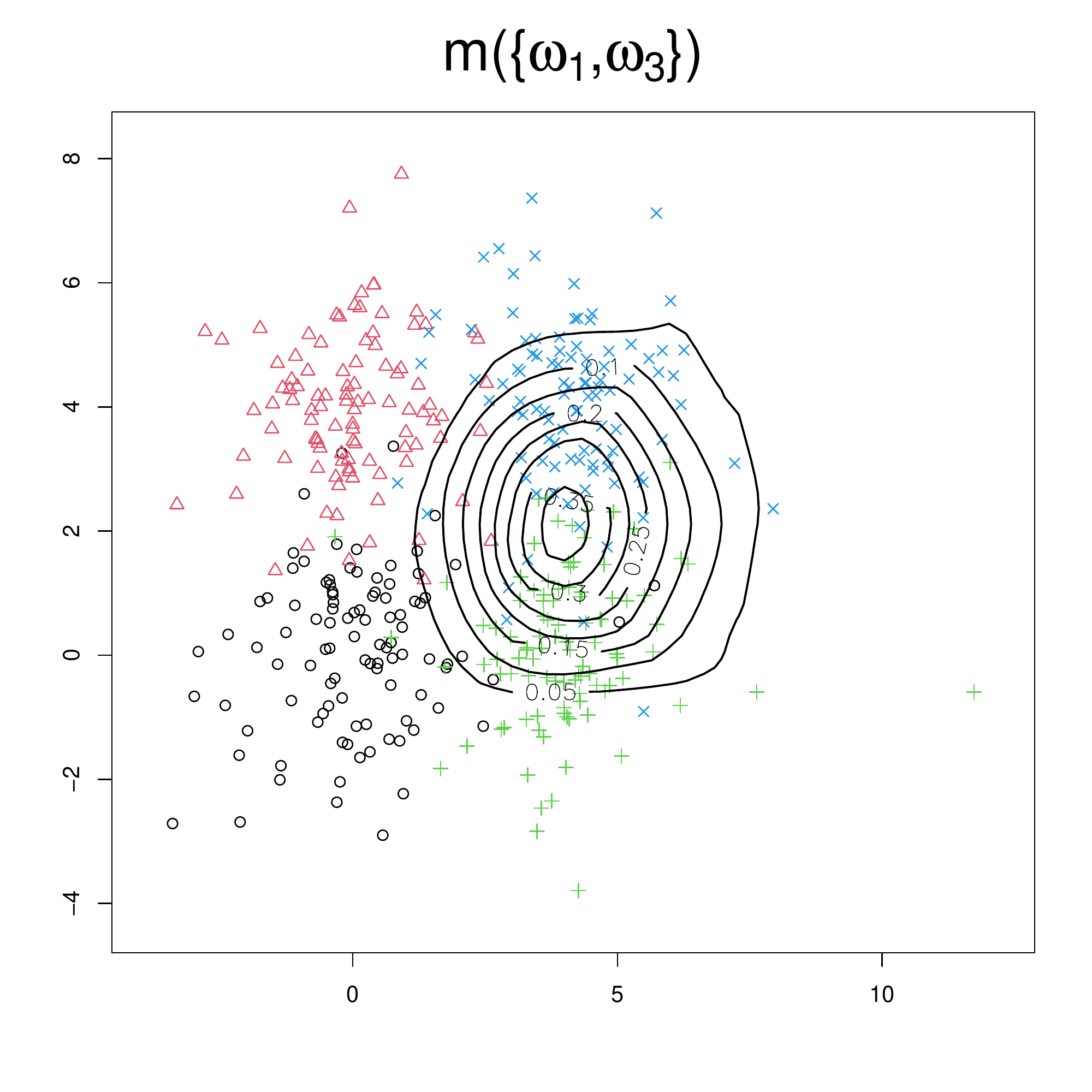}}
\subfloat[\label{fig:fourclass_omega14}]{\includegraphics[width=0.25\textwidth]{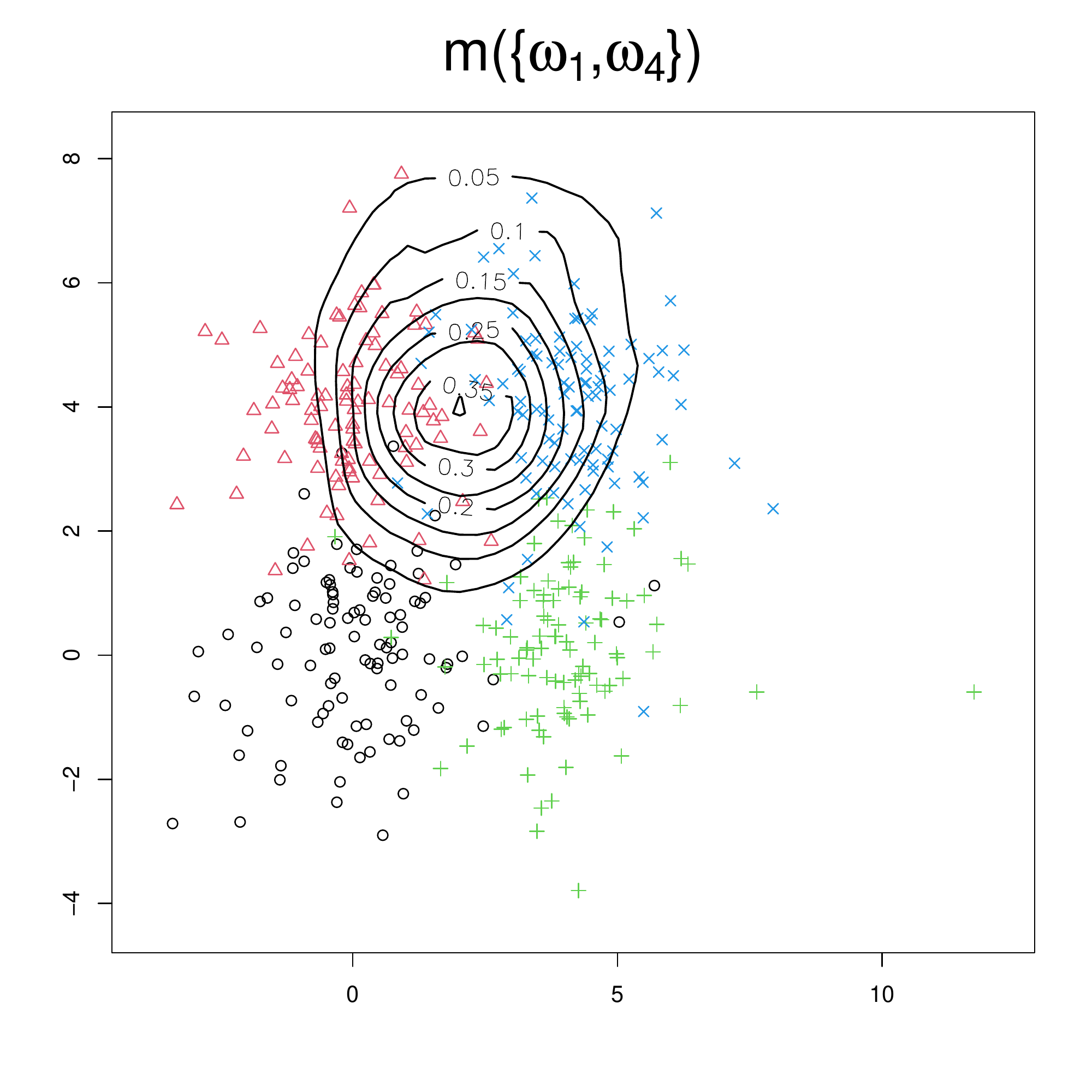}}
\subfloat[\label{fig:fourclass_omega23}]{\includegraphics[width=0.25\textwidth]{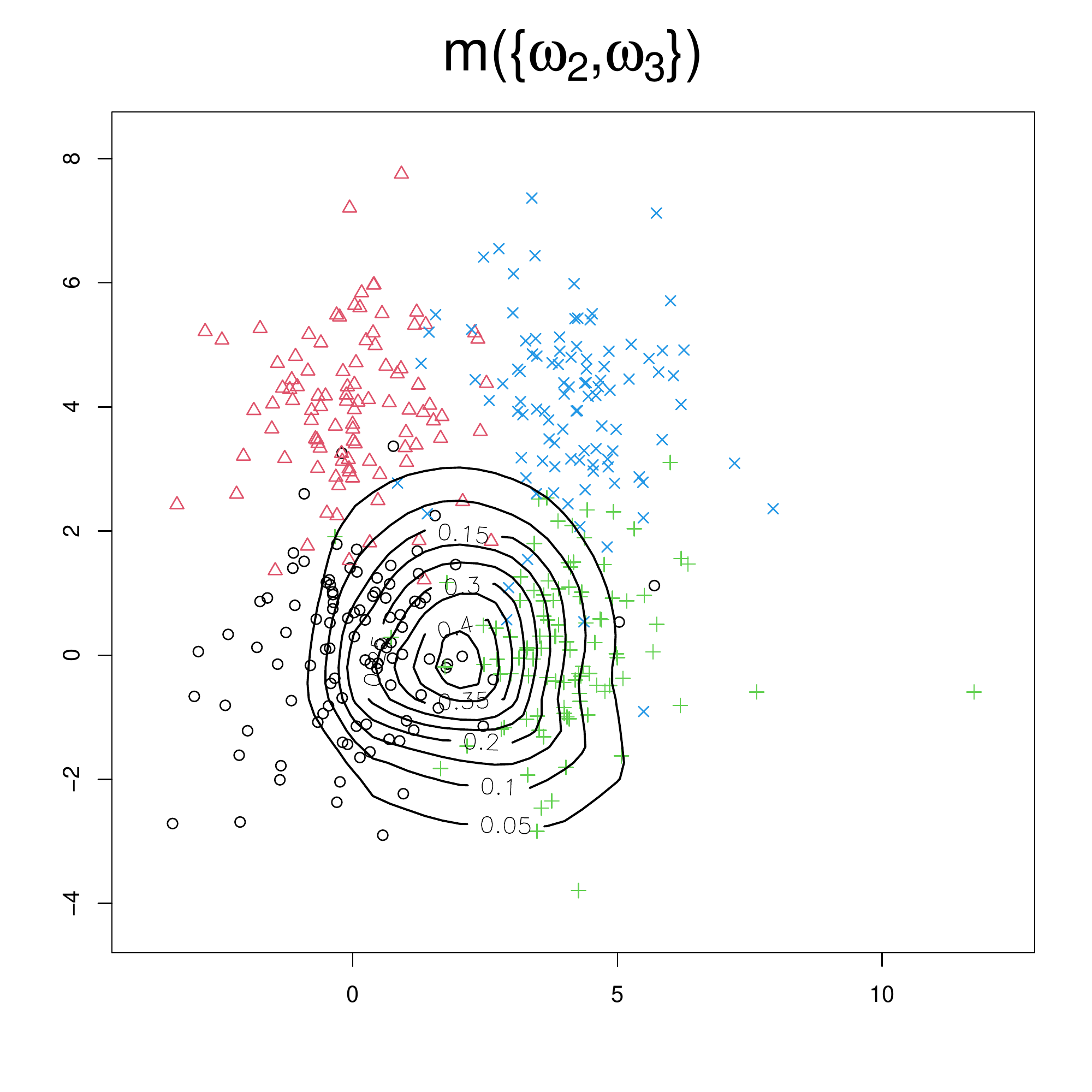}}
\subfloat[\label{fig:fourclass_omega24}]{\includegraphics[width=0.25\textwidth]{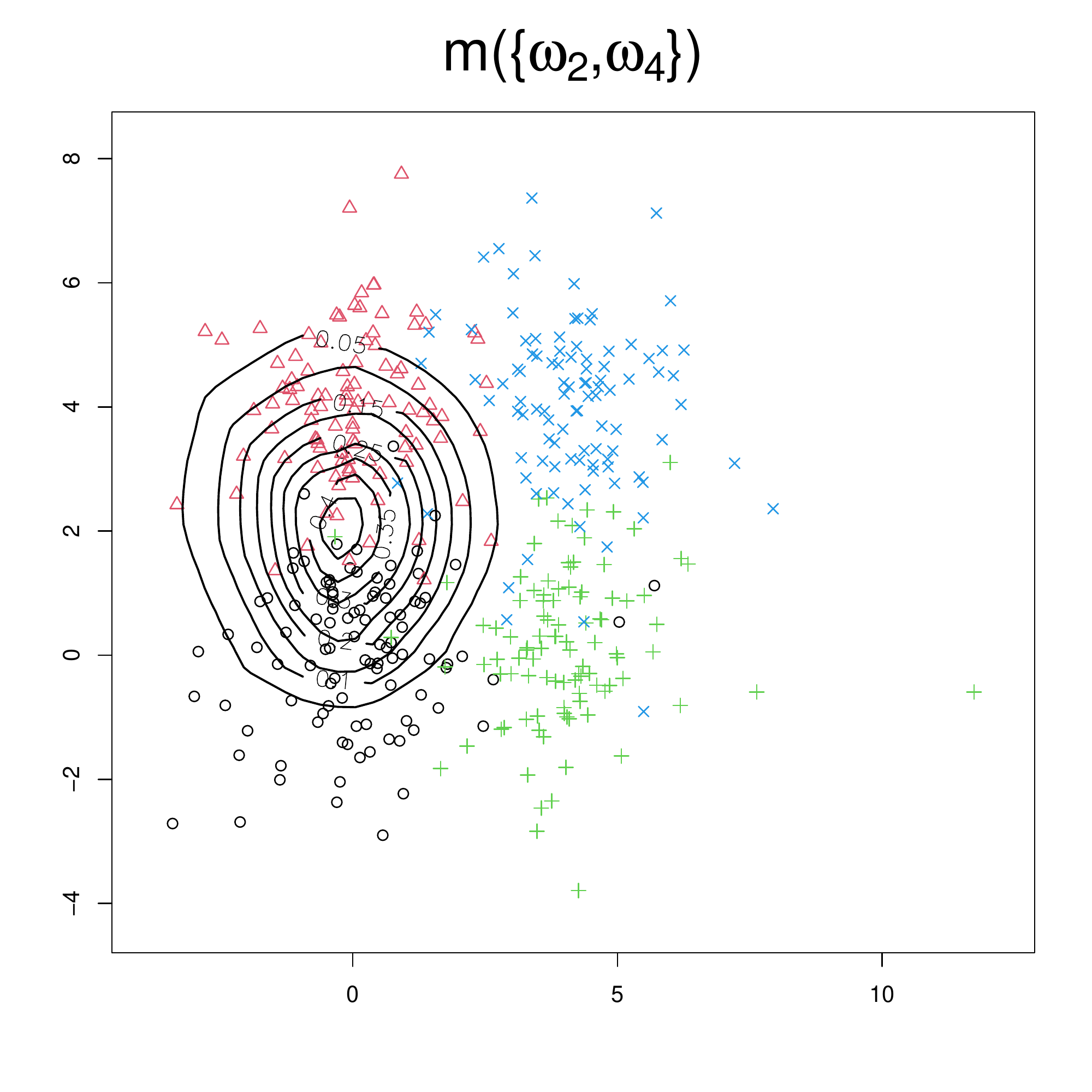}}
\caption{Contour plot of  the masses assigned to singletons (a-d) and pairs of contiguous clusters (e-h) by NN-EVCLUS for the \textsf{fourclass} dataset.  \label{fig:NN_fourclass_contour}}
\end{figure}

\begin{figure}
\centering
\subfloat[\label{fig:fourclass_pl1}]{\includegraphics[width=0.4\textwidth]{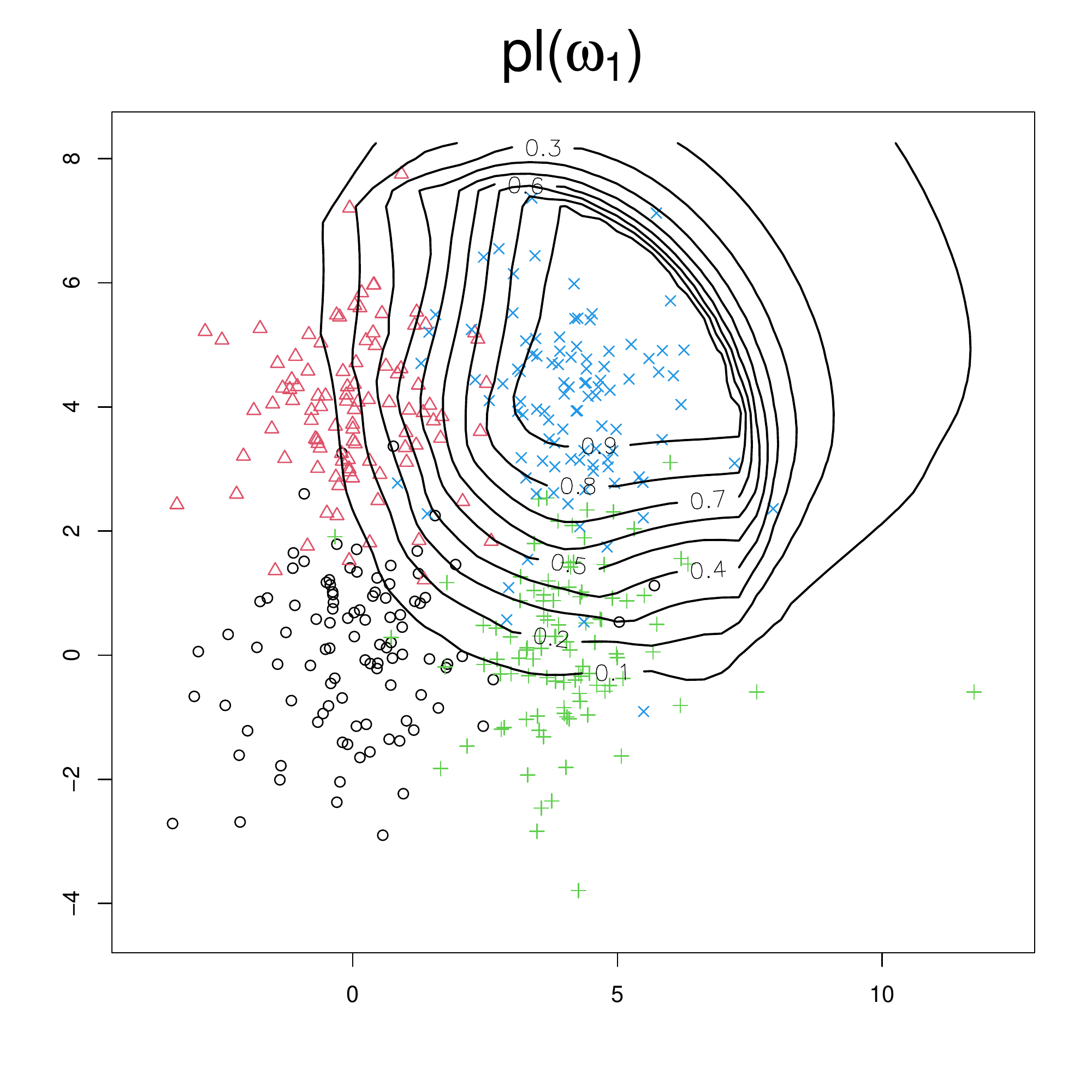}}
\subfloat[\label{fig:fourclass_pl2}]{\includegraphics[width=0.4\textwidth]{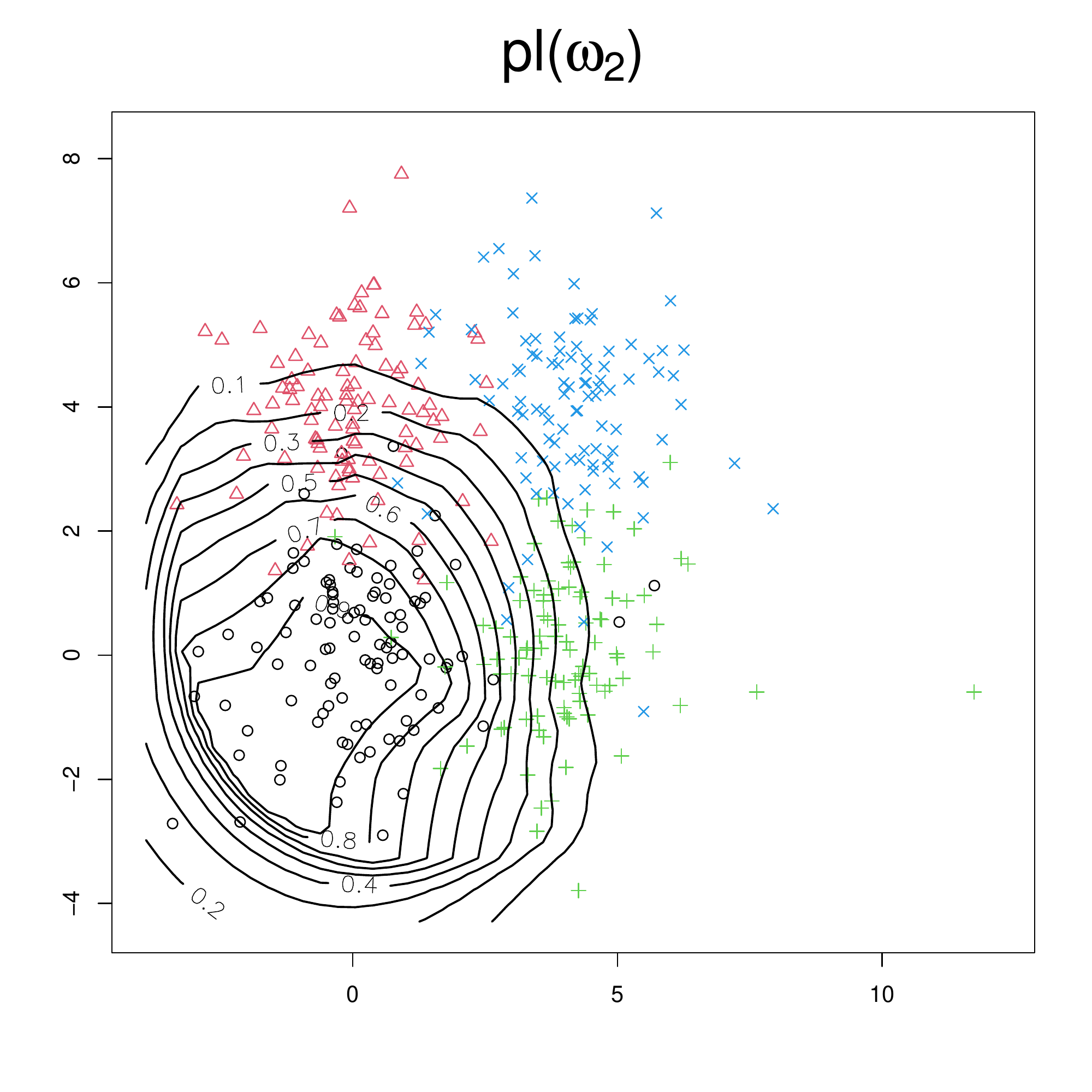}}\\
\subfloat[\label{fig:fourclass_pl3}]{\includegraphics[width=0.4\textwidth]{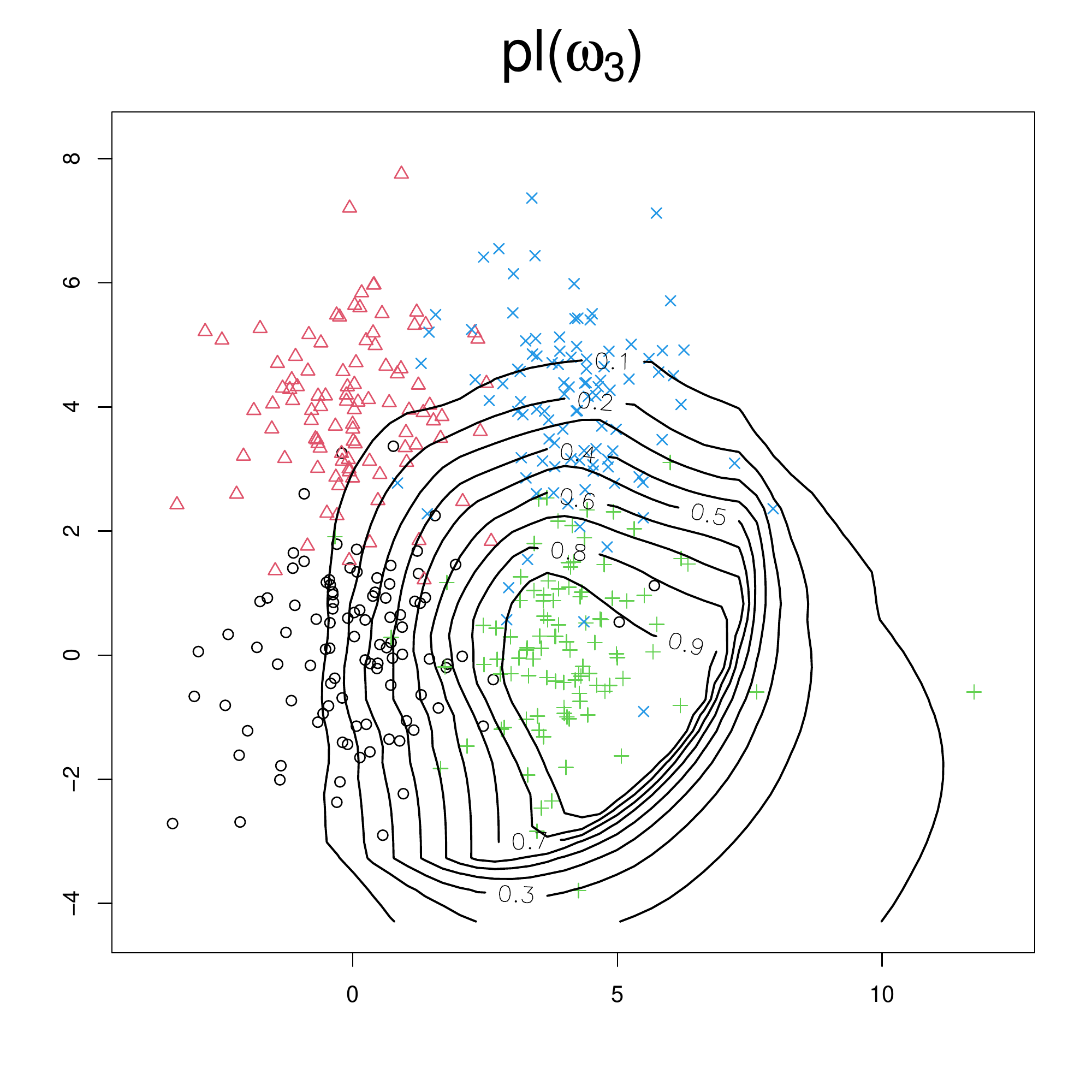}}
\subfloat[\label{fig:fourclass_pl4}]{\includegraphics[width=0.4\textwidth]{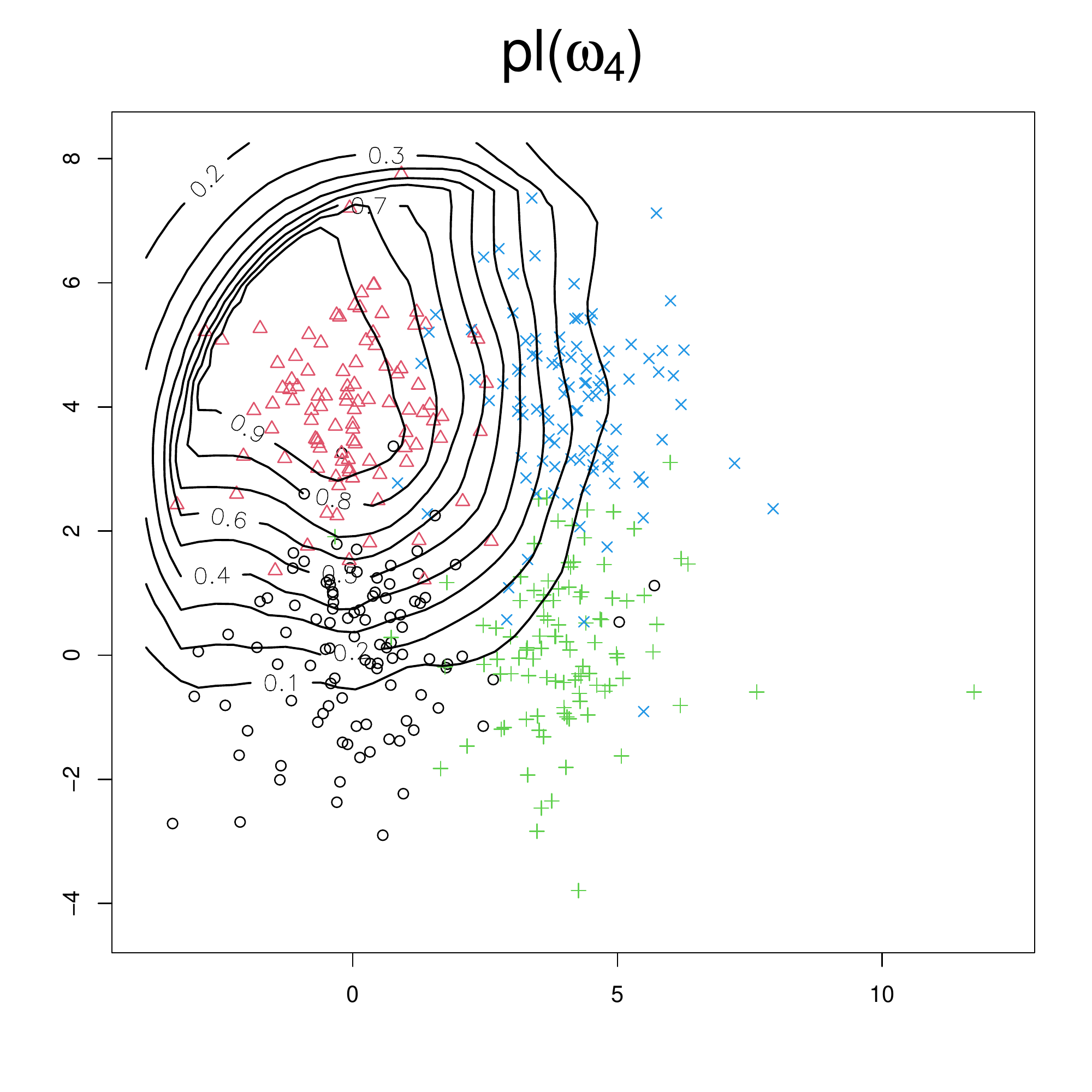}}
\caption{Contour plot of  plausibilities of each of the four clusters obtained by NN-EVCLUS for the \textsf{fourclass} dataset.  \label{fig:NN_fourclass_contour_pl}}
\end{figure}

\begin{figure}
\centering  
\subfloat[\label{fig:Loss_nH}]{\includegraphics[width=0.35\textwidth]{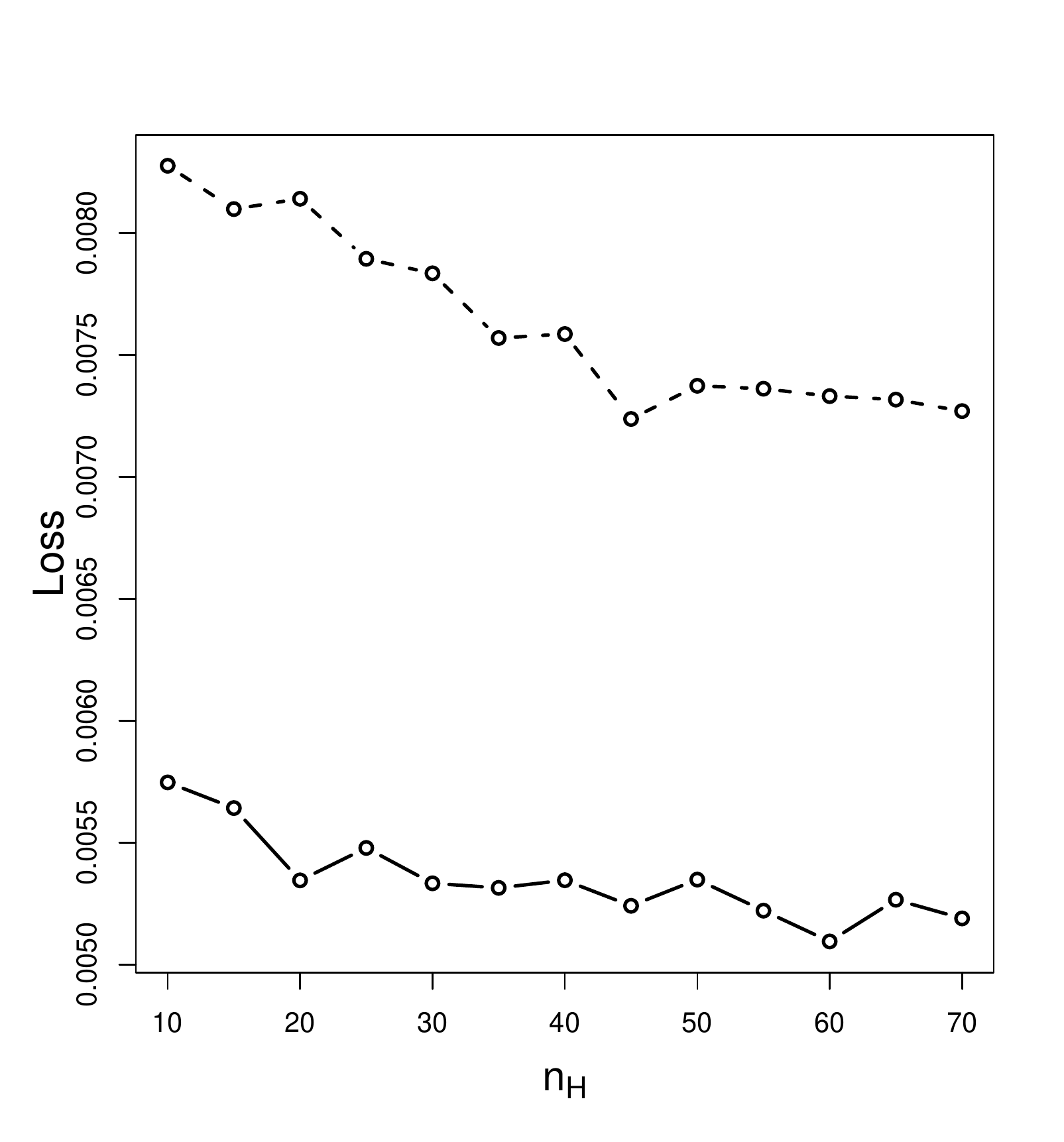}}
\subfloat[\label{fig:Loss_lambda}]{\includegraphics[width=0.35\textwidth]{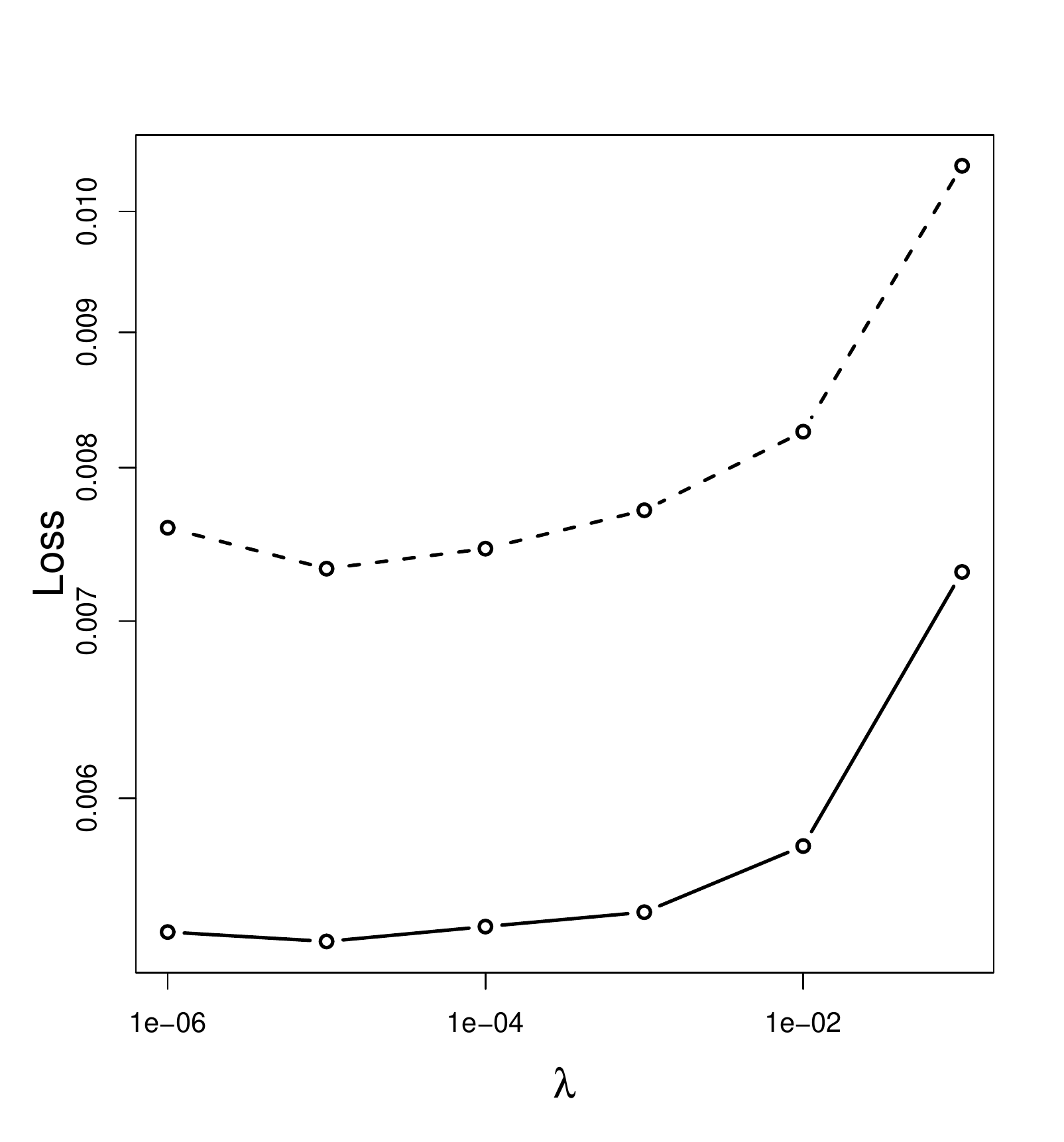}}
\caption{Training (solid lines) and test (broken lines) loss of NN-EVCLUS on the \textsf{fourclass} dataset vs. number $n_H$ of hidden units (a) and vs. regularization coefficient $\lambda$ (b) with 50 hidden units. \label{fig:Loss_nH_Lamda}}
\end{figure}

\end{Ex}

\subsection{Using side information}
\label{subsec:semi}

In many cases,  additional knowledge about some objects can guide the learning process. In clustering, such knowledge may take the form of pairwise constraints specifying that some objects belong to the same class (\emph{must-link} constraints), or belong to different classes (\emph{cannot-link} constraints). In evidential clustering, a variant of ECM (called CECM) dealing with such constraints was first introduced in \cite{antoine12}. The constrained version of EVCLUS (called CEVCLUS) was then introduced in \cite{antoine14}, and improved in   \cite{li18}. 

CEVCLUS minimizes a loss function that is the sum of the squared error loss and a penalization term defined as follows. Let $S_{ij}$ denote the event that objects $i$ and $j$ belong to the same cluster, and  $\overline{S}_{ij}$  the complementary event. Given mass functions $m_i$ and $m_j$ about the cluster membership of objects $i$ and $j$, the plausibility of $S_{ij}$ and $\overline{S}_{ij}$ can be computed as follows \cite{antoine12}:
\begin{subequations}
\begin{align}
\label{eq:PlS}
Pl_{ij}(S_{ij}) & =   1-\kappa_{ij}\\
\label{eq:PlnegS}
Pl_{ij}(\overline{S}_{ij}) & =   1-m_i(\emptyset)-m_j(\emptyset)+m_i(\emptyset)m_j(\emptyset)-\sum_{k=1}^c m_i(\{\omega_k\})m_j(\{\omega_k\}).
\end{align}
\end{subequations}
\new{We note that the pair $(Pl_{ij}(S_{ij}),Pl_{ij}(\overline{S}_{ij}))$ can  take any value in $[0,1]^2$; for instance, if $m_i(\emptyset)=m_j(\emptyset)=1$, then $Pl_{ij}(S_{ij})=Pl_{ij}(\overline{S}_{ij})=0$, and if $m_i(\Omega)=m_j(\Omega)=1$, then $Pl_{ij}(S_{ij})=Pl_{ij}(\overline{S}_{ij})=1$.} For objects $i$ and $j$  known to belong to the same cluster, $Pl_{ij}(S_{ij})$ should be high and $Pl_{ij}(\overline{S}_{ij})$ should be low, and the converse holds  if objects $i$ and $j$ are known to belong to different clusters. The loss function of CEVCLUS is thus defined as
\begin{equation}
\label{eq:stress}
  \calL_c(\calM)= \calL(\calM) + \frac{\xi}{2(|\mathsf{ML}|+|\mathsf{CL}|)}(\calP_\mathsf{ML}+\calP_\mathsf{CL}),
\end{equation}
with
\begin{subequations}
\label{eq:JMLCL}
\begin{align}
\calP_\mathsf{ML}&=\sum_{(i,j)\in\mathsf{ML}} \left(Pl_{ij}(\overline{S}_{ij})+1-Pl_{ij}(S_{ij}) \right),\\
\calP_\mathsf{CL}&=\sum_{(i,j)\in\mathsf{CL}} \left(Pl_{ij}(S_{ij})+1-Pl_{ij}(\overline{S}_{ij})\right),
\end{align}
\end{subequations}
where $\calL(\calM)$ is the squared error loss defined by \eqref{eq:lossEVCLUS} or \eqref{eq:lossEVCLUS1}, $\xi$ is a hyperparameter that controls the trade-off between the stress and the constraints, and $\mathsf{ML}$ and $\mathsf{CL}$ are the sets of must-link and cannot-link constraints, respectively. The constrained loss \eqref{eq:stress}-\eqref{eq:JMLCL} was minimized by gradient descent in the original version of CEVCLUS \cite{antoine14}; a more efficient cyclic coordinate descent algorithm was proposed in \cite{li18}.

Another form of prior information may come as a subset of labeled objects. This is the point of view of semi-supervised learning \cite{chapelle06}.  This approach is relevant when classes are already defined, but only a subset of objects can be labeled because of time or cost constraints. Given a subset of labeled data, it is possible to derive must-link and cannot-link contraints, but the converse is false in general: the pairwise constraint formalism is thus more general. Semi-supervised learning is rather seen as an extension of supervised classification, whereas constrained clustering is seen as an extension of fully unsupervised learning. Both types of side information can easily be exploited by NN-EVCLUS, as will be explained below.

\paragraph{Integration of pairwise constraints} Pairwise constraints can be easily integrated in NN-EVCLUS using a penalized loss such as \eqref{eq:stress}. From \eqref{eq:conflict1}, we have
\[
Pl_{ij}(S_{ij}) = 1-\bm_i^{*T} \bC \bm_j^*= \bm_1^{*T}( \bone\bone^T-\bC)\bm_j^*,
\]
where $\bone$ is a column vector of length $f$ whose components are all equal to 1. Similarly, we can write 
\[
Pl_{ij}(\overline{S}_{ij}) = 1-\bm_i^{*T} \bE \bm_j^*-\bm_i^{*T} \bS \bm_j^*= \bm_1^{*T}( \bone\bone^T-\bE-\bS)\bm_j^*,
\]
where $\bE$ is a square matrix of size $f$ defined as
\begin{equation*}
\label{eq:B}
\bE=\begin{pmatrix}
               1 & 1 & \cdots & 1 \\
               1 & 0 & \cdots & 0 \\
               \vdots & \vdots & \ddots & \vdots \\
               1 & 0 & \cdots & 0
                     \end{pmatrix}
\end{equation*}
and  $\bS$ is the square matrix of size $f$ with general term
\begin{equation*}
\label{eq:A}
(\bS)_{k\ell}=\begin{cases}
                   1 & k=\ell,\ |F_k|=1 \\
                   0 & \text{otherwise}.
                 \end{cases}
 \end{equation*}
 Consequently, we can rewrite \eqref{eq:JMLCL} as
 \begin{subequations}
\label{eq:JMLCL1}
\begin{align}
\calP_\mathsf{ML}&=\sum_{(i,j)\in\mathsf{ML}} \bm_i^{*T} \bQ \bm_j^*\\
\calP_\mathsf{CL}&=\sum_{(i,j)\in\mathsf{CL}} \left(2-\bm_i^{*T} \bQ \bm_j^*\right),
\end{align}
\end{subequations}
with $\bQ=\bone\bone^T+\bC-\bE-\bS$. The gradients of $\calP_\mathsf{ML}$ and $\calP_\mathsf{CL}$ with respect to the network parameters are given in  \ref{sec:pairwise_grad}.

\paragraph{Integration of labeled data} Let us assume that we have $n_s$ labeled attribute vectors $\{(\bx_i,y_i), i\in \calI_s\}$, where $\calI_s\subseteq \{1,\ldots,n\}$, and $y_i\in\Omega$ is the class label of object $i$. We can use this information by minimizing a penalized loss of the form
\begin{equation}
\label{eq:loss_semi}
\calL_{S}(\btheta)=(1-\nu) \calL(\btheta) + \nu  \calP_s,
\end{equation}
where $\nu$ is a coefficient and $\calP_s$ is a penalization term defined as
\begin{equation}
\label{eq:reg_ss}
\calP_s=\frac{1}{n_s}\sum_{i\in \calI_s} \sum_{l=1}^c (pl^*_{il}-y_{il})^2.
\end{equation}
In \eqref{eq:reg_ss},  $y_{il}=I(y_i=\omega_l)$, $pl^*_{il}=pl_i^*(\omega_l)$,  and $pl^*_i$ is the contour function corresponding to $m^*_i$. The rationale behind \eqref{eq:reg_ss} is that, when object $i$ is known to belong to class $l$, the plausibility of that class should be high, while the plausibility of the other classes should be low. We can notice than, when $n_s=n$, the learning task becomes fully supervised. The gradient of $\calP_s$ with respect to the model parameters is given in \ref{sec:ns_grad}

\paragraph{Metric adaptation}

Although it has been shown that CEVCLUS has the ability to use of pairwise constraints to improve clustering results \cite{antoine14,li18}, it can fail to do so effectively when the constraints are inconsistent with the distance matrix. As recalled in Section \ref{subsec:evclus},  EVCLUS is based on the assumption that similar objects are likely to belong to the same cluster and, conversely, dissimilar objects plausibly belong to different clusters. If pairwise constraints are provided, which force similar objects to belong to different clusters, or dissimilar objects to belong to the same cluster, then the two terms in loss function \eqref{eq:stress} may become strongly inconsistent and CEVCLUS may fail to find a suitable evidential partition. 

An approach to solve this problem is to use the additional information (pairwise constraints or labeled data) to learn a metric such that objects that are known to belong to different clusters become further apart, while objects in a given cluster are as similar as possible. When labeled data are provided, we can extract discriminant features using classical Fisher Discriminant Analysis (FDA) or a nonlinear version such as Local Fisher Discriminant Analysis (FDA) \cite{sugiyama06} or Kernel Fisher Discriminant Analysis (KFDA) \cite{yang04}, and compute the distance matrix in the new feature space. When pairwise constraints are available, we can use feature extraction techniques such as Learning with Side Information (LSI) \cite{xing03}, Distance Metric Learning with Eigenvalue Optimization (DML-eig) \cite{ying12}, Pairwise Constrained Component Analysis (PCCA) or its kernelized version KPCCA  \cite{mignon12}.

\begin{Ex}
The \textsf{circles} dataset shown in Figure \ref{fig:circles_MLCL} is composed of 500 two-dimensional vectors distributed in a spherical cluster surrounded by a circular-shaped cluster. The proportions of the two clusters are, respectively, 2/5 and 3/5. NN-EVCLUS cannot find this partition without additional knowledge, because it violates the fundamental assumption that  two dissimilar objects are unlikely to belong to the same cluster: maximally distant points on the circle at the extremities of a diameter actually belong to the same cluster. We can, however, use additional information in the form of pairwise constrained or labeled data and modify the distance matrix accordingly. 

To illustrate this approach, we randomly generated 50 object pairs, which gave us 22 must-link constraints and 28 cannot-link constraints as shown, respectively, in Figure \ref{fig:circles_ML} and \ref{fig:circles_CL}.  These constraints  represent a tiny fraction of the $500\times 499/2=124750$ object pairs. We used this information to extract a discriminant feature  by KPCCA\cite{mignon12} using a Gaussian kernel with inverse kernel width $\sigma=0.3$. As shown in Figure \ref{fig:circles_PCCA}, the data are linearly separable in this new one-dimensional feature space. The Euclidean distance matrix in the original space and in the transformed space are shown, respectively, in Figure \ref{fig:circles_dist} and \ref{fig:circles_dist_PCCA}. 

We trained NN-EVCLUS with the original features $\bx$ and the Euclidean matrix in the KPCCA feature space, with $n_H=30$ hidden units, $\lambda=0.01$, and the loss function \eqref{eq:stress} with $\xi=0.1$. (The results are not sensitive to $\xi$ in this example, and even setting $\xi=0$, i.e., ignoring the constraints gives good results thanks to the very good separation in the transformed feature space).  The results are shown as contour plots of the masses assigned to each of the four focal sets in Figure \ref{fig:NN_circles_contour}, and as contour plots of the plausibilities of the two classes in Figure \ref{fig:NN_circles_contour_pl}. We can see that a meaningful evidential partition has been found and the two clusters are perfectly separated (as shown by the decision boundary plotted in red in Figure \ref{fig:NN_circles_contour_pl}).

Similar results were obtained with  50 randomly selected labeled instances as shown in Figure \ref{fig:circles_semi},  using KFDA instead of KPCCA, and penalized loss function \eqref{eq:loss_semi}. The distributions of the discriminant feature extracted by KFDA in each of the two classes are shown in Figure \ref{fig:circles_kfda}, and contour plots of the plausibilities of the two classes computed by NN-EVCLUS trained in semi-supervised mode with the  50 labeled data are displayed in Figure \ref{fig:NN_circles_contour_pl_semi}. By comparing Figures  \ref{fig:NN_circles_contour_pl} and  \ref{fig:NN_circles_contour_pl_semi}, we can see that the results are almost identical.

\begin{figure}
\centering  
\subfloat[\label{fig:circles_ML}]{\includegraphics[width=0.35\textwidth]{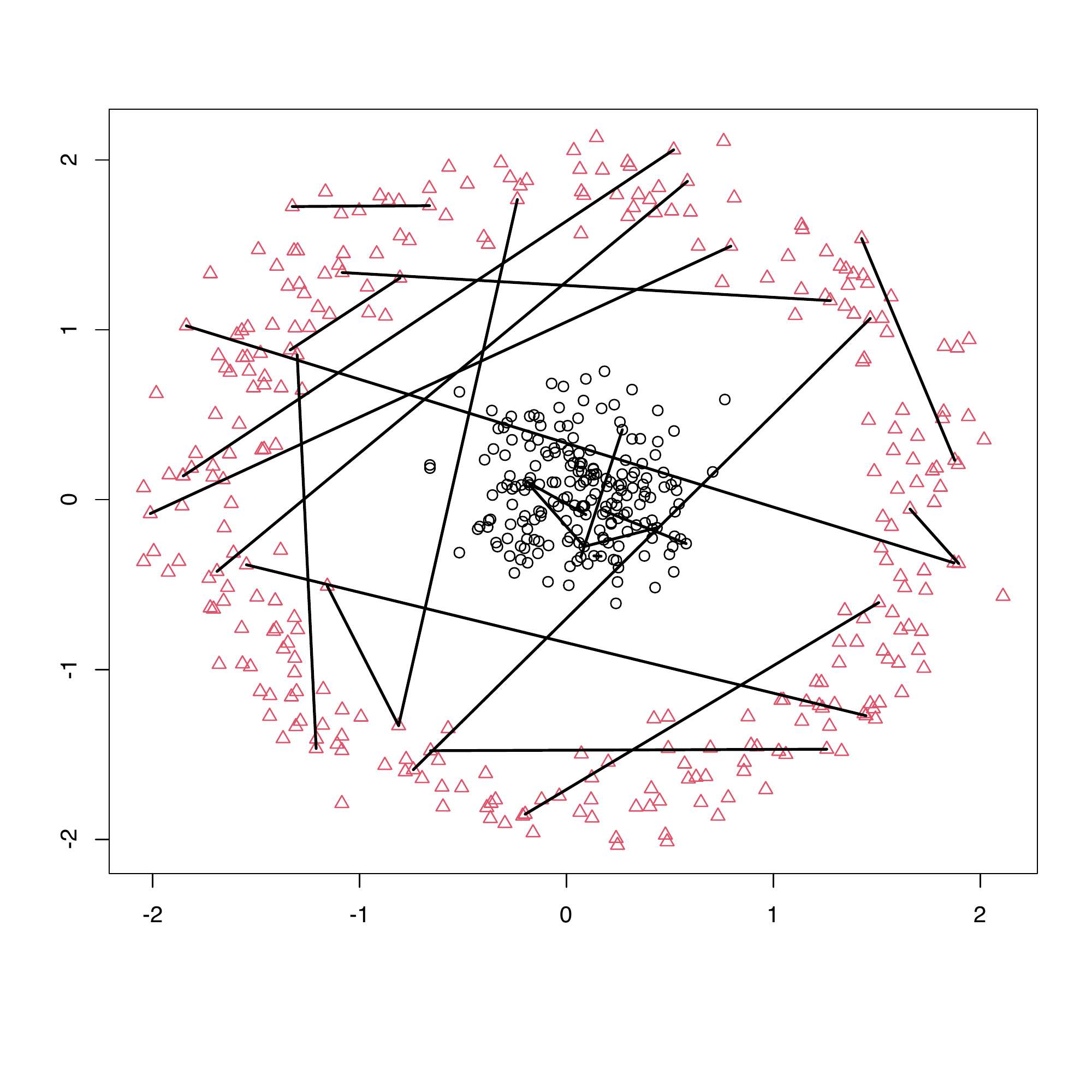}}
\subfloat[\label{fig:circles_CL}]{\includegraphics[width=0.35\textwidth]{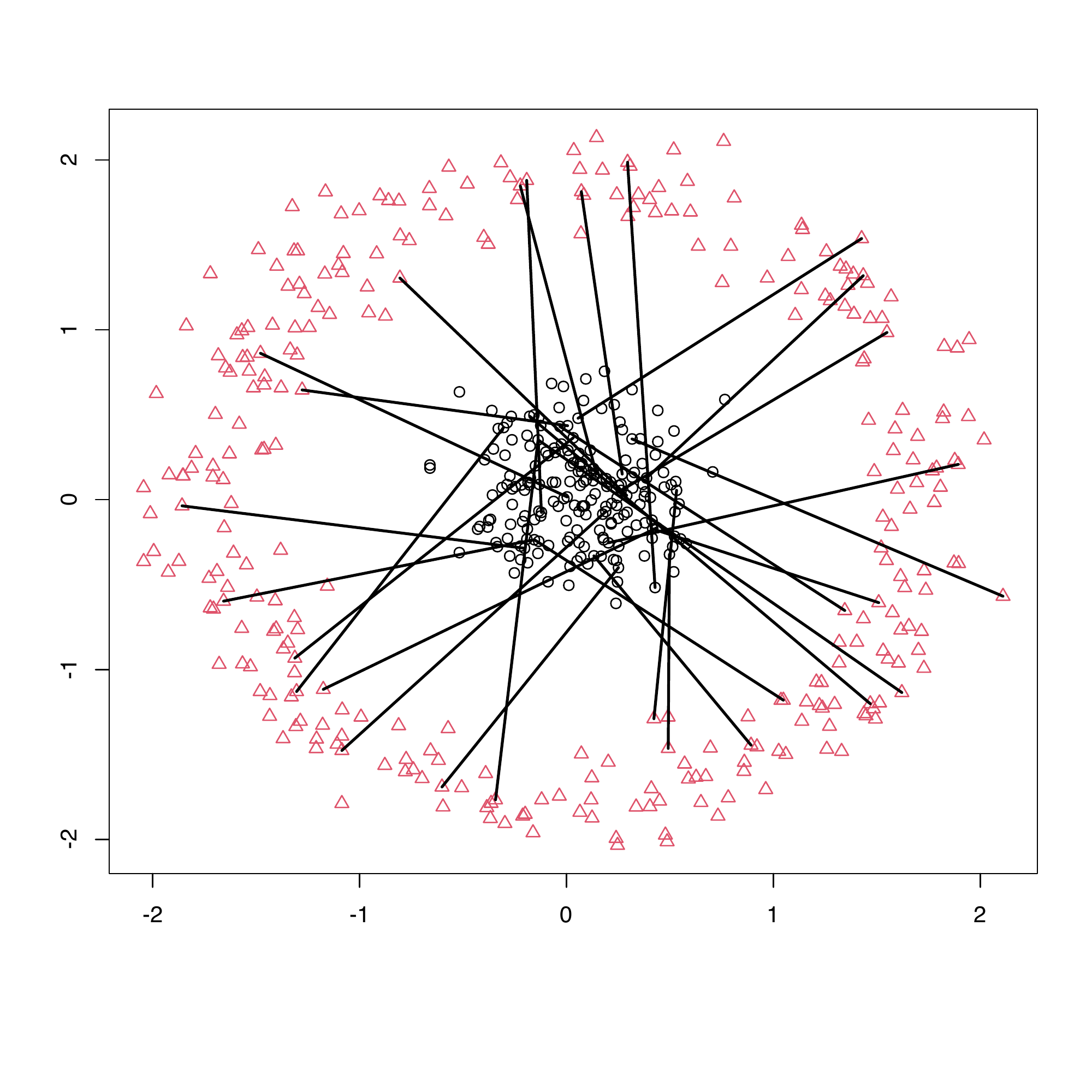}}
\caption{The \textsf{circles} dataset with must-link (a) and cannot-link (b) constraints.  \label{fig:circles_MLCL}}
\end{figure}

\begin{figure}
\centering  
\includegraphics[width=0.4\textwidth]{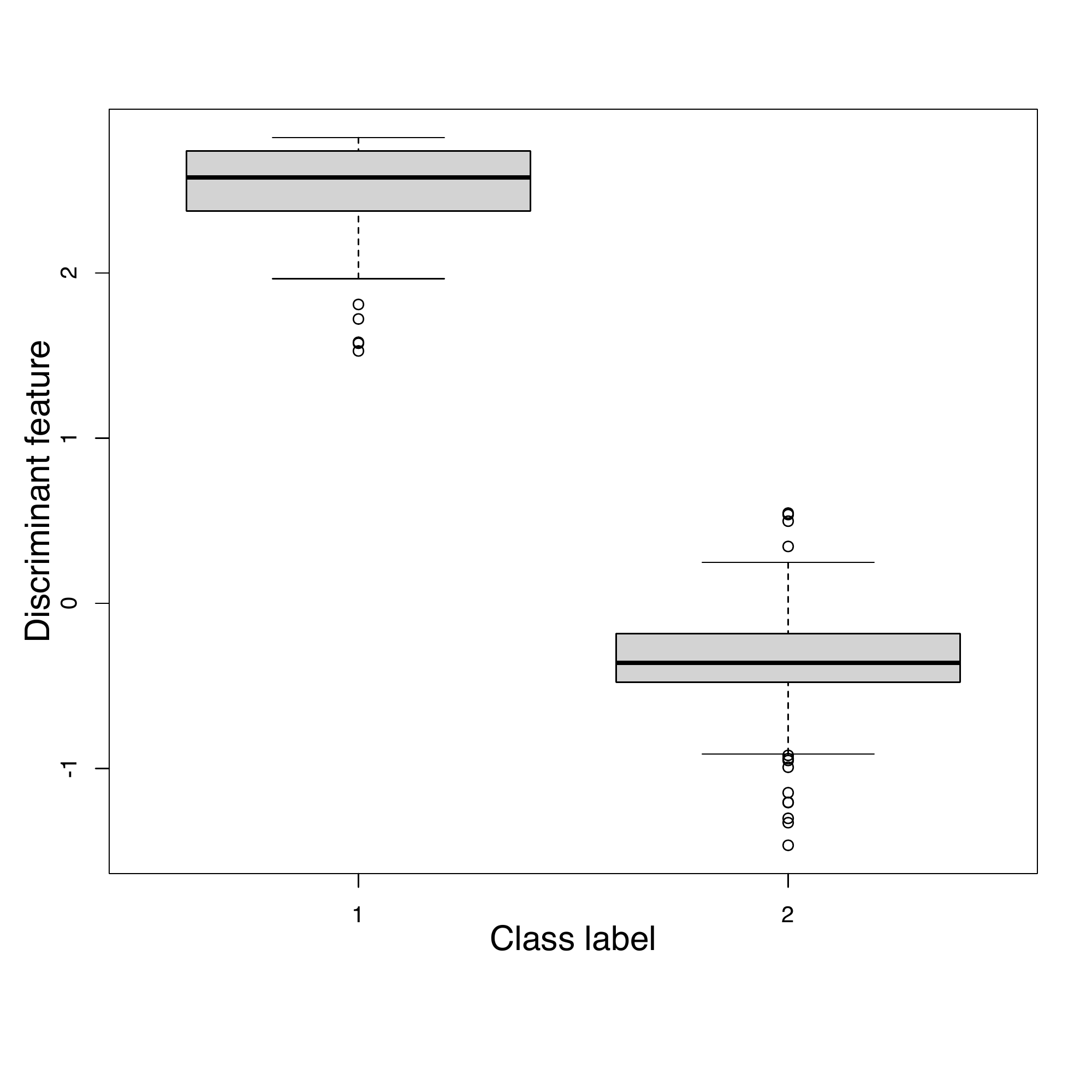}
\caption{Boxplots of the feature extracted by KPCCA for the  \textsf{circles} data with the pairwise constraints shown in Figure \ref{fig:circles_MLCL}. \label{fig:circles_PCCA}}
\end{figure}

\begin{figure}
\centering  
\subfloat[\label{fig:circles_dist}]{\includegraphics[width=0.35\textwidth]{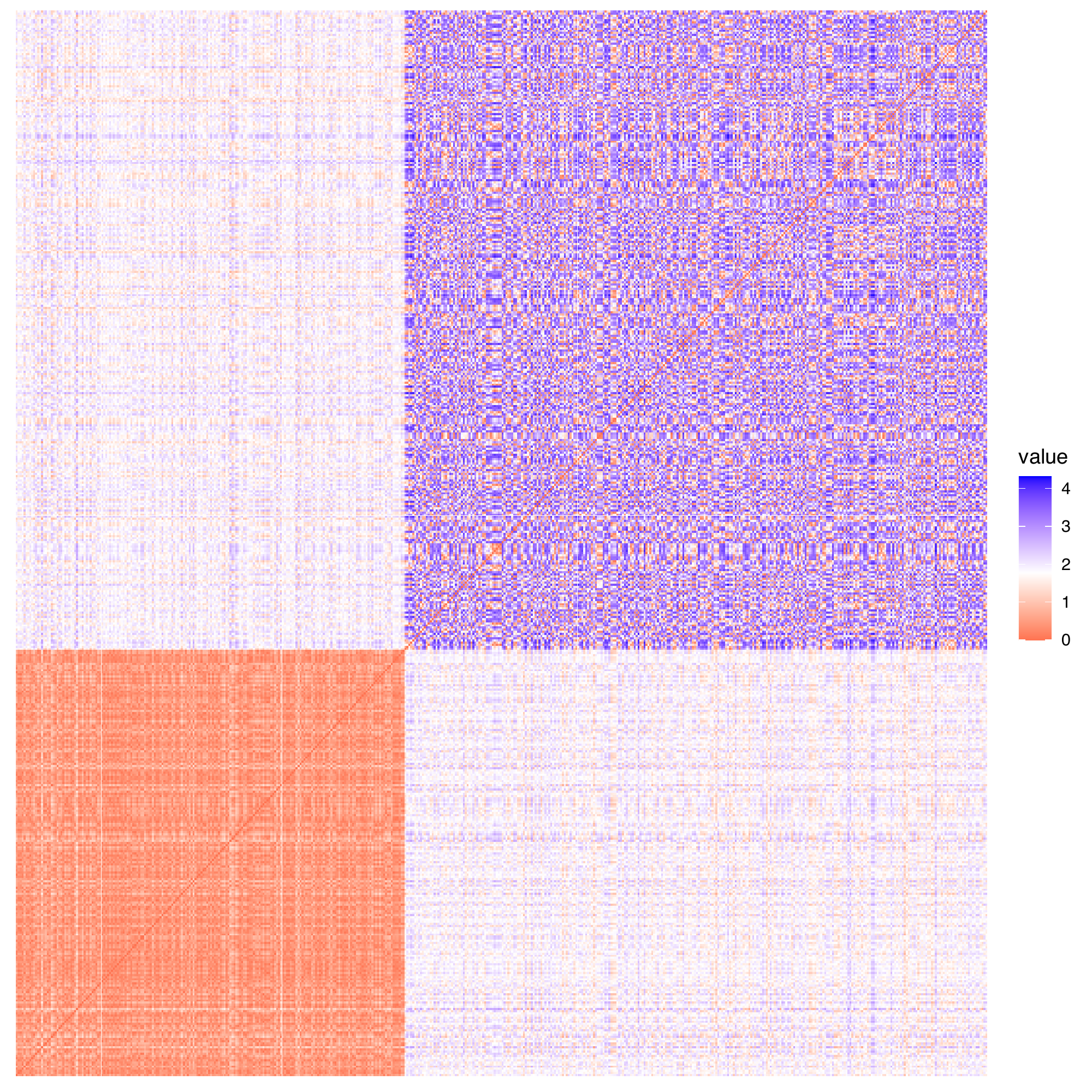}}
\subfloat[\label{fig:circles_dist_PCCA}]{\includegraphics[width=0.35\textwidth]{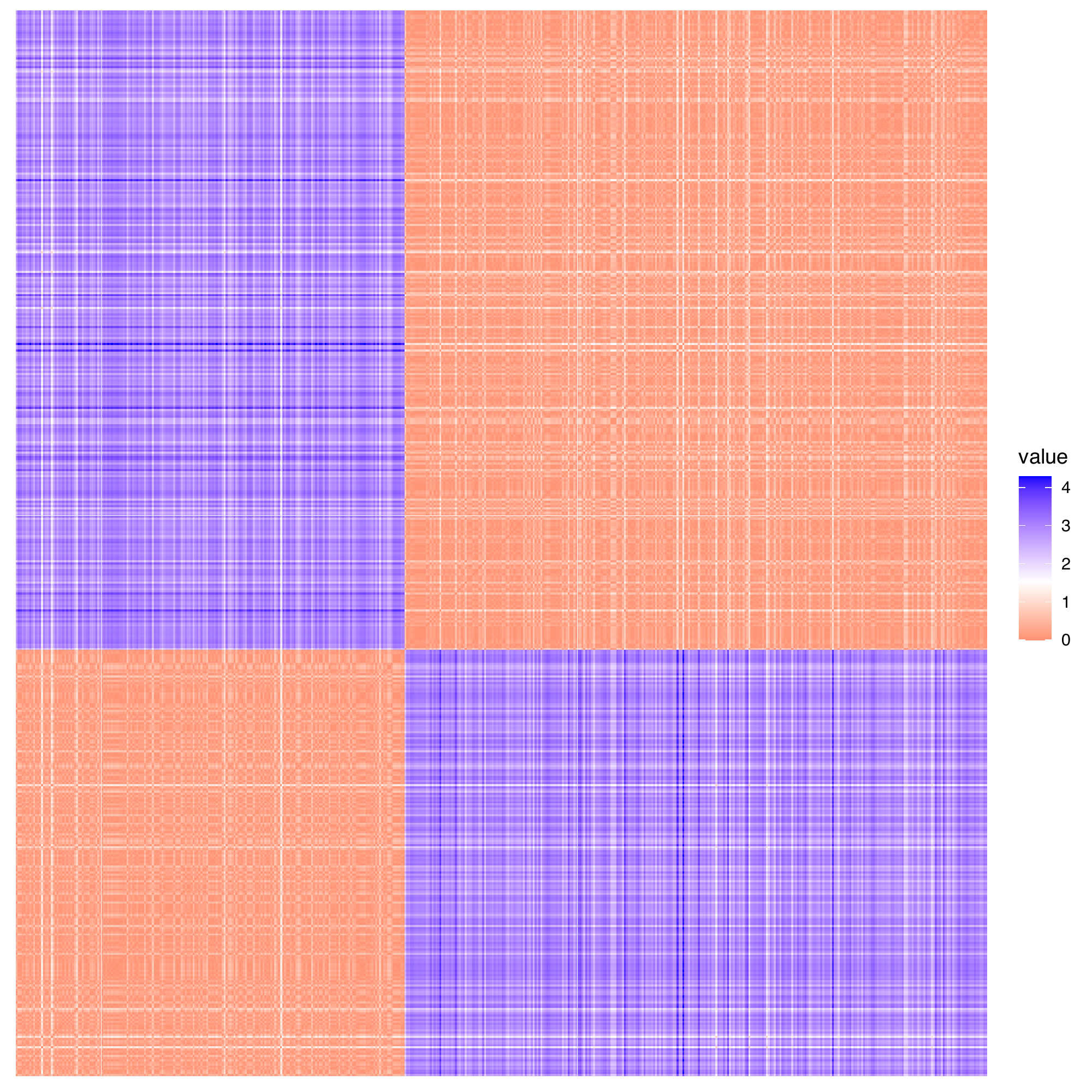}}
\caption{Image representations of the distances matrices of the \textsf{circles} data  in the original space (a) and in the transformed space generated by KPCCA (b). (This figure is better viewed in color). \label{fig:circles_dist1}}
\end{figure}

\begin{figure}
\centering  
\subfloat[\label{fig:circles_empty}]{\includegraphics[width=0.4\textwidth]{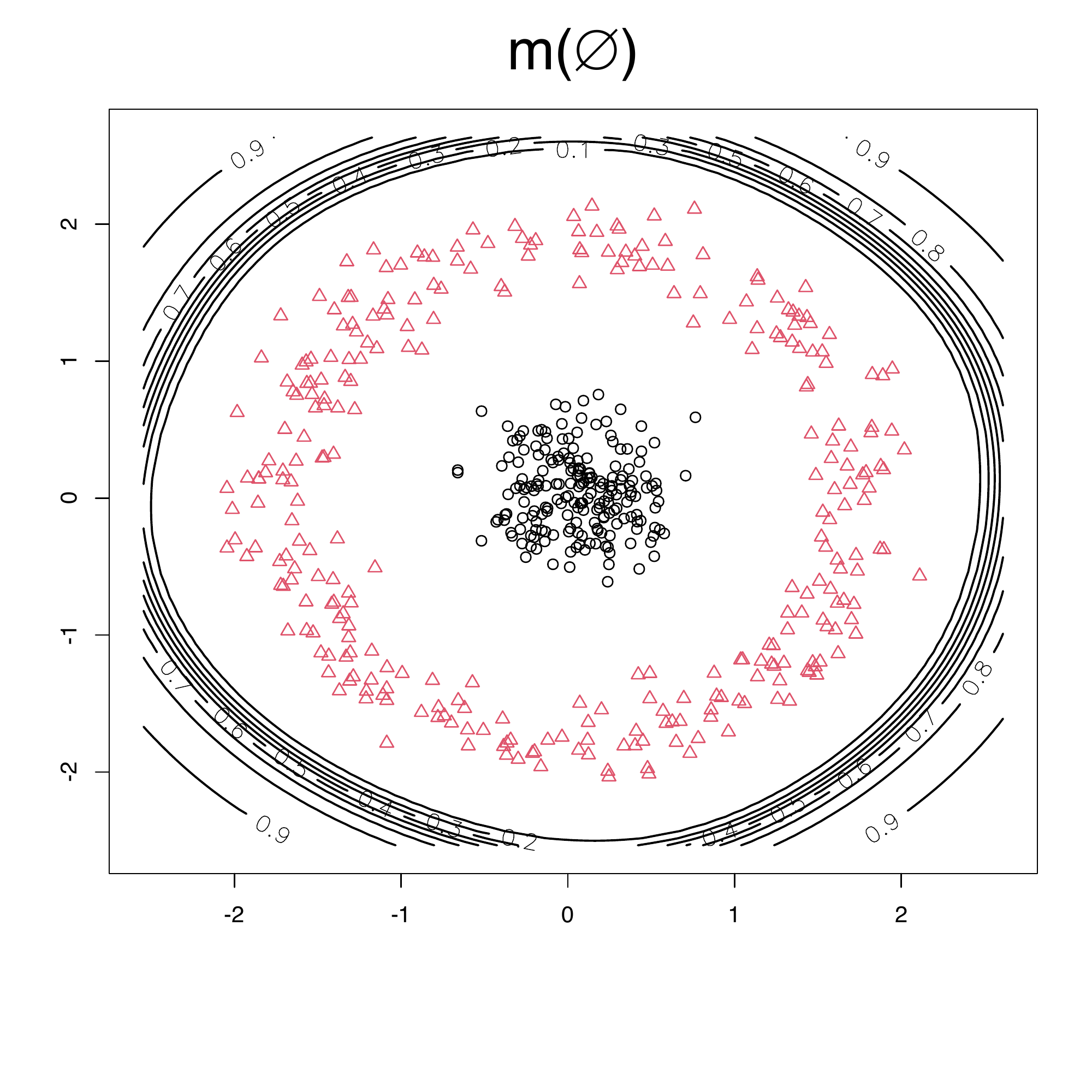}}
\subfloat[\label{fig:circles_omega1}]{\includegraphics[width=0.4\textwidth]{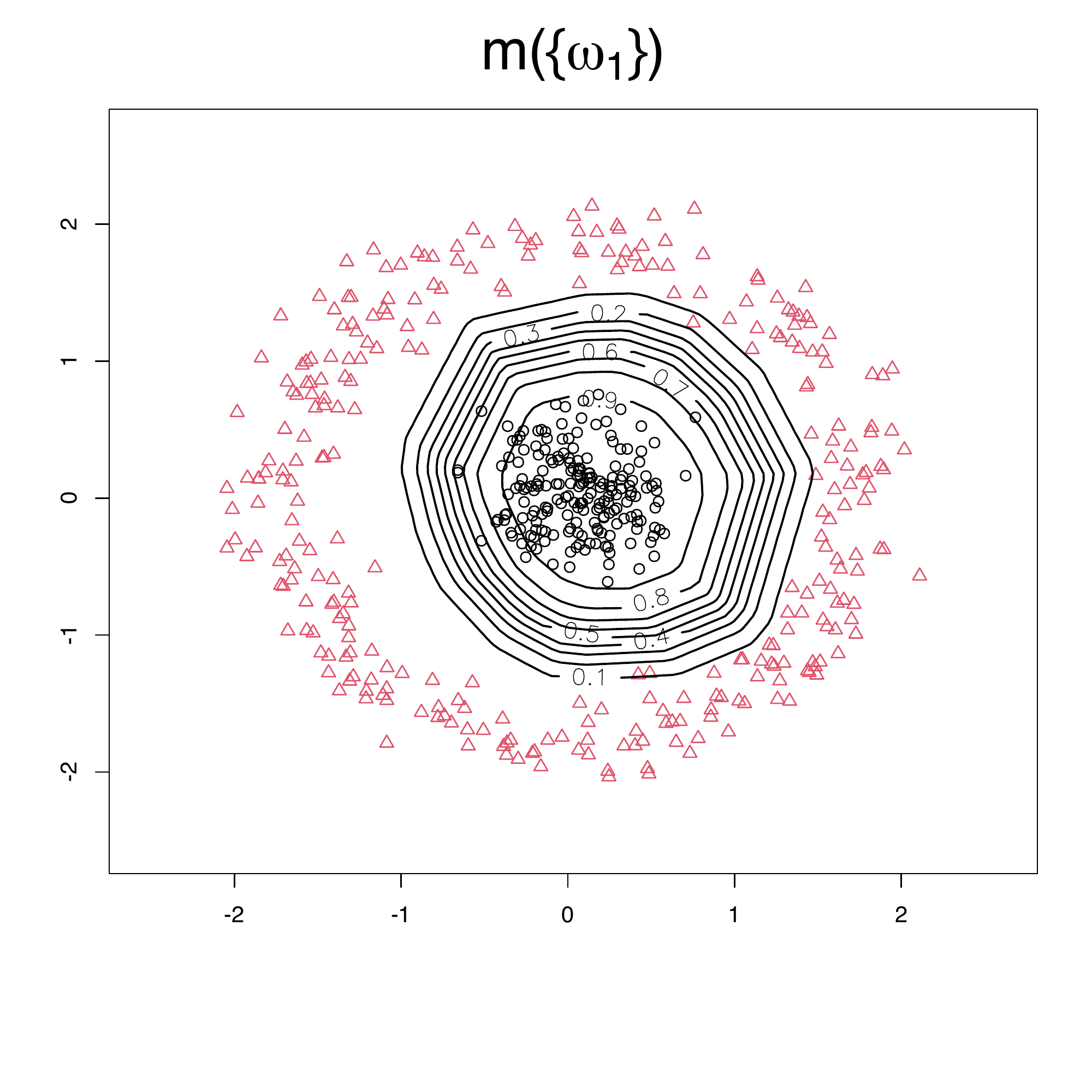}}\\
\subfloat[\label{fig:circles_omega2}]{\includegraphics[width=0.4\textwidth]{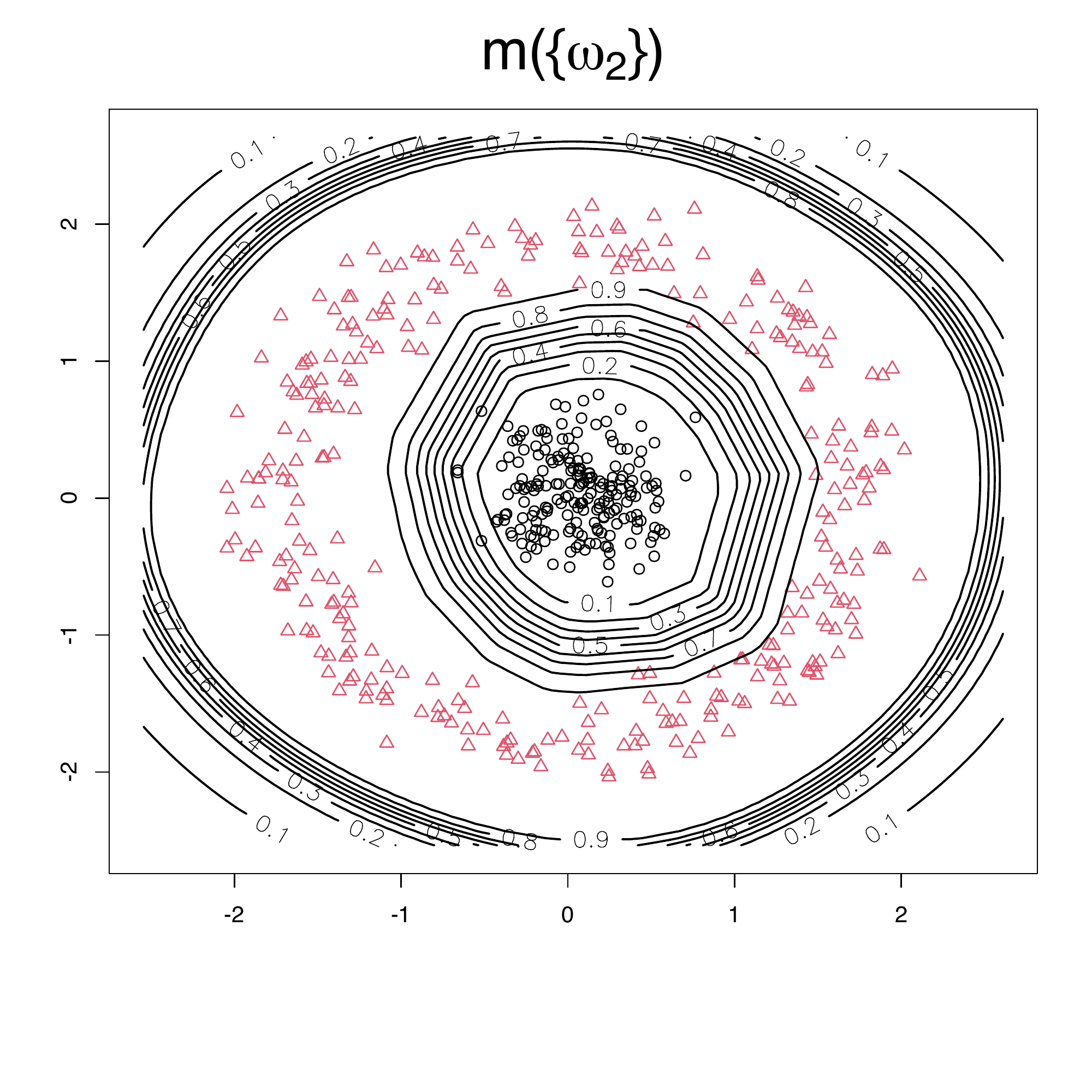}}
\subfloat[\label{fig:circles_Omega}]{\includegraphics[width=0.4\textwidth]{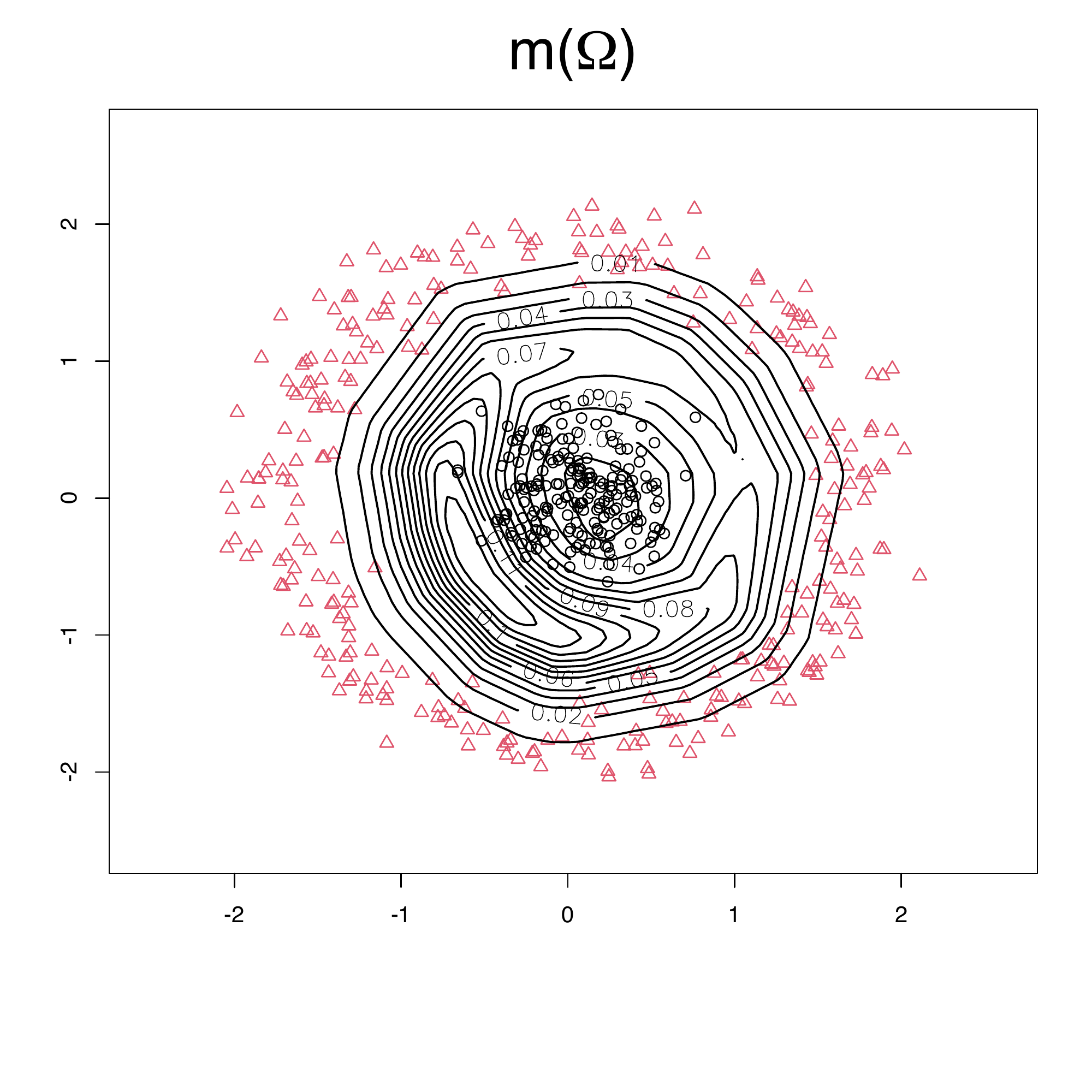}}
\caption{Contour plots of  the masses assigned to each of the four focal set by NN-EVCLUS trained on the \textsf{circles} dataset with the pairwise constraints shown in Figure \ref{fig:circles_MLCL}.  \label{fig:NN_circles_contour}}
\end{figure}

\begin{figure}
\centering
\subfloat[\label{fig:circles_pl1}]{\includegraphics[width=0.4\textwidth]{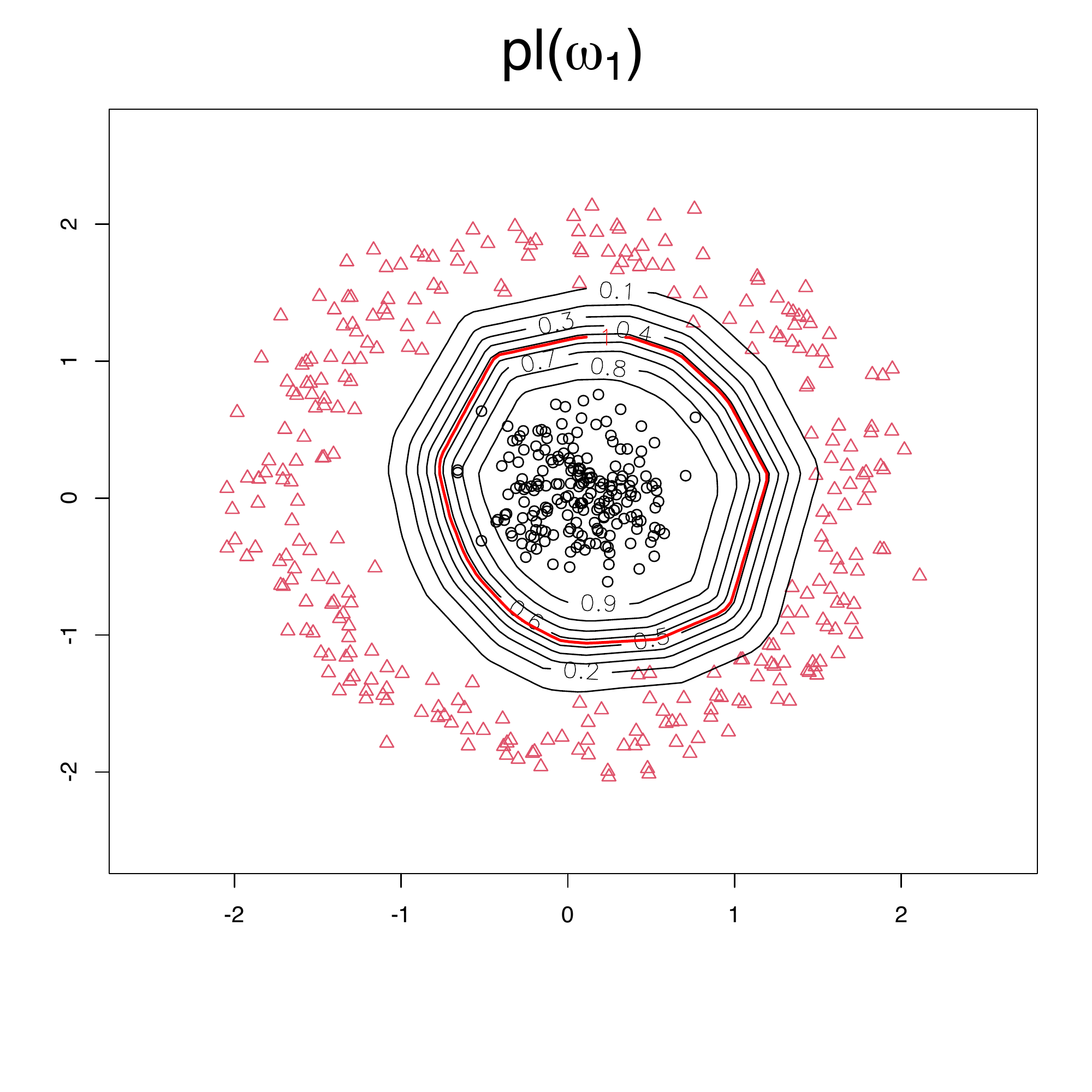}}
\subfloat[\label{fig:circles_pl2}]{\includegraphics[width=0.4\textwidth]{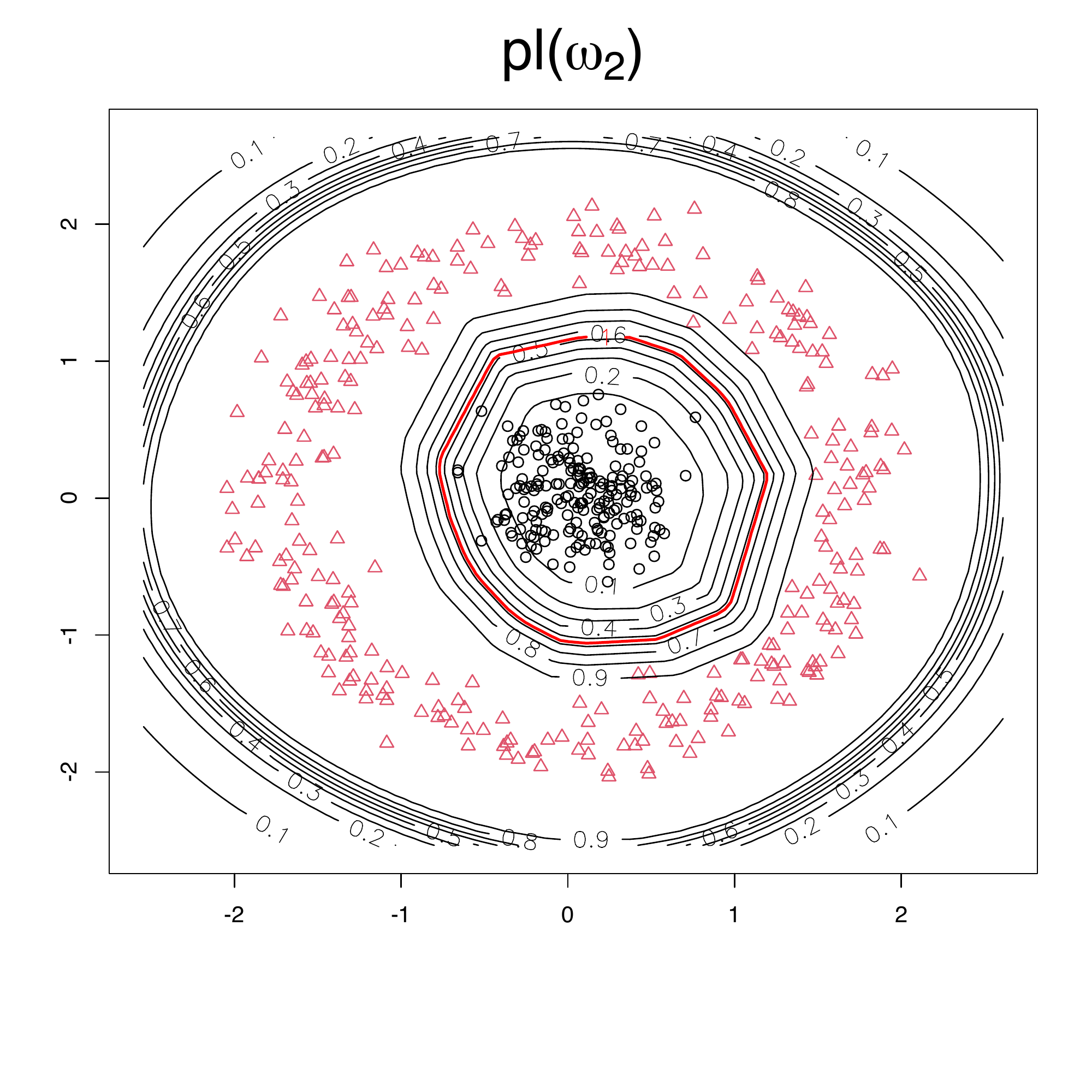}}
\caption{Contour plot of  plausibilities of each of the two clusters obtained by NN-EVCLUS for the \textsf{circles} dataset with the pairwise constraints shown in Figure \ref{fig:circles_MLCL}.  The red thick line represents the decision boundary between the two clusters defined as the curve with equation $pl(\omega_1)= pl(\omega_2)$. \label{fig:NN_circles_contour_pl}}
\end{figure}

\begin{figure}
\centering
\subfloat[\label{fig:circles_semi}]{\includegraphics[width=0.4\textwidth]{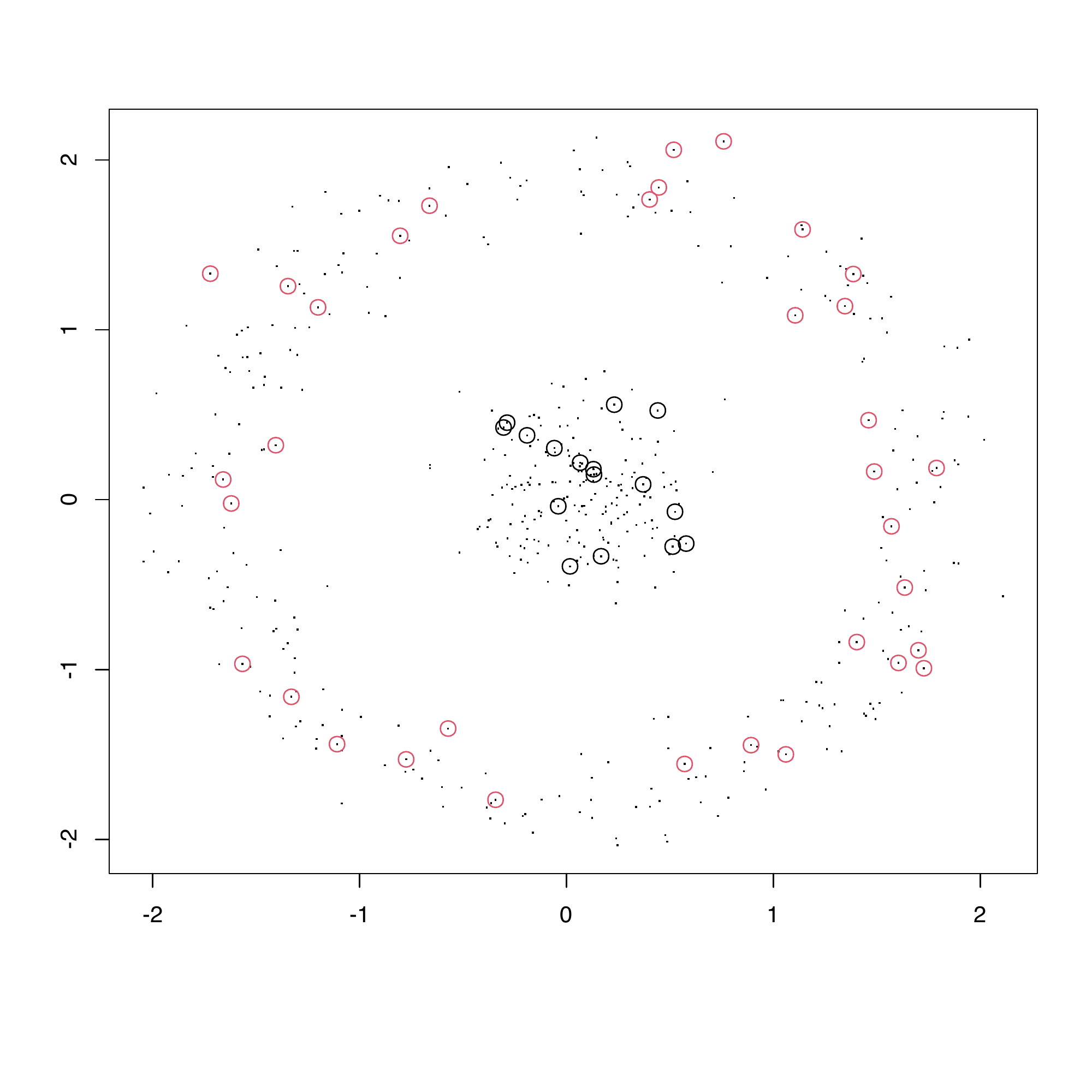}}
\subfloat[\label{fig:circles_kfda}]{\includegraphics[width=0.4\textwidth]{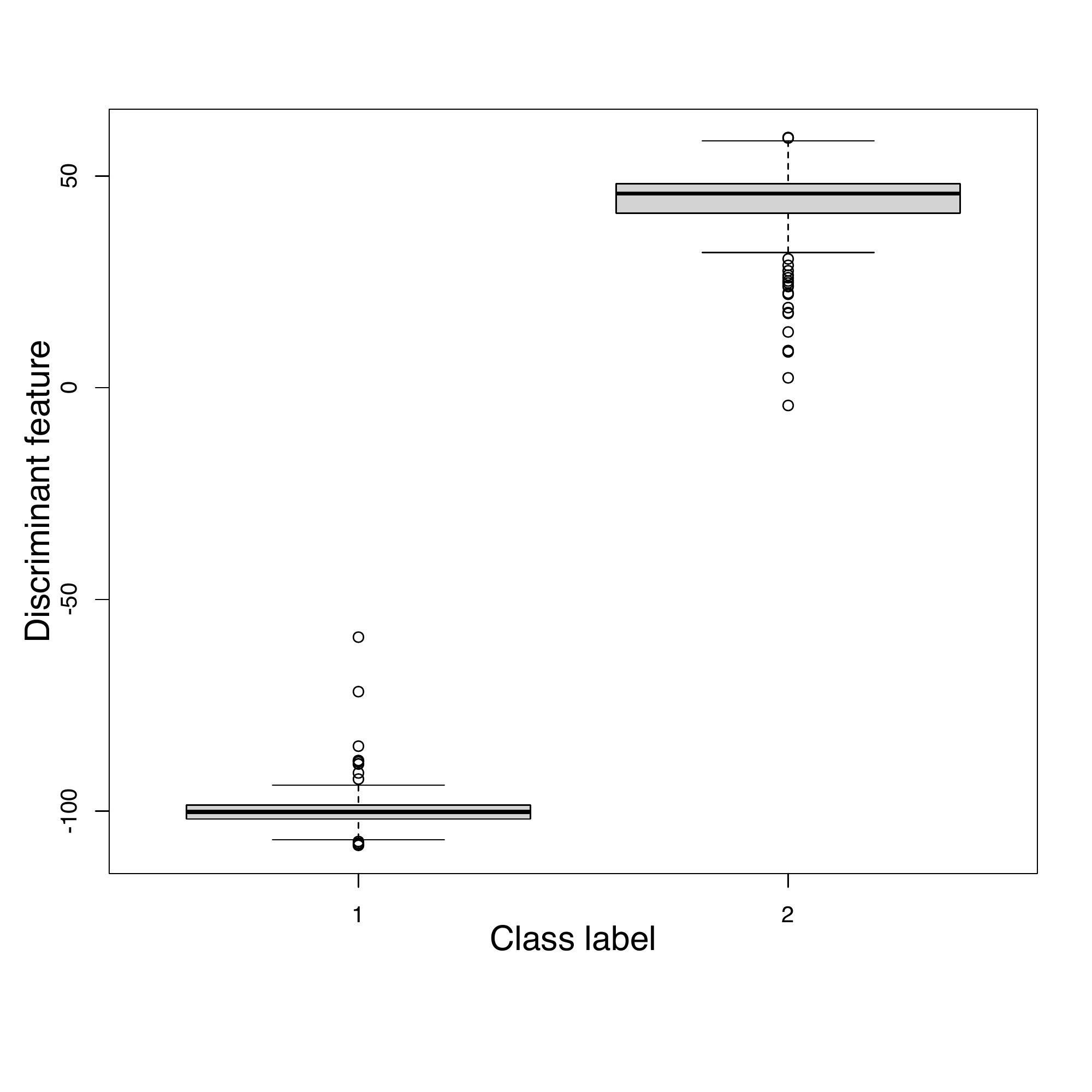}}
\caption{The \textsf{circles} dataset with 50 randomly selected labeled instances (a), and boxplot of the discriminant feature extracted by KFDA (b) using the labeled data. \label{fig:NN_circles_semi}}
\end{figure}

\begin{figure}
\centering
\subfloat[\label{fig:circles_pl1_semi}]{\includegraphics[width=0.4\textwidth]{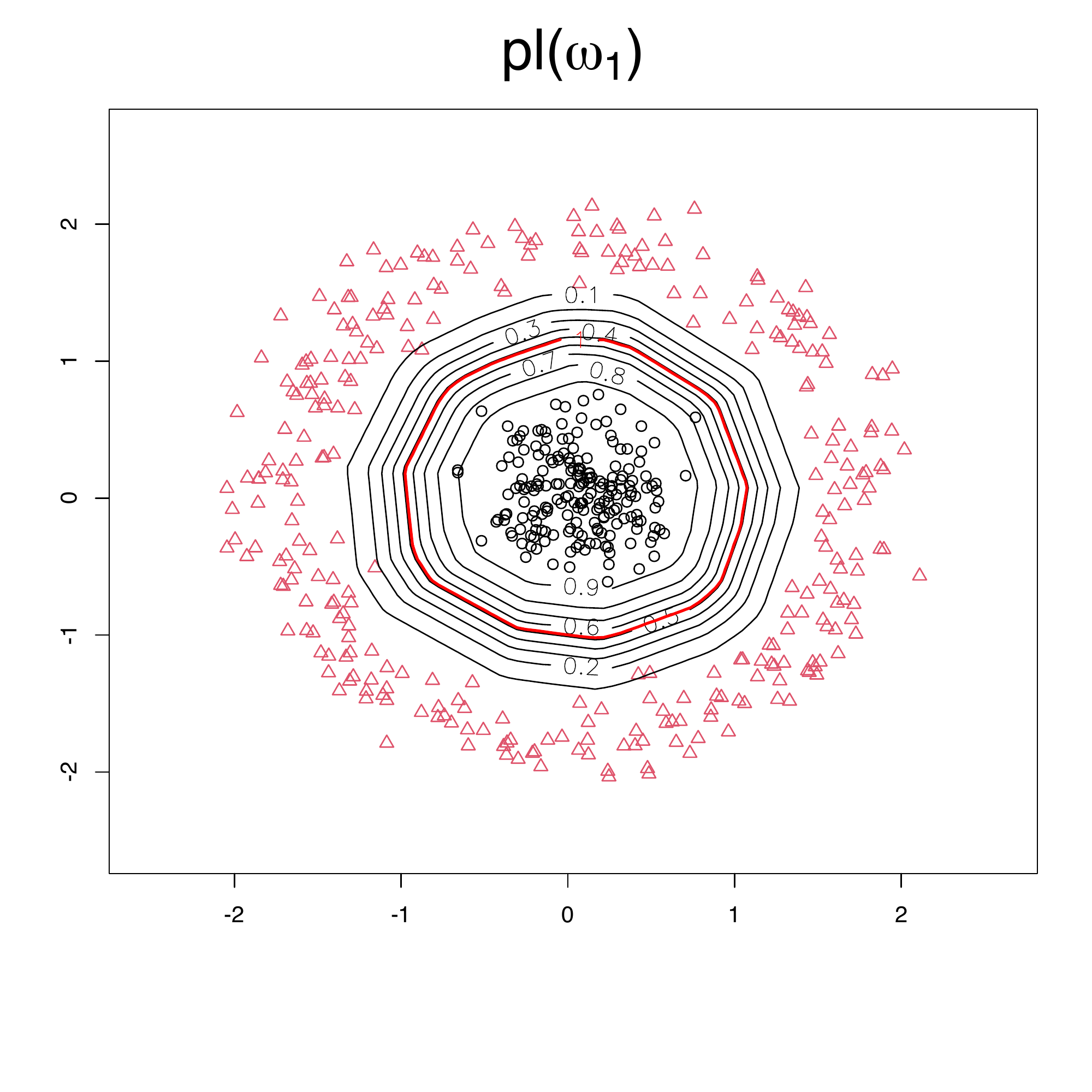}}
\subfloat[\label{fig:circles_pl2_semi}]{\includegraphics[width=0.4\textwidth]{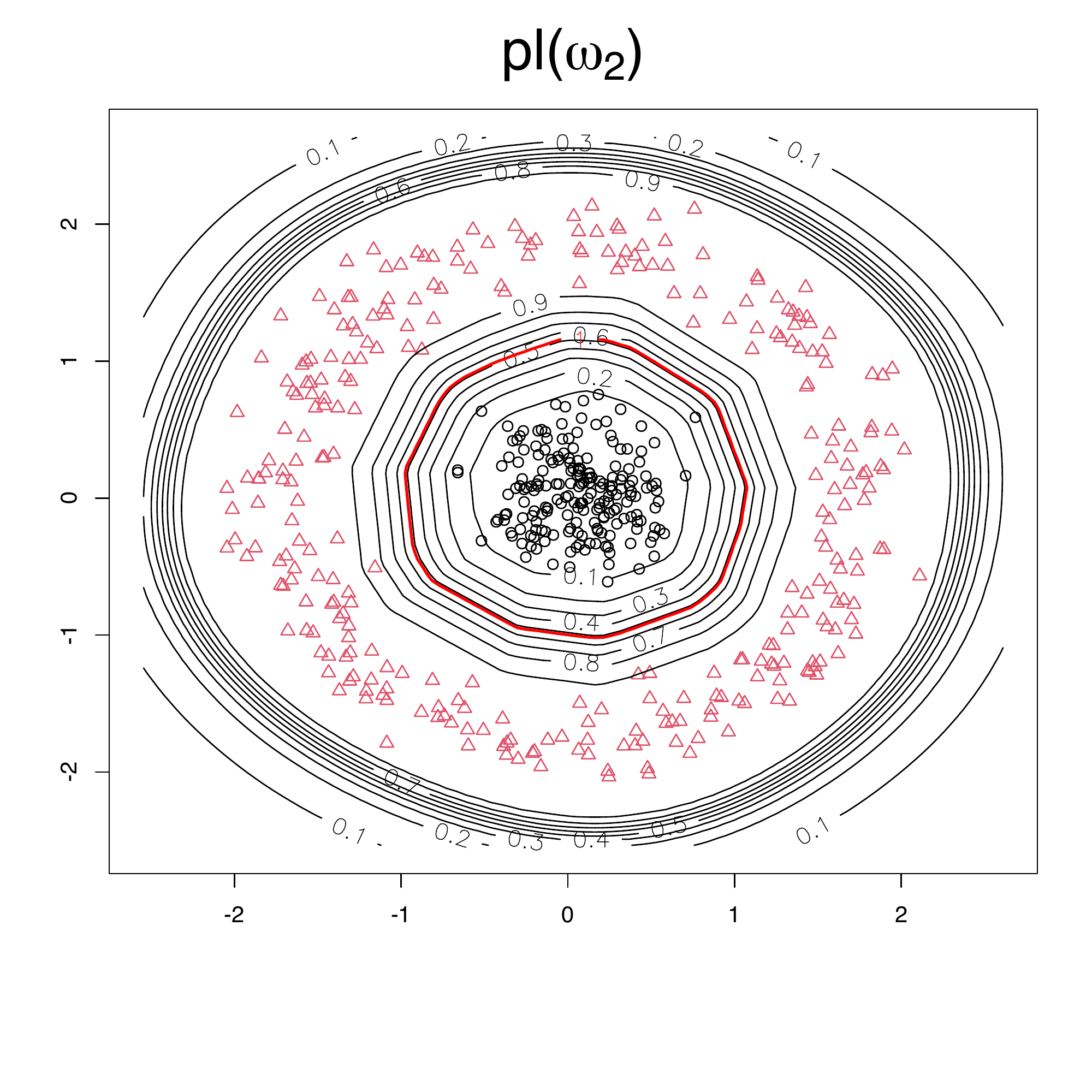}}
\caption{Contour plot of  plausibilities of each of the two clusters obtained by NN-EVCLUS for the \textsf{circles} dataset with the 50 labeled instances shown in Figure \ref{fig:circles_semi}.  The red thick line represents the decision boundary between the two clusters defined as the curve with equation $pl(\omega_1)= pl(\omega_2)$. \label{fig:NN_circles_contour_pl_semi}}
\end{figure}

\end{Ex}

\section{Comparative experiments}
\label{sec:exper}

In this section, we compare the performances of NN-EVCLUS with those of alternative evidential clustering algorithms on a sample of publicly available datasets. \new{All the simulations  were performed using an  implementation of our algorithm in R publicly available  in  package {\tt evclust} \cite{denoeux21}.} Fully unsupervised clustering of attribute and dissimilarity data will first be addressed, respectively,  in Sections \ref{subsec:attr} and \ref{subsec:rel}. Clustering with pairwise constraints will then be studied in Section \ref{subsec:exper_semi}.

\subsection{Unsupervised clustering of attribute data}
\label{subsec:attr}

\paragraph{Data sets} We  considered  the \new{14} publicly available real and artificial datasets summarized in Table \ref{tab:attribute_data}. These datasets all contain attribute data and have a wide range of characteristics in terms of input dimension and number of clusters. For the \textsf{Ecoli} dataset, we used only the quantitative attributes (2, 3, 6, 7, and 8) and the four most frequent classes: `im', `pp', `imU' and `cp';  we then merged the classes `im' and `imU', resulting in a dataset with 307 objects described by five attributes and partitioned into three clusters. The \textsf{Letters4p1} is a subset of the ``Letter Recognition'' dataset from the UCI machine learning repository \cite{dua19} containing six clusters. \new{The \textsf{Mice} dataset is a part of the data analyzed in \cite{higuera15}; it contains the expression levels of  22  proteins for the trisomic mice: we considered the 26 proteins  listed in columns 4 and 5 of Table 3 in \cite{higuera15}, and we retained only the 22 of them without missing values. The three classes  are: ``t-SC-m'' (shock-context with  memantine), ``t-SC-s'' (shock-context with  saline) and ``t-CS'' (context-shock with either memantine or saline). The \textsf{DryBean} dataset is a subset of the data analyzed in \cite{koklu20}, with 200 randomly selected objects in each class. The \textsf{Leaves5p1} dataset \cite{franti18} is a subset of the ``One-hundred plant species leaves'' in the UCI database \cite{dua19} containing 320 objects from 20 classes. All datasets contain only numerical attributes; the dissimilarities were computed as Euclidean distances between attribute vectors.}

\begin{table}
\caption{Number $n$ of objects, number $d$ of attributes and number $c$ of clusters for each of the 14 attribute datasets used in this study. \label{tab:attribute_data}}
\begin{center}
\begin{tabular}{ccccc}
\hline
Name & $n$ & $d$ & $c$ & Source\\
\hline
\textsf{Wine} & 178& 13 & 3 & \cite{dua19} \\
\textsf{Iris} & 150 & 4 & 3 & \cite{R20}\\
\textsf{Ecoli} & 307 & 5 & 3 &  \cite{dua19}\\
\textsf{Heart} & 270& 13 & 2 &  \cite{dua19}\\
\textsf{Seeds} & 210& 7 & 3 &  \cite{dua19}\\
\textsf{Letters4p1} & 4634& 16 & 6 & \cite{franti18}\\
\textsf{Glass} & 214&10 & 6 &  \cite{dua19}\\
\textsf{Segment} &2310  & 16& 7 &  \cite{dua19}\\
\textsf{S2} & 5000 & 2 & 15 & \cite{franti18}\\
\textsf{S4} & 5000 & 2  & 15 &\cite{franti18}\\
\textsf{D31} & 3100 & 2& 31 & \cite{franti18}\\
 \new{\textsf{Mice}} & \new{507} & \new{22} & \new{3} & \new{\cite{dua19}\cite{higuera15}}\\
 \new{\textsf{DryBean}} & \new{1400} & \new{16} & \new{7} & \new{\cite{dua19}\cite{koklu20}}\\
 \new{\textsf{Leaves5p1}} & \new{320} & \new{64} & \new{20} & \new{\cite{franti18}}\\
\hline
\end{tabular}
\end{center}
\end{table}

\paragraph{Algorithms} As alternative evidential clustering algorithms, we considered EVCLUS \cite{denoeux16a}, ECM \cite{masson08}, CCM\footnote{Our R code for CCM was translated from Matlab code provided by Prof. Zhunga Liu.} \cite{liu15}, sECMdd, wECMdd\footnote{The R code for sECMdd and wECMdd was provided by Dr. Kuang Zhou.} \cite{zhou16},  BPEC \cite{su19} \new{and Bootclus \cite{denoeux20d}}. The number of clusters was assumed to be known. For EVCLUS and NN-EVCLUS, we restricted the focal sets to the empty set, singletons, pairs and $\Omega$, except when the number $c$ of clusters was equal to or larger than 5, in which case the pairs were not included in the focal sets. Parameter $\delta_0$ was set to the 0.9-quantile of distances for $c\le 4$ and to a smaller value (0.5, 0.2 or 0.1 quantile) for datasets with a larger number of clusters (as a heuristic, $\delta_0$ should be smaller when the number of clusters is larger). For NN-EVCLUS, the number of hidden units was set to 1.5 times the number of focal sets. We used batch learning for small datasets ($n \le 1000$) and minibatch learning with the RMSprop algorithm \cite[page 300]{goodfellow16} for larger datasets. For ECM, we used as focal sets the empty set, singletons and pairs. When the number of clusters was strictly greater that 3, we used the two-step strategy described in \cite{denoeux16a}: we first trained the model with the empty set and singletons; we then identified pairs of clusters that are mutual nearest neighbors according to a similarity measure, and we re-trained the model after including these pairs as focal sets. The same strategy was applied with BPEC, for which we used the version of ECM with an adaptive metric \cite{antoine12}. \new{The Bootclus method is based on the bootstrap and Gaussian mixture models (GMMs); we used the default settings of function {\tt bootclus} in package {\tt evclus}: $B=500$ bootstrap samples and bootstrap percentile confidence intervals at level $1-\alpha=0.9$. Function {\tt bootclus} calls function {\tt Mclust} of package {\tt mclust}, which selects the best GMM according to the Bayesian information criterion (BIC).}

The $\delta-Bel$ method was used to identify cluster centers. When these centers were clearly discernible in the $\delta-Bel$ graph, they were used with BPEC. We also considered ways of exploiting this information with other methods. For ECM and CCM, the cluster centers were used as initial prototypes (the corresponding methods will be denoted, respectively,  as ECM-bp and CCM-bp). For NN-EVCLUS, we treated these centers as labeled data and optimized loss function \eqref{eq:loss_semi} with $\nu=0.5$; the corresponding method will be denoted as NN-EVCLUS-bp. \new{For the \textsf{Leaves5p1} dataset, as the belief peaks did not identify all clusters,  we used the attribute vectors with maximum plausibility in each class (one vector per class) obtained with EVCLUS instead.} The CCM algorithm was used with the empty set, the singletons, the pairs and $\Omega$ as focal sets; the parameters were set to the default parameters as recommended by the authors \cite{liu15}, i.e., $\gamma=1$, $\beta=2$ and $T_c=2$. As in ECM and BPEC, parameter $\delta$, which controls the number of outliers was set to 5. Similarly, we used the default values recommended in \cite{zhou16} for sECMdd ($\beta = 2$, $\alpha = 2$, $\eta = 1$, $\gamma = 1$) and wECMdd ($\beta = 2$, $\alpha = 2$, $\xi = 5$, $\psi = 2$). All algorithms were run five times and we kept the best solution in terms of loss or objective function.

\begin{figure}
\centering  
\includegraphics[width=0.35\textwidth]{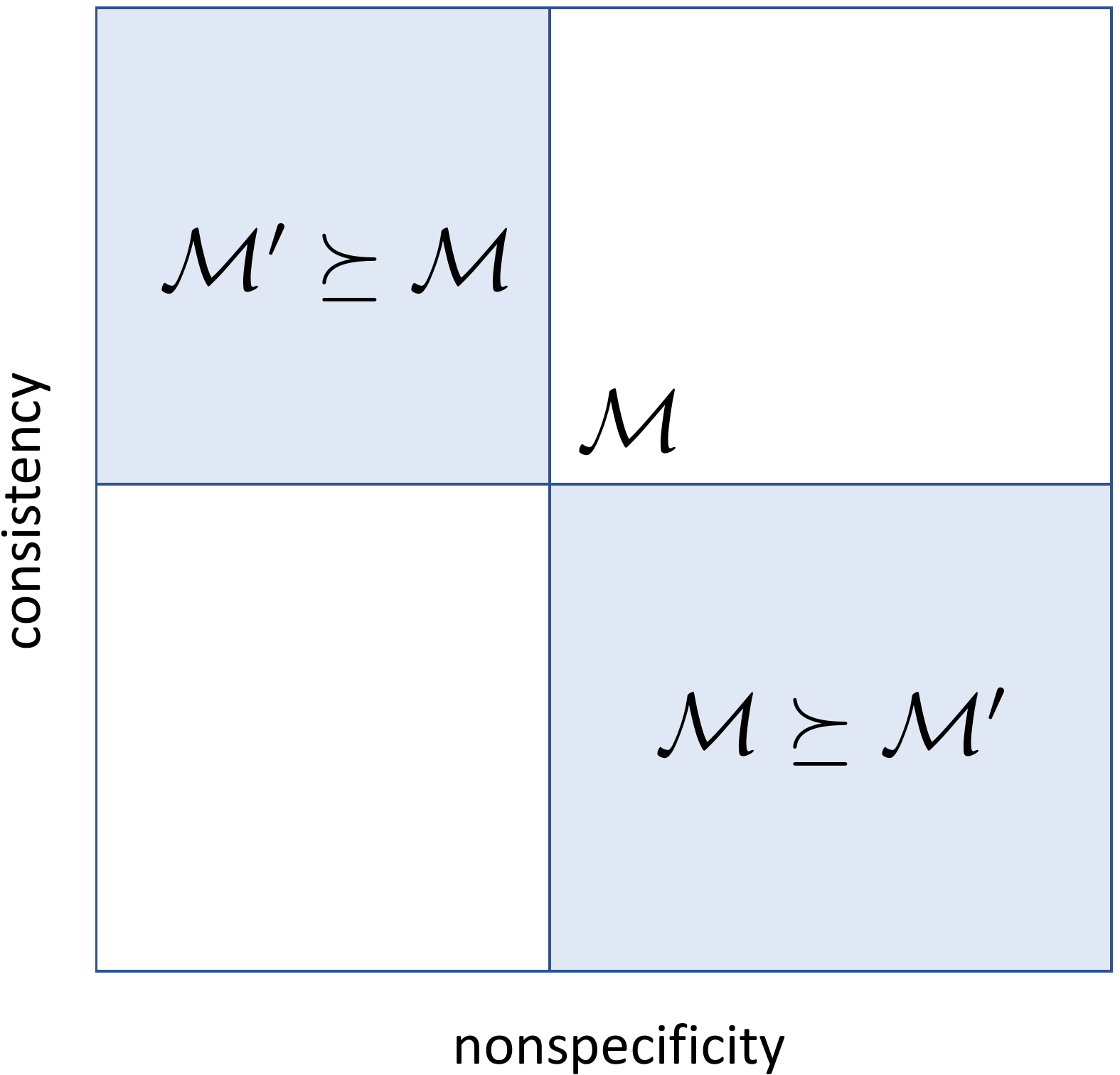}
\caption{Partial order induced by consistency and nonspecificity: evidential partition $\calM'$ dominates $\calM$ (denoted as $\calM' \succeq \calM$) if it has higher consistency and lower nonspecificity, and it is dominated by $\calM$ (denoted as $\calM \succeq \calM'$) if it has lower consistency and higher nonspecificity.  \label{fig:NSCI_dominance}}
\end{figure}

\paragraph{Performance criteria} Measuring the quality of evidential partitions is a difficult problem. One approach is to convert the evidential partition into a hard partition by assigning each object to the cluster with the highest plausibility, and compare the resulting hard partition to the ground-truth partition using, e.g., the adjusted Rand index (ARI) \cite{hubert85}. This approach provides easily interpretable results and makes it possible to rank evidential clustering algorithms according to the similarity between the evidential partition and the true partition. However, it does not account  for the specific characteristics of evidential partitions. In \cite{denoeux17a}, we proposed  evidential extensions of the Rand index, and we  argued that the quality of an evidential partition could be described by two numbers: a consistency index (CI) measuring its agreement with the true partition, and a nonspecificity index (NS) measuring its imprecision. We generally  aim at high consistency and low nonspecificity, so that the pair of indices (CI,NS) induces a partial order: an evidential partition $\calM'$ is better than an evidential partition $\calM$ if it has a higher consistency index and a lower nonspecificity (see Figure \ref{fig:NSCI_dominance}) \cite{denoeux17a}. Here, we computed the three indices (ARI, CI and NS) for each of the methods applied to each dataset.

\paragraph{Results and discussion} The ARI values obtained by the 11 methods on the 14 datasets are shown in Table \ref{tab:ARI_attribute_data}. For each dataset, the best result is printed in bold, and the values within 5\% of the best result are underlined. The $\delta-Bel$ method failed to identify the centers of all clusters for the \textsf{Letters4p1}, \textsf{Glass}, \textsf{Segment} \new{and \textsf{Leaves5p1}} datasets; for this reason, the methods using cluster centers identified by this method (NN-EVCLUS-bp, ECM-bp, CCM-bp and BPEC) are not given for these datasets. Also, our implementations of sECMdd, wECMdd \new{and Bootclus} failed to converge in a reasonable amount of time for the \textsf{S2}, \textsf{S4} and \textsf{D31} characterized by large numbers of objects and clusters; no results are thus reported for these methods on these datasets. 

We can see from Table \ref{tab:ARI_attribute_data} that NN-EVCLUS  yielded either the best results in terms of ARI, or results close to the best ones \new{for all datasets, except \textsf{Iris} and \textsf{Segment}, for which Bootclus gave better results, which can be explained by the presence of nonspherical clusters; even for these two datasets, NN-EVCLUS yielded the second best results in terms of ARI.} Using the cluster centers identified by the $\delta-Bel$ method makes it possible to improve the results of NN-EVCLUS most of the time, particular when the number of clusters is large (as in the \textsf{S2}, \textsf{S4} and \textsf{D31} datasets). Whereas the methods based on prototypes (ECM, CCM and BPEC) work well on artificial datasets with well-separated clusters, they are outperformed by EVCLUS and NN-EVCLUS on real datasets characterized by highly overlapping clusters. The sECMdd and wECMdd were outperformed by other methods on all datasets. NN-EVCLUS with random initialization performs equally well as, or better than EVCLUS on most datasets, except when the number of clusters is large (as in the \textsf{S2}, \textsf{S4} and \textsf{D31} datasets), in which case NN-EVCLUS may fail to find a deep minimum of the loss function; in these cases, using prior information provided by the  $\delta-Bel$ method is crucial. \new{When the $\delta-Bel$ fails, using maximum plausibility  attribute vectors provided by EVCLUS seems to be a good strategy for training NN-EVCLUS when the number of clusters is large, as shown by the good results obtained on the \textsf{Leaves5p1} dataset.}

\begin{sidewaystable}
\caption{ARI values for the \new{11} methods on the \new{14} attribute datasets. The best value for each dataset is printed in bold, and the values within 5\% of the best value are underlined. (NN-EV and NN-EV-bp stand, respectively, for NN-EVCLUS and NN-EVCLUS-bp). \label{tab:ARI_attribute_data}}
\begin{center}
\begin{tabular}{cccccccccccc}
\hline
Dataset &  EVCLUS & NN-EV & NN-EV-bp& ECM & ECM-bp & CCM & CCM-bp &sECMdd & wECMdd & BPEC&\new{Bootclus}\\
\hline
\textsf{Wine} & \underline{0.91} &\underline{0.91} &\bf{0.93} &0.85& 0.85& 0.45 &0.45&0.24 &0.39 &0.80&\new{\bf{0.93}}\\
\textsf{Iris} &0.77 &  0.77 &  0.70 &0.62& 0.62& 0.63& 0.63& 0.54& 0.66& 0.64& \new{\bf{0.90}}\\
\textsf{Ecoli} & \underline{0.80}& \underline{0.80} & \bf{0.82} & \underline{0.79} &  \underline{0.79} &  0.70 & 0.70 & 0.11 & 0.64& \underline{0.78}&\new{0.60}\\
\textsf{Heart} & \underline{0.41} &  \bf{0.42}&  \underline{0.41}&  0.38 &  0.38 &  0.37 &  0.40& -0.0058&  0.39  &0.22&\new{0.27}\\
\textsf{Seeds} & \bf{0.81}& \underline{0.80} & 0.75& 0.73 &0.73 & \underline{0.79} & \underline{0.79} & 0.45 & 0.69 & 0.72&\new{0.73}\\
\textsf{Letters4p1} & \underline{0.50} & \bf{0.52} & $\cdot$& 0.050 & $\cdot$&0.089 & $\cdot$&0.075 &0.099 & $\cdot$&\new{0.10}\\
\textsf{Glass} & \underline{0.35}&  \bf{0.36} & $\cdot$&0.31 & $\cdot$&0.21 &$\cdot$ &0.20 &0.26& $\cdot$& \new{0.18}\\
\textsf{Segment} & 0.51 &  0.54 & $\cdot$&0.50 & $\cdot$&0.36 & $\cdot$&0.44 & 0.23& $\cdot$ & \new{\bf{0.61}}\\
\textsf{S2} &0.88 & 0.81 & \underline{0.93} & \underline{0.93} & \underline{0.93} & 0.72 & \bf{0.94}& $\cdot$&$\cdot$ & \underline{0.93}&  $\cdot$\\
\textsf{S4} & \underline{0.63} & 0.55& \underline{0.64} & \underline{0.63}& \underline{0.63} & 0.52& \underline{0.63}& $\cdot$& $\cdot$& \textbf{0.65}&$\cdot$\\
\textsf{D31} & \underline{0.91} & 0.69 & \underline{0.94} & 0.86 & \textbf{0.95} &  0.48& \bf{0.95} & $\cdot$& $\cdot$&\underline{0.94}&$\cdot$ \\
\new{\textsf{Mice}} &\new{0.47} &\new{ 0.69} & \new{\textbf{0.80}} & \new{0.60}  &\new{0.60}  &\new{0.66} & \new{0.70} &\new{-0.02} & \new{0.58} & \new{0.75} & \new{\underline{0.77}}\\
\new{\textsf{DryBean}} & \new{0.62} &\new{0.63} &\new{\textbf{0.75}} &\new{\underline{0.73}} &\new{0.66} &\new{0.51} &\new{0.60} &\new{0.47} &\new{0.28} &\new{0.67} &\new{0.67}\\
\new{\textsf{Leaves5p1}} & \new{0.60} & \new{\textbf{0.64}} & $\cdot$ &\new{0.42} & $\cdot$& \new{0.25} &$\cdot$ & \new{0.20} &\new{0.25}& $\cdot$&\new{\underline{0.62}}\\
\hline
\end{tabular}
\end{center}
\end{sidewaystable}

\begin{figure}
\centering  
\subfloat[\label{fig:NSCI_Wine}]{\includegraphics[width=0.33\textwidth]{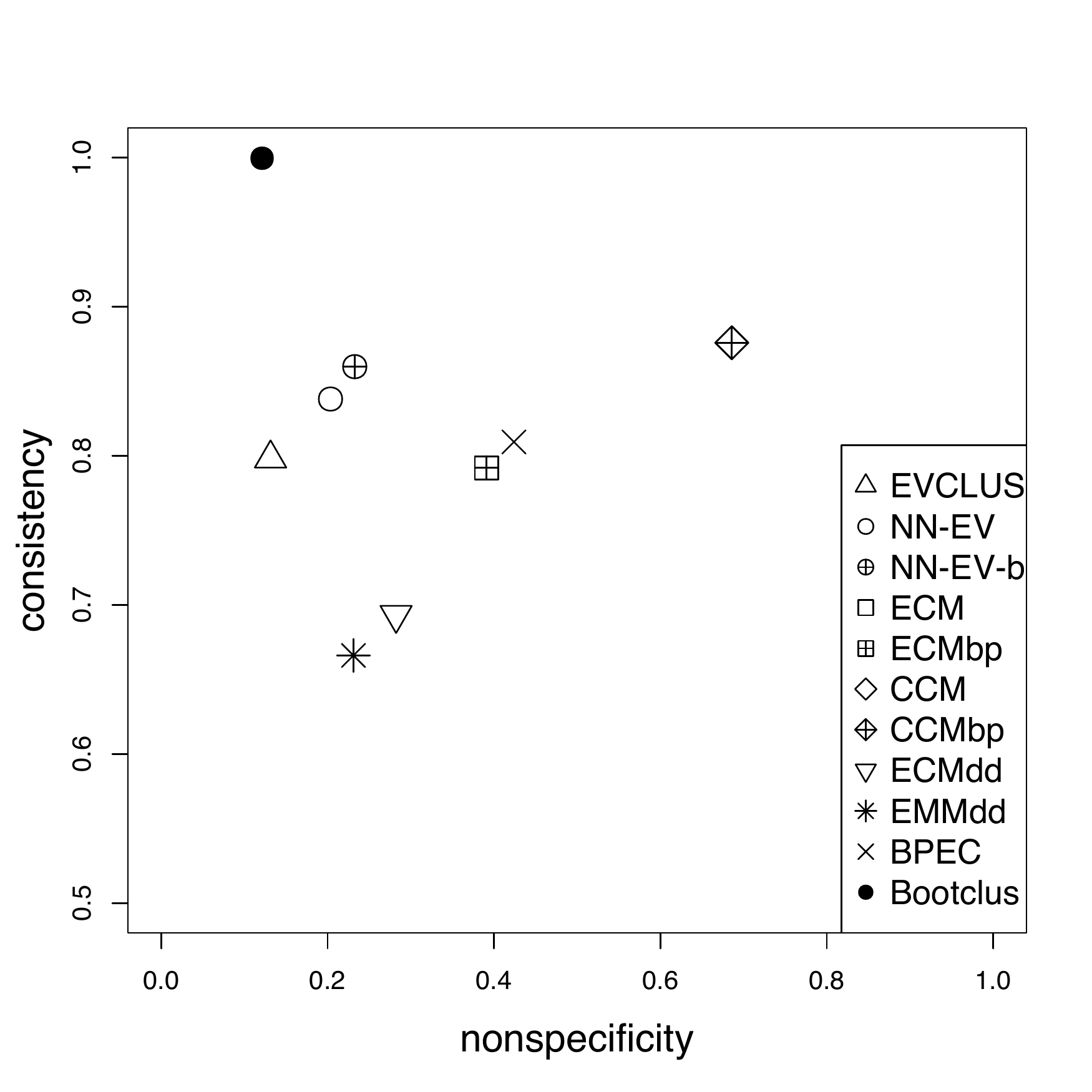}}
\subfloat[\label{fig:NSCI_Iris}]{\includegraphics[width=0.33\textwidth]{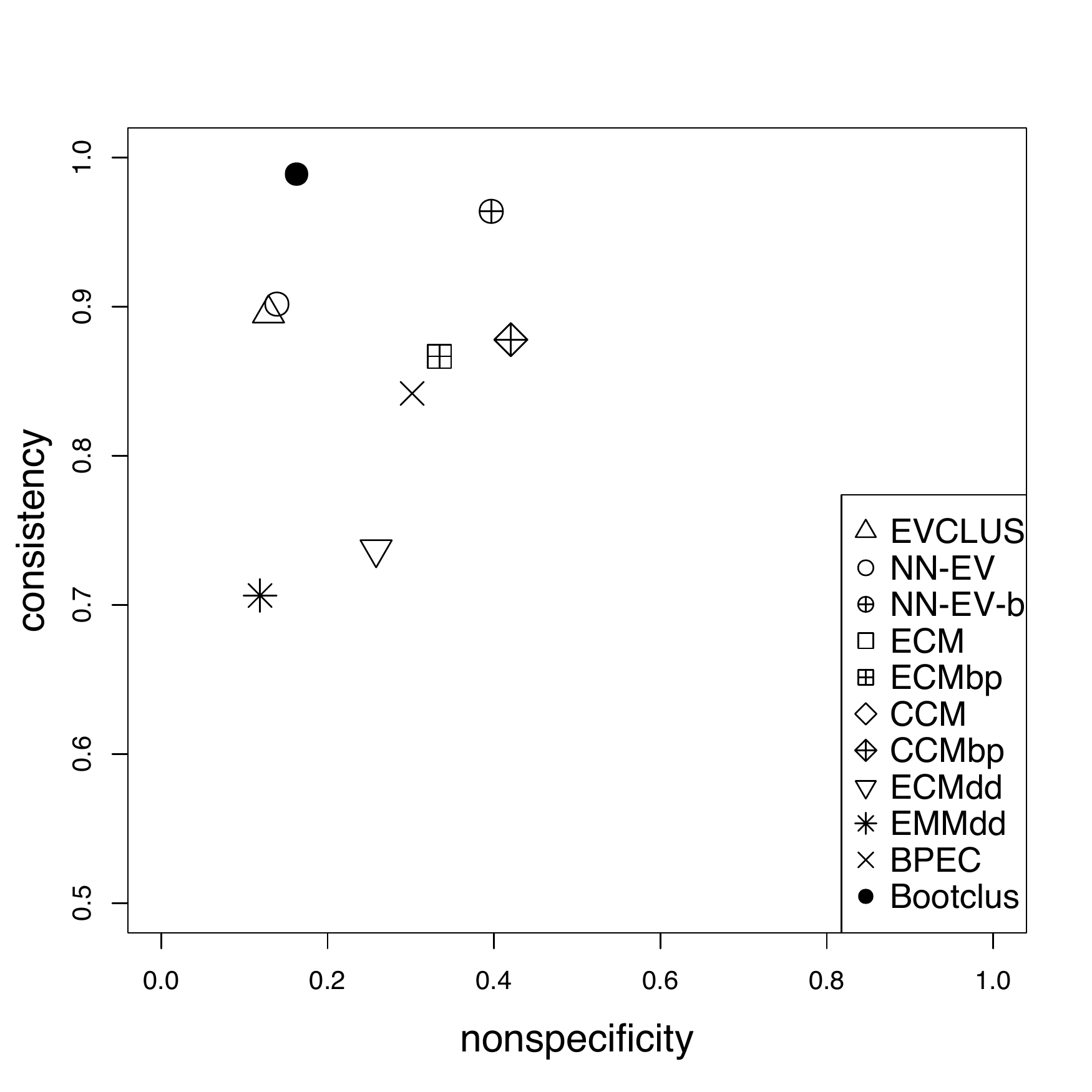}}
\subfloat[\label{fig:NSCI_Ecoli}]{\includegraphics[width=0.33\textwidth]{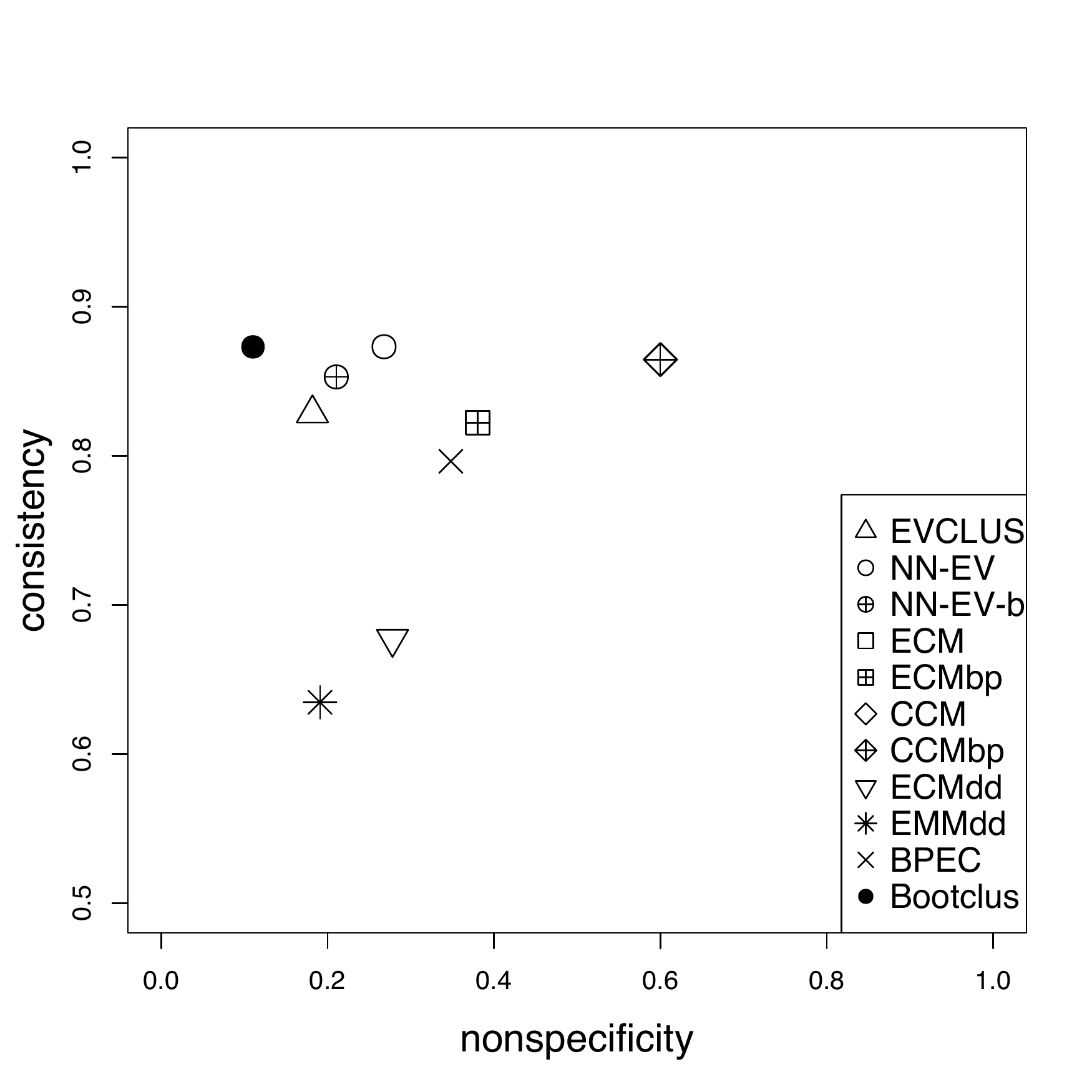}}\\
\subfloat[\label{fig:NSCI_Heart}]{\includegraphics[width=0.33\textwidth]{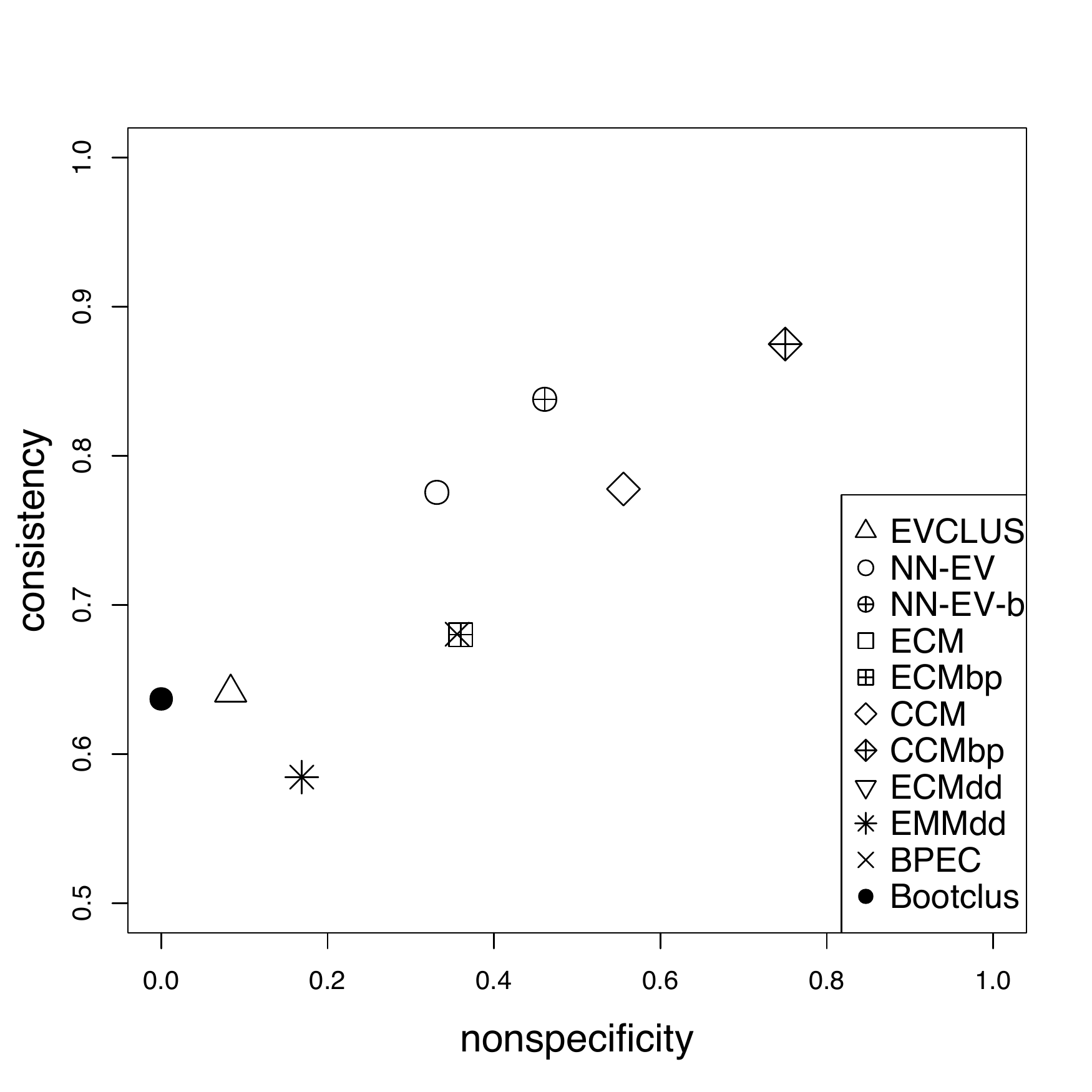}}
\subfloat[\label{fig:NSCI_Seeds}]{\includegraphics[width=0.33\textwidth]{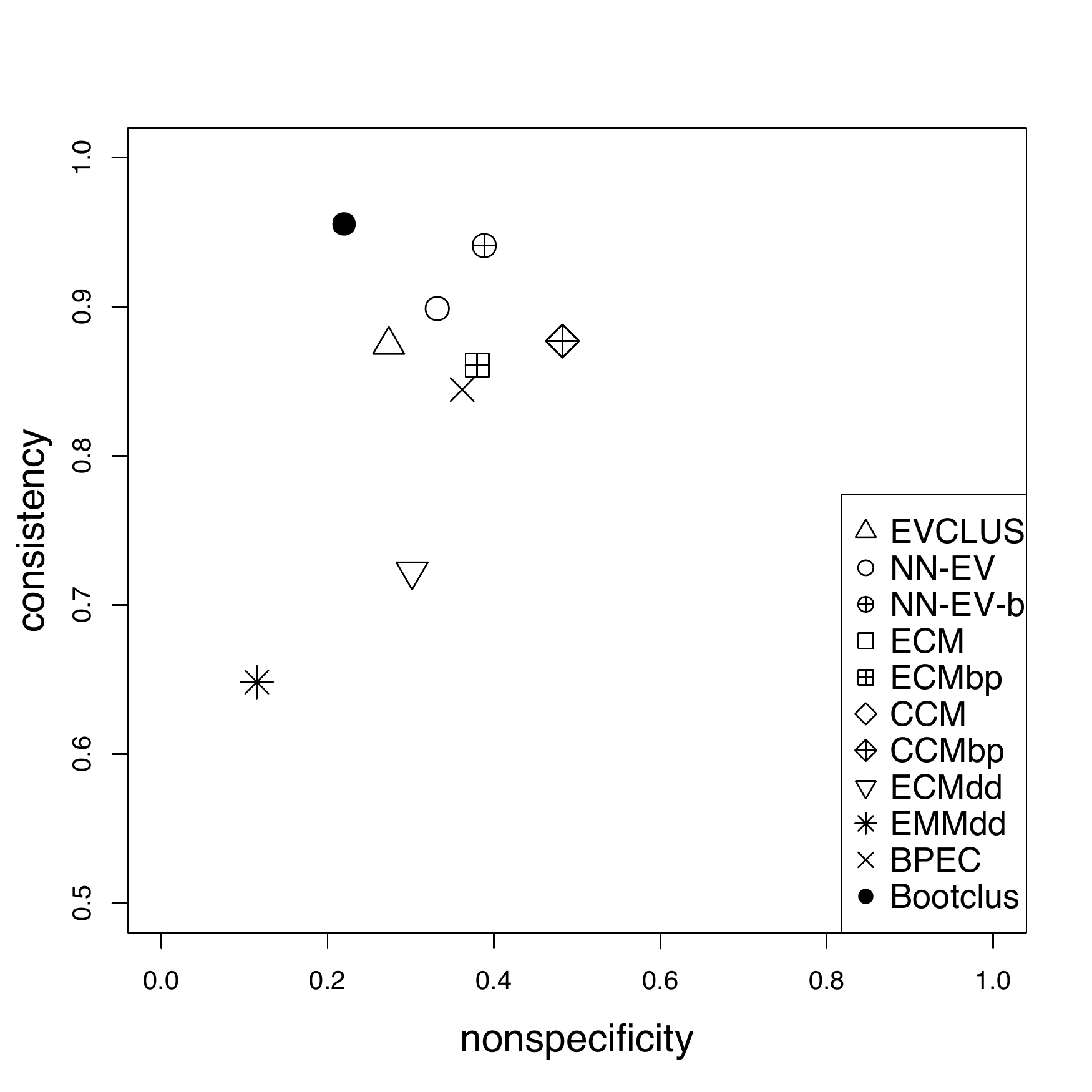}}
\subfloat[\label{fig:NSCI_Letters}]{\includegraphics[width=0.33\textwidth]{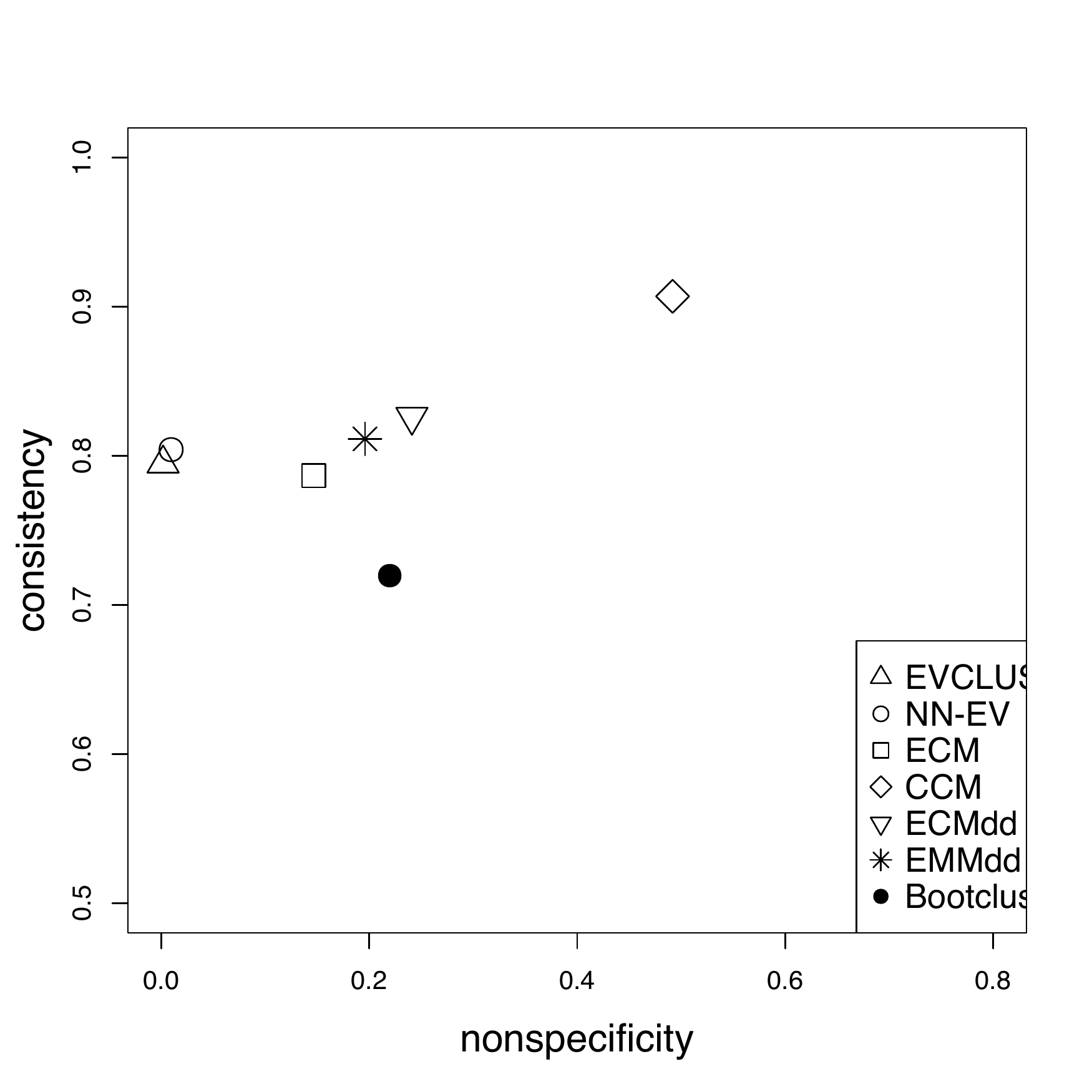}}
\caption{Consistency index (vertical axis) vs. nonspecificity (horizontal axis) for the  \textsf{Wine} (a), \textsf{Iris} (b), \textsf{Ecoli} (c), \textsf{Heart} (d), \textsf{Seeds} (e) and \textsf{Letters4p1} (f) datasets. \label{fig:NSCI1}}
\end{figure}

\begin{figure}
\centering  
\subfloat[\label{fig:NSCI_Glass}]{\includegraphics[width=0.33\textwidth]{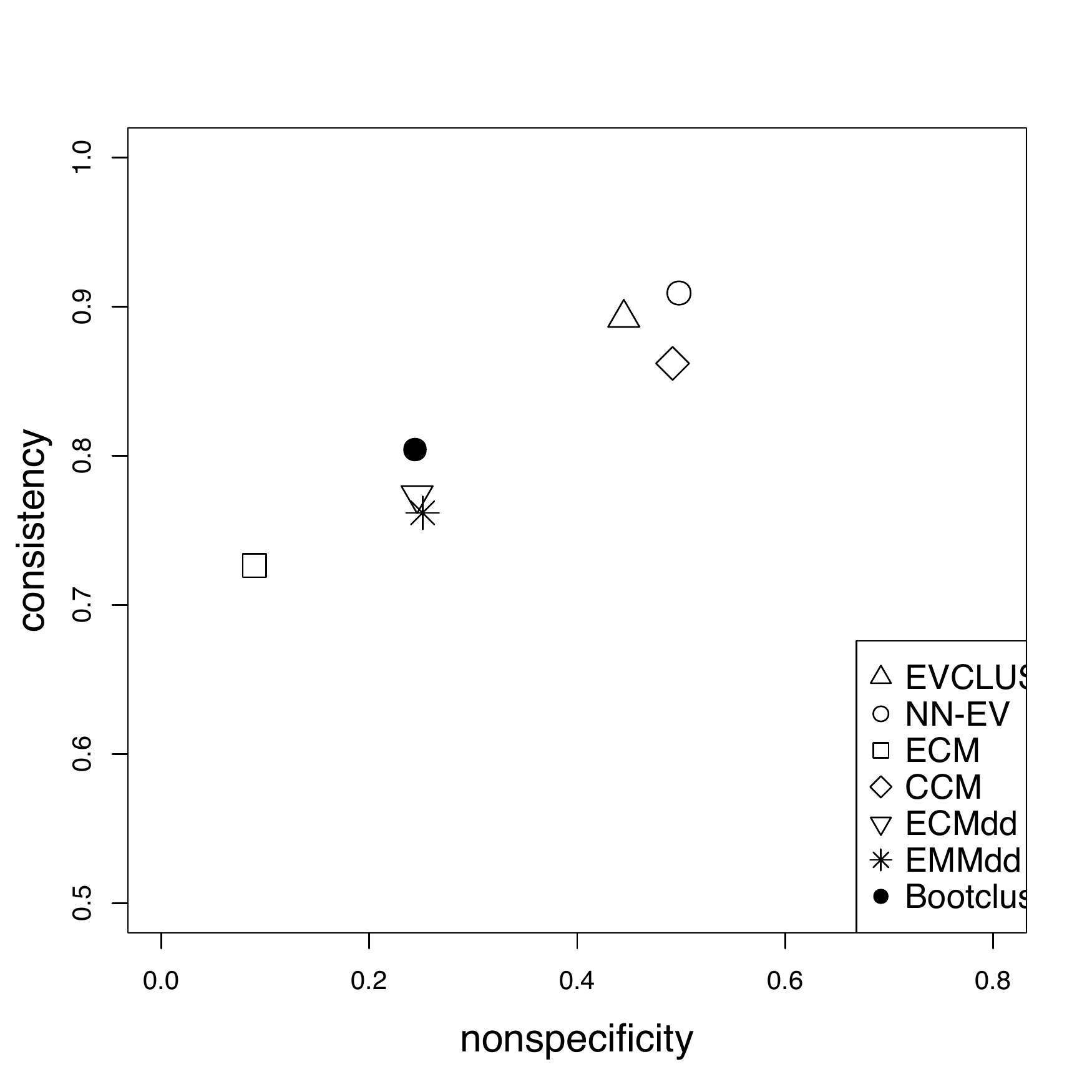}}
\subfloat[\label{fig:NSCI_Segment}]{\includegraphics[width=0.33\textwidth]{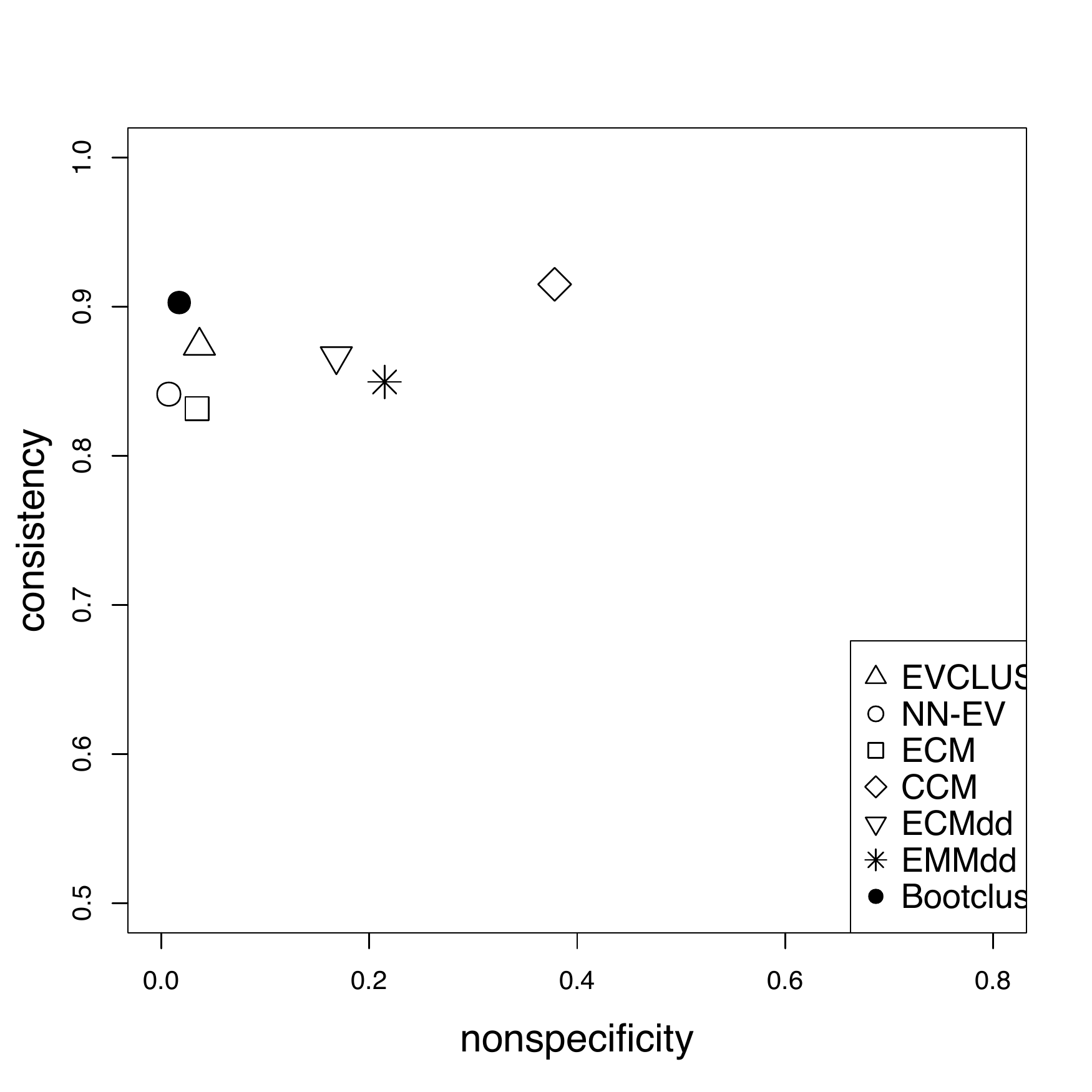}}
\subfloat[\label{fig:NSCI_S2}]{\includegraphics[width=0.33\textwidth]{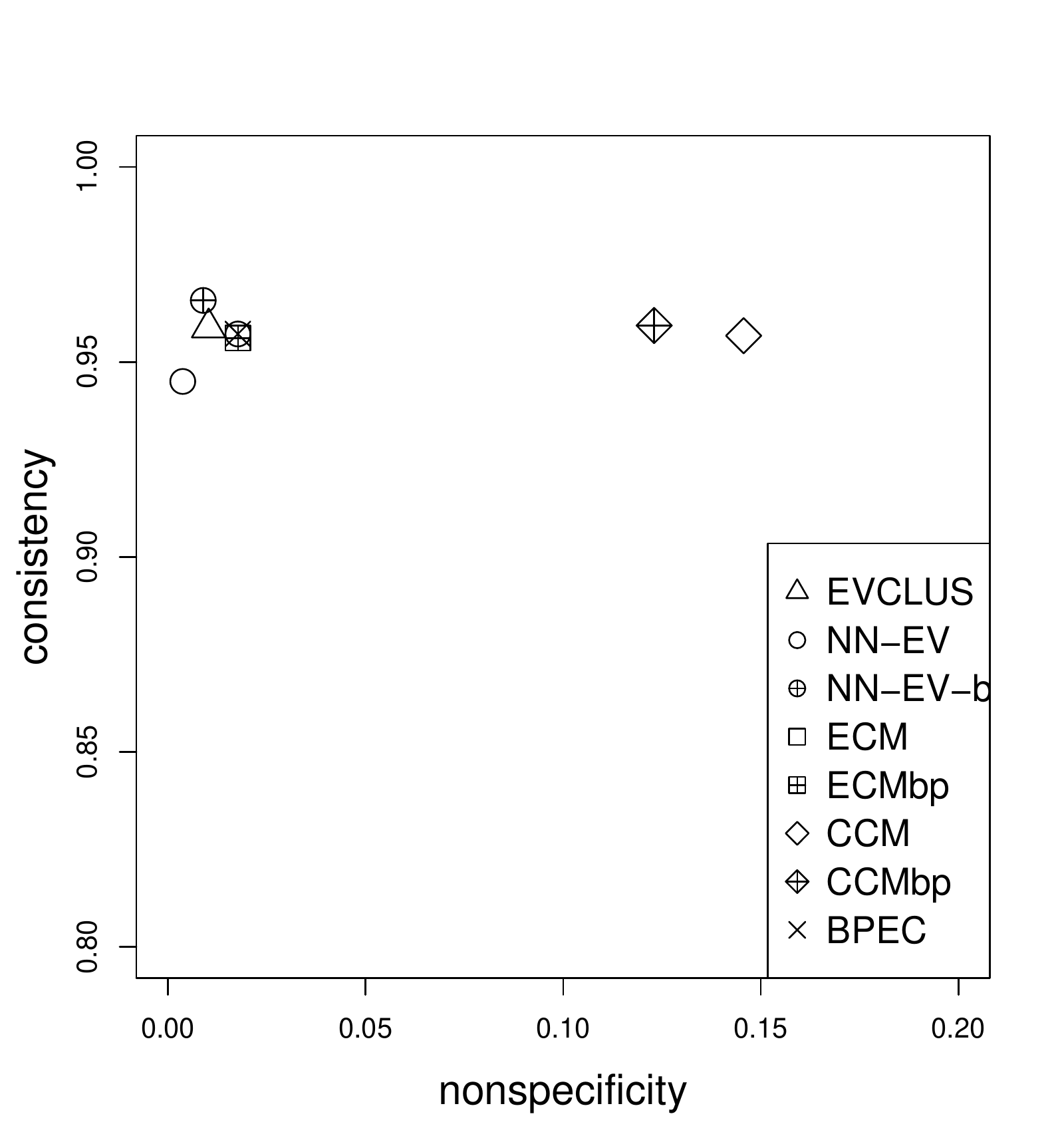}}\\
\subfloat[\label{fig:NSCI_S4}]{\includegraphics[width=0.33\textwidth]{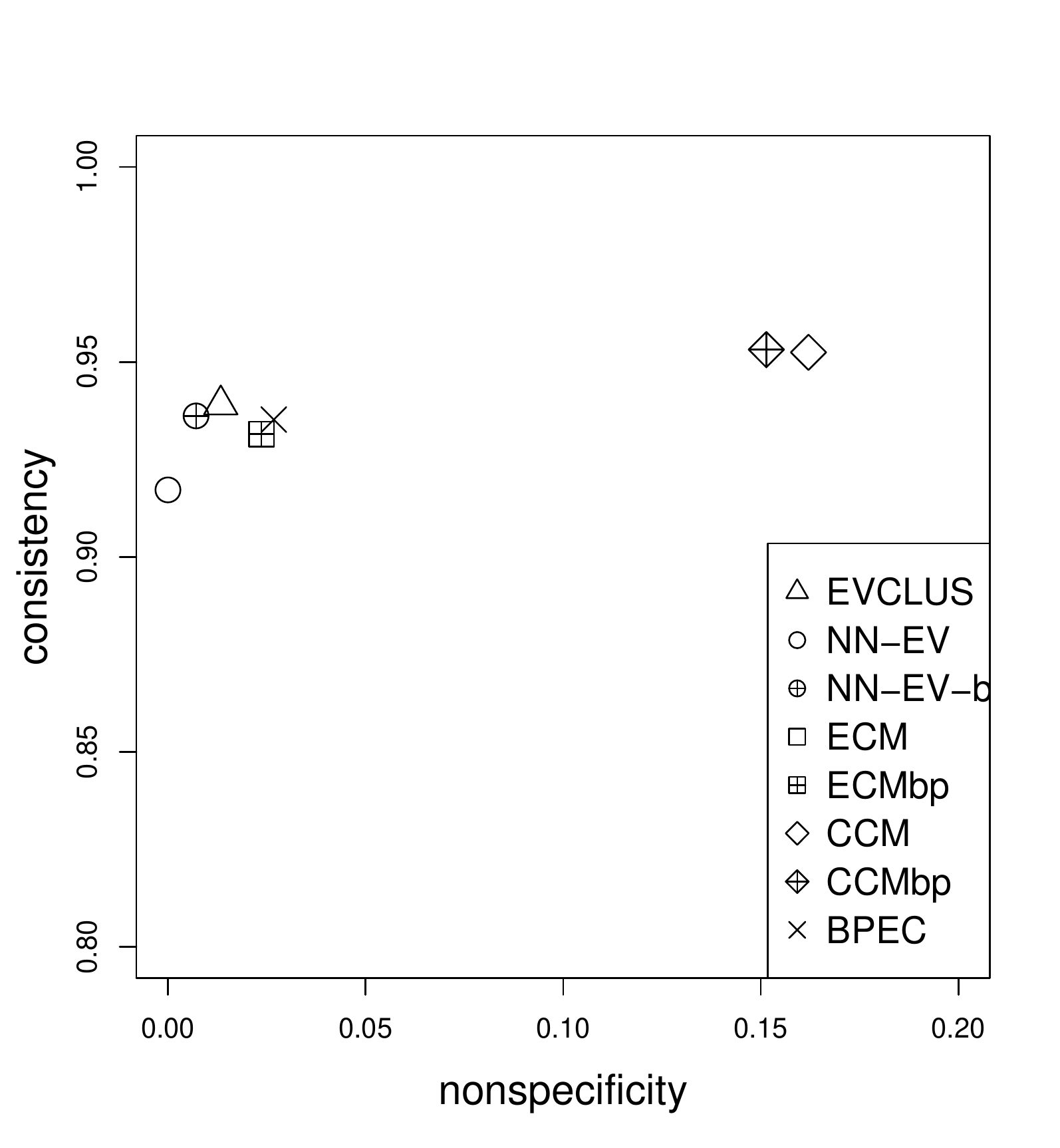}}
\subfloat[\label{fig:NSCI_D31}]{\includegraphics[width=0.33\textwidth]{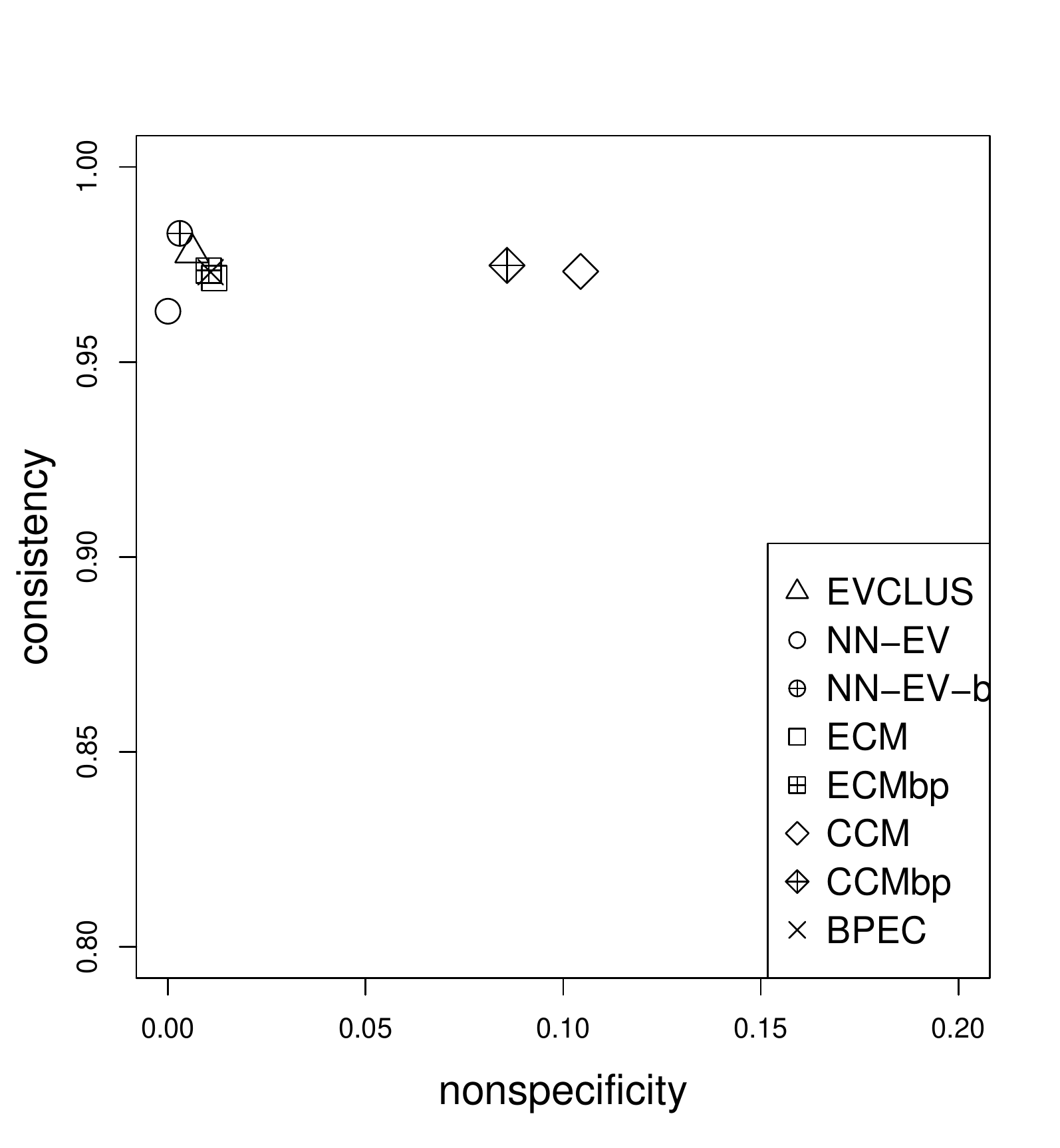}}
\caption{Consistency index (vertical axis) vs. nonspecificity (horizontal axis) for the  \textsf{Glass} (a), \textsf{Segment} (b), \textsf{S2} (c), \textsf{S4} (d) and \textsf{D31} (e) datasets. \label{fig:NSCI2}}
\end{figure}

\begin{figure}
\centering  
\subfloat[\label{fig:NSCI_Mice}]{\includegraphics[width=0.33\textwidth]{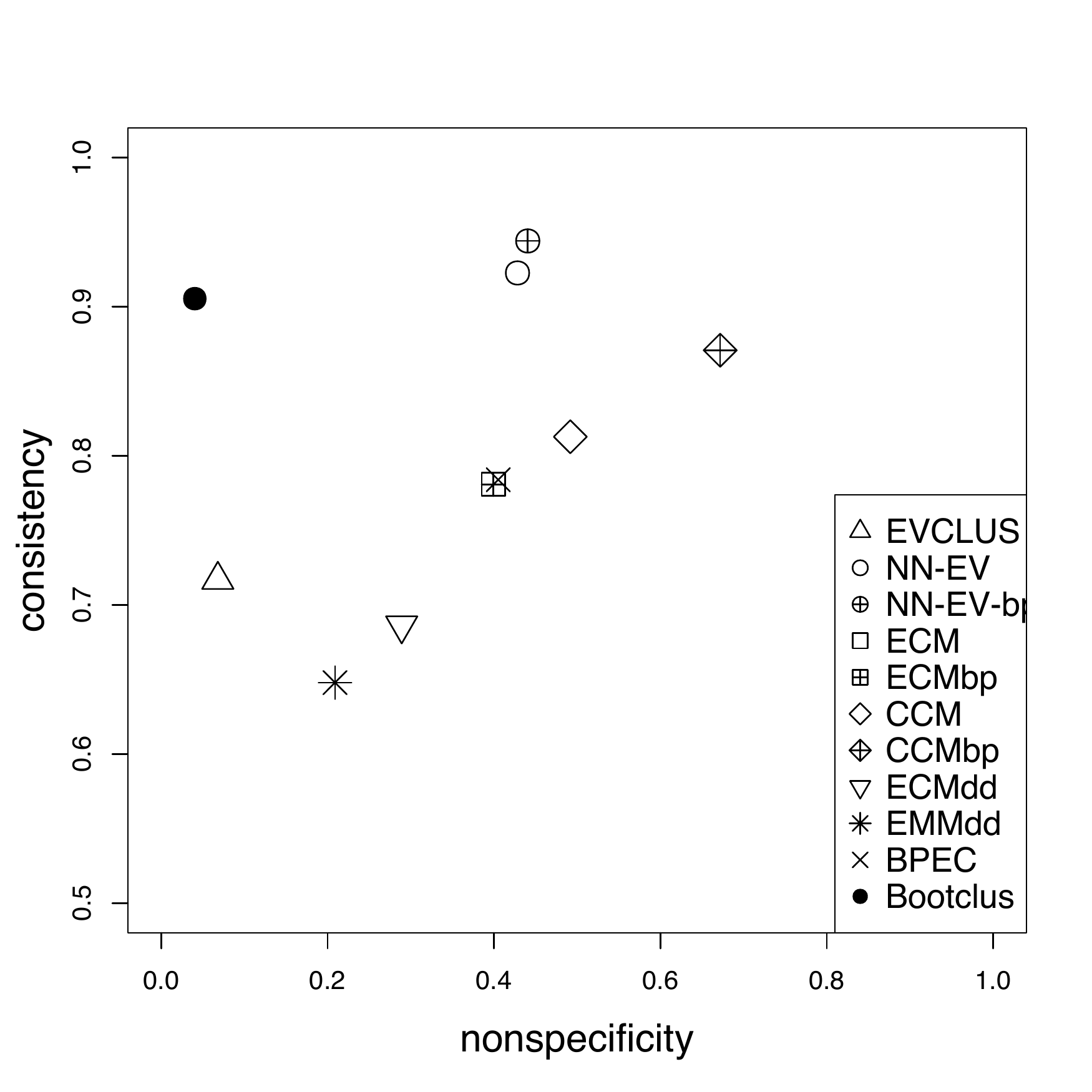}}
\subfloat[\label{fig:NSCI_Beans}]{\includegraphics[width=0.33\textwidth]{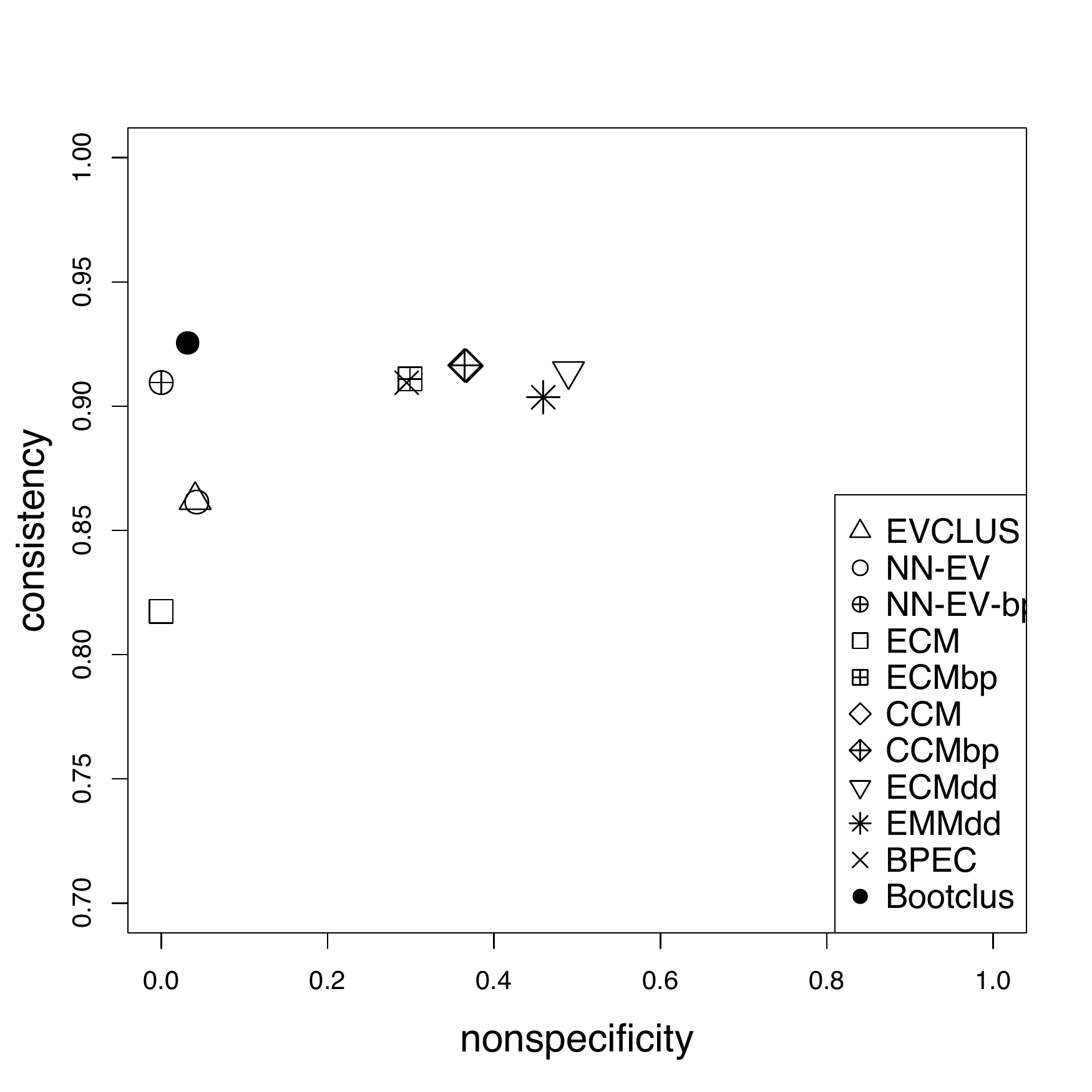}}
\subfloat[\label{fig:NSCI_Leaves}]{\includegraphics[width=0.33\textwidth]{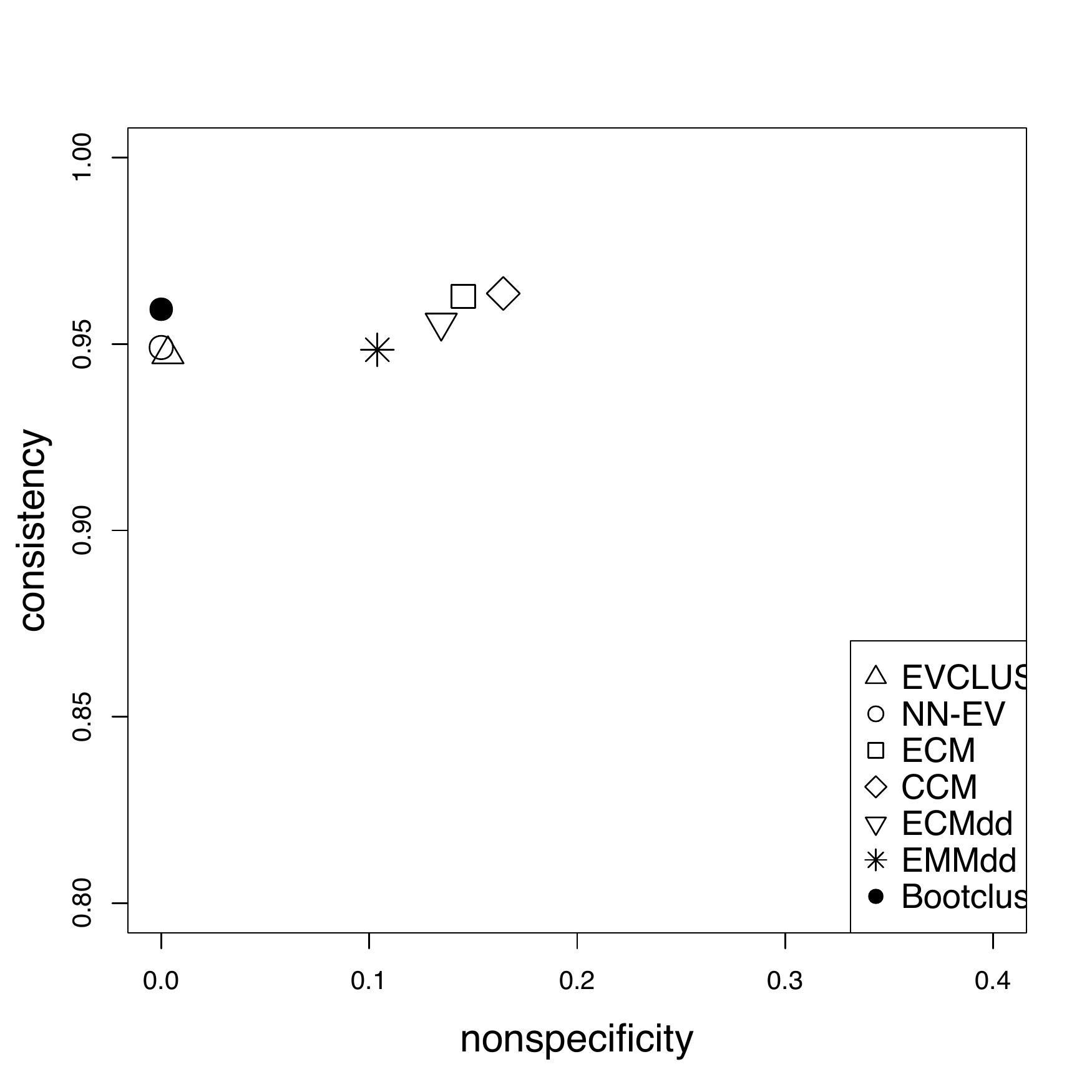}}
\caption{\new{Consistency index (vertical axis) vs. nonspecificity (horizontal axis) for the  \textsf{Mice} (a), \textsf{DryBean} (b),  and \textsf{Leaves5p1} (c) datasets.} \label{fig:NSCI2bis}}
\end{figure}

Figures \ref{fig:NSCI1}-\ref{fig:NSCI2bis} display the consistency indices and nonspecificities of the evidential partitions generated for the 14 datasets. These graphs allow us to visualize the  dominance relations between evidential partitions. \new{For instance, in Figure \ref{fig:NSCI_Heart} related to the \textsf{Heart} data, we can see that the evidential partitions generated by  Bootclus and EVCLUS dominate that generated by EMMdd, and that the evidential partition generated by CCM is dominated by that generated by EVCLUS-bp; the evidential partitions generated by Bootclus, EVCLUS, NN-EVCLUS, NN-EVCLUS-bp,  and CCM-bp are not dominated. From Figures \ref{fig:NSCI1}-\ref{fig:NSCI2bis}, we can see that the evidential partitions generated by NN-EVCLUS and NN-EVCLUS-bp are generally not dominated by those obtained by any of the others algorithms, except Bootclus on the \textsf{Wine} (Figure \ref{fig:NSCI_Wine}), \textsf{Ecoli} (Figure \ref{fig:NSCI_Ecoli}), \textsf{Seeds} (Figure \ref{fig:NSCI_Seeds}), and \textsf{Leaves5p1} (Figure \ref{fig:NSCI_Leaves}) datasets. In contrast, NN-EVCLUS dominates Bootclus on the  \textsf{Letters4p1} data (Figure \ref{fig:NSCI_Letters}). These results confirm the very good performance of NN-EVCLUS, which is only outperformed by Bootclus on a minority of datasets. }

\paragraph{Computing time} Computing time is an important issue when clustering large datasets. It is not so easy to measure intrinsically because it depends on the implementation of the algorithms. Table \ref{tab:time} reports the mean and standard deviations of the CPU times (in seconds) over five runs of EVCLUS, NN-EVCLUS, ECM and CCM applied to two of the largest datasets studied in this section: \textsf{S4} and \textsf{D31}. The algorithms were coded in R and executed on a 2019 16"   MacBook Pro with a 2.4 GHz 8-core Intel i9 processor. EVCLUS was run with $p=500$ and with the empty set, the singletons and $\Omega$ as focal sets. For NN-EVCLUS, we used the same focal sets as with EVCLUS and 30 hidden units; the neural network was randomly initialized and trained using the RMSprop algorithm with $s=30$ mini-batches and coefficients $\epsilon=0.001$, $\rho=0.9$ and $\delta=10^{-8}$. For ECM, we used the two-step procedure recalled above (the model was first trained with the empty set and singletons, and was then re-trained  after including  selected pairs of clusters as focal sets). For CCM, we used the empty set, the singletons, the pairs and $\Omega$ as focal sets, but the coefficient $T_c$ limiting the cardinality of the focal set was set to 2. We note that we included neither sECMdd nor wECMdd in this comparison as these algorithms require to store the whole dissimilarity matrix and they are extremely slow when applied to dataset of more than 1000 objects. We can see that NN-EVCLUS consumes significantly more time than EVCLUS  and, to an even greater extent, ECM. The CCM algorithm was very slow on the \textsf{D31} dataset because our implementation requires to use all pairs of clusters as focal sets. We can remark that  the execution of NN-EVCLUS could be dramatically accelerated by running the code on GPU, which is left for further development.

\begin{table}
\caption{Means and standard deviations (in parentheses) of CPU times (in seconds) over five runs of EVCLUS,  NN-EVCLUS,  ECM and CCM applied to the \textsf{S4} and \textsf{D31} datasets (see implementation details in the text). \label{tab:time}}
\begin{center}
\begin{tabular}{lcccc}
\hline
& EVCLUS & NN-EVCLUS & ECM & CCM\\
\hline
\textsf{S4} & 55.75 (11.2) &430.02 (21.5) &10.51 (3.3) &248.09 (64.7)\\
\textsf{D31} & 94.88  (19.8) & 400.60 (82.7) &   46.08 (10.6) & 1728.82 ( 430.5) \\
\hline
\end{tabular}
\end{center}
\end{table}

\subsection{Unsupervised clustering of dissimilarity data}
\label{subsec:rel}

Whereas the EVCLUS algorithm was initially introduced for clustering dissimilarity data \cite{denoeux04b,denoeux16a}, it may seem that this possibility is lost with NN-EVCLUS, which uses attributes as input, in addition to a distance of dissimilarity matrix. However, it is still possible to cluster dissimilarity data with NN-EVCLUS by using dissimilarities  as attributes. More precisely, let $\bD=(\delta_{ij})$ be the $n \times n$ dissimilarity matrix. Each object $i$ can be described by the $n$-dimensional vector $\bdelta_i=(\delta_{i,1},\ldots,\delta_{i,n})$ of its distances to the $n$ objects (including itself), which corresponds to a row of matrix $\bD$ and can be regarded as a vector of $n$ attributes. To reduce the dimensionality of the representation, we can apply Principal Component Analysis (PCA) to these vectors and project the data on the  subspace spanned by the $p$ first principal components, resulting in the description of each object $i$ by a $p$-dimensional attribute vector $\bx_i$. We note that  the attributes of any new object can be obtained from its  dissimilarities to the $n$ objects in the learning set by multiplying the centered vector of dissimilarities by the projection matrix.

\paragraph{Datasets} To study the application of NN-EVCLUS to nonmetric dissimilarity data, we considered four datasets:
\be
\item The \textsf{Protein} dataset\footnote{This dataset is part of  the {\tt evclust} R package \cite{denoeux21}.} \cite{hofmann97,graepel99,denoeux04b} consists of a dissimilarity matrix derived from the structural comparison of 213 protein sequences. Each of these proteins is known to belong to one of four classes of globins: hemoglobin-$\alpha$ (HA), hemoglobin-$\beta$ (HB), myoglobin (M) and heterogeneous globins (G).
\item The \textsf{ChickenPieces} dataset\footnote{The \textsf{ChickenPieces}, \textsf{Zongker} and \textsf{Gestures} datasets are available at \url{http://prtools.org/disdatasets}.} is composed of dissimilarities between 446 binary images represents the silhouettes of five parts of  chickens. There are thus $c=5$ clusters. The dataset is composed of  44 dissimilarity matrices corresponding to different ways of computing the dissimilarities. As in \cite{antoine12}, we used matrix {\tt chickenpieces-20-90} in our experiments. Since the data are slightly asymmetric, we computed a new matrix $\bD=(\delta_{ij})$ by the transformation $\delta_{ij}\leftarrow (\delta_{ij}+\delta_{ji})/2$.
\item The \textsf{Zongker} dataset contains similarities between 2000 handwritten digits in 10 classes, based on deformable template matching. The dissimilarity measure is the result of an iterative optimization of the non-linear deformation of the grid \cite{jain97}. Again, we made the dissimilarity matrix  symmetric  by the transformation $\delta_{ij}\leftarrow (\delta_{ij}+\delta_{ji})/2$.
\item The \textsf{Gestures}  dataset consists of the dissimilarities computed from a set of gestures in a sign-language study \cite{lichtenauer08}. They were measured by two video cameras observing the positions the two hands in 75 repetitions of creating 20 different signs. There are thus 1500 objects grouped in 20 clusters. The dissimilarities were computed by  a dynamic time warping procedure.  
\ee

\paragraph{Algorithms} As alternative evidential relational clustering algorithms, we considered EVCLUS \cite{denoeux16a}, RECM \cite{masson09a}, sECMdd and wECMdd \cite{zhou16}.  We used the same focal sets for EVCLUS, NN-EVCLUS and RECM (the empty set, the singletons and $\Omega$). For sECMdd and wECMdd, we used the empty set, the singletons, the pairs and $\Omega$. The other parameter values for each of these algorithms are summarized in Table \ref{tab:rel_param}. For RECM, sECMdd and wECMdd, we used the default settings recommended by the authors  \cite{masson09a,zhou16}. \new{The batch version of NN-EVCLUS was run for the \textsf{Protein} dataset, and the mini-batch version (with 10 mini-batches and the RMSprop algorithm) was applied to the three other datasets.} Each algorithm was run five times, and the best solution in terms of the loss or objective function was retained.

\begin{sidewaystable}
\caption{Parameter values for the five methods applied the three dissimilarity datasets. The notation $\delta_{(\alpha)}$ stands for the $\alpha$-quantile of dissimilarities. \label{tab:rel_param}}
\begin{center}
\begin{tabular}{|l|p{2.5cm}|p{3cm}|p{3cm}|p{3.5cm}|p{3.5cm}|}
\hline
& EVCLUS & NN-EVCLUS & RECM & sECMdd & wECMdd\\
\hline
\textsf{Protein} &  $\delta_0=\max \delta_{ij}$ &  $p=3$, $n_H=20$, $\delta_0=\max \delta_{ij}$ & $\alpha = 1$,
  $\beta = 1.5$, $\delta^2= \max \delta_{ij}$& $\alpha = 2$, $\beta = 2$, $\eta = 1$, $\gamma = 1$, $\delta= \max \delta_{ij}$&$\alpha = 2$, $\beta = 2$, $\xi = 5$, $\psi = 2$, $\delta= \max \delta_{ij}$\\
  \hline
\textsf{ChickenPieces} & $\delta_0=\delta_{(0.2)}$ & $p=5$, $n_H=30$, $\delta_0=\delta_{(0.2)}$ &$\alpha = 1$,
  $\beta = 1.5$, $\delta^2=\delta_{(0.95)}$ &$\alpha = 2$, $\beta = 2$, $\eta = 1$, $\gamma = 1$, $\delta= \delta_{(0.95)}$&  $\alpha = 2$, $\beta = 2$, $\xi = 5$, $\psi = 2$, $\delta= \delta_{(0.95)}$\\
  \hline
\textsf{Zongker} & $\delta_0=\delta_{(0.3)}$  & $p=20$, $n_H=50$, $\delta_0=\delta_{(0.3)}$ &$\alpha = 1$,
  $\beta = 1.5$, $\delta^2=\delta_{(0.95)}$ &$\alpha = 2$, $\beta = 2$, $\eta = 1$, $\gamma = 1$, $\delta= \delta_{(0.95)}$ &$\alpha = 2$, $\beta = 2$, $\xi = 5$, $\psi = 2$, $\delta= \delta_{(0.95)}$\\
  \hline
 \textsf{Gestures} & $\delta_0=\delta_{(0.2)}$  & $p=20$, $n_H=50$, $\delta_0=\delta_{(0.2)}$ &$\alpha = 1$,
  $\beta = 1.5$, $\delta^2=\delta_{(0.95)}$ &$\alpha = 2$, $\beta = 2$, $\eta = 1$, $\gamma = 1$, $\delta= \delta_{(0.95)}$ &$\alpha = 2$, $\beta = 2$, $\xi = 5$, $\psi = 2$, $\delta= \delta_{(0.95)}$\\
\hline
\end{tabular}
\end{center}
\end{sidewaystable}

\begin{figure}
\centering  
\subfloat[\label{fig:NSCI_Protein}]{\includegraphics[width=0.33\textwidth]{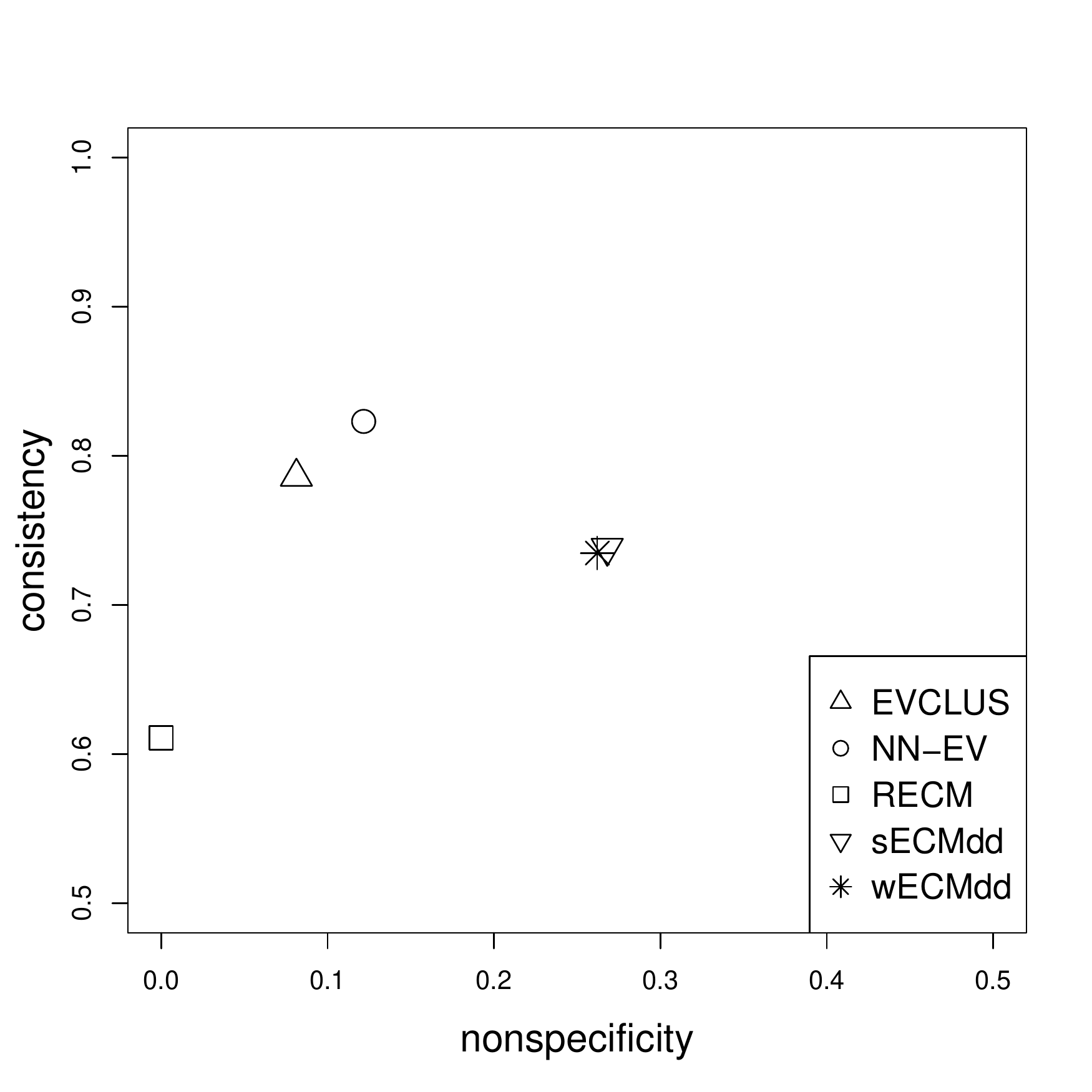}}
\subfloat[\label{fig:NSCI_Chicken}]{\includegraphics[width=0.33\textwidth]{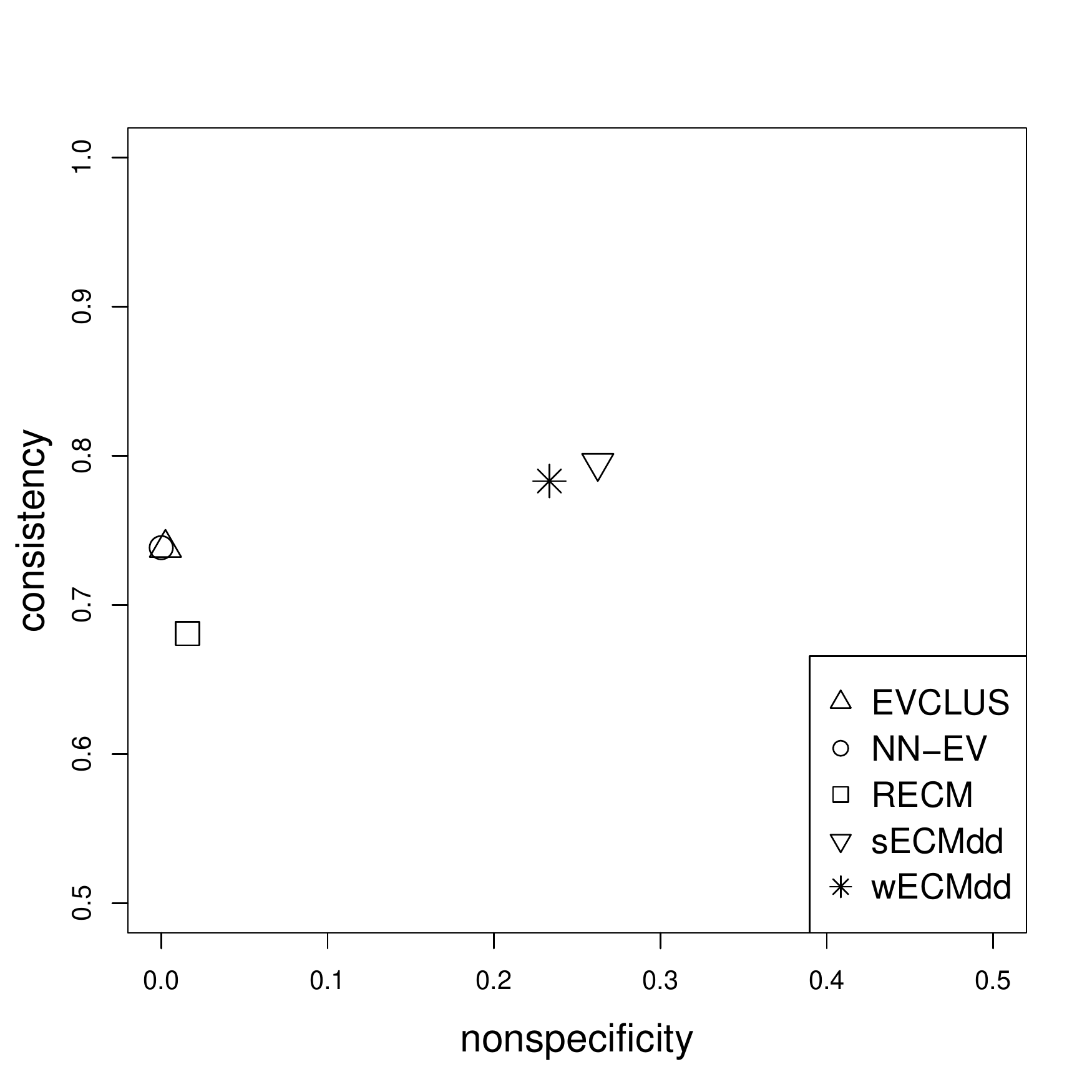}}\\
\subfloat[\label{fig:NSCI_Zongker}]{\includegraphics[width=0.33\textwidth]{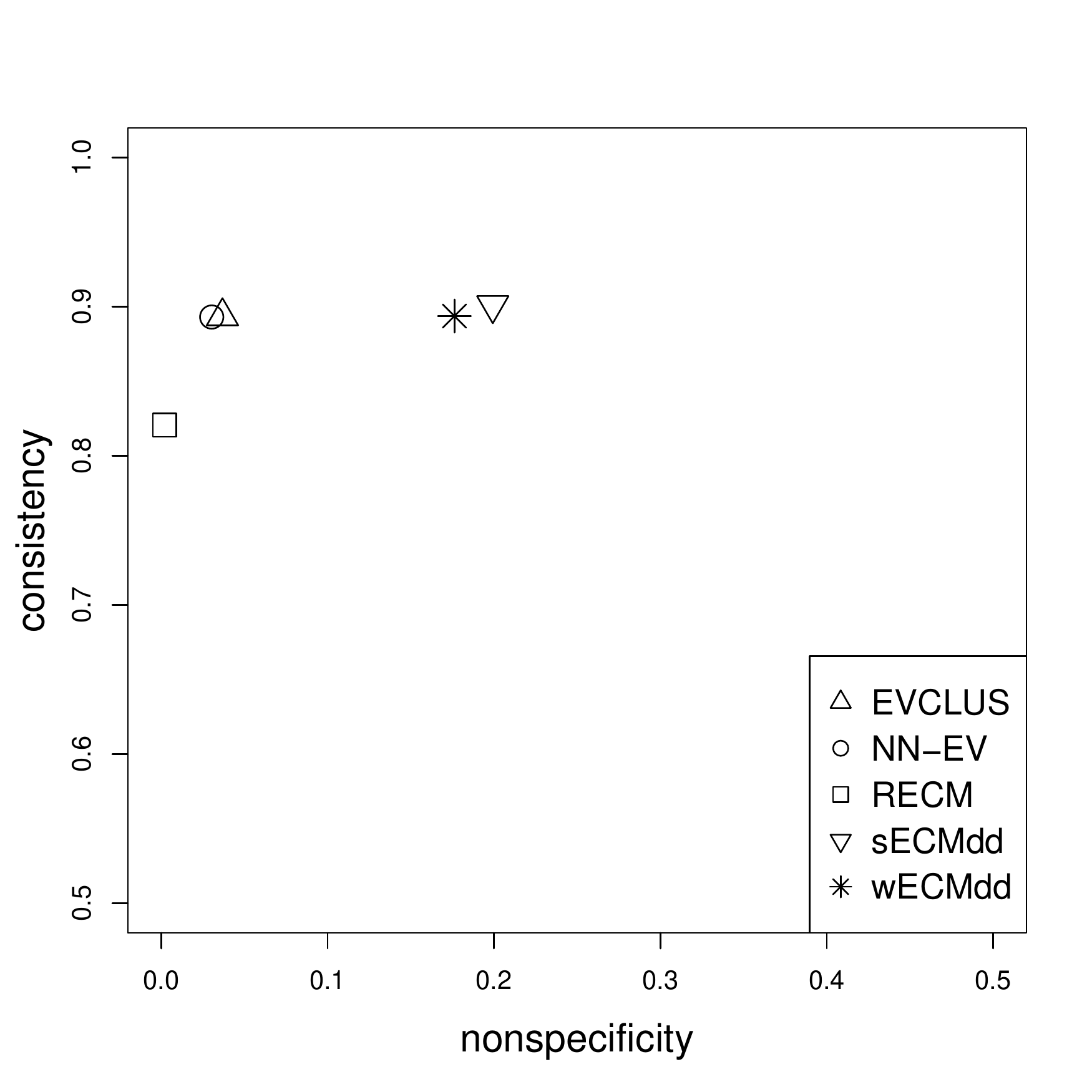}}
\subfloat[\label{fig:NSCI_Gestures}]{\includegraphics[width=0.33\textwidth]{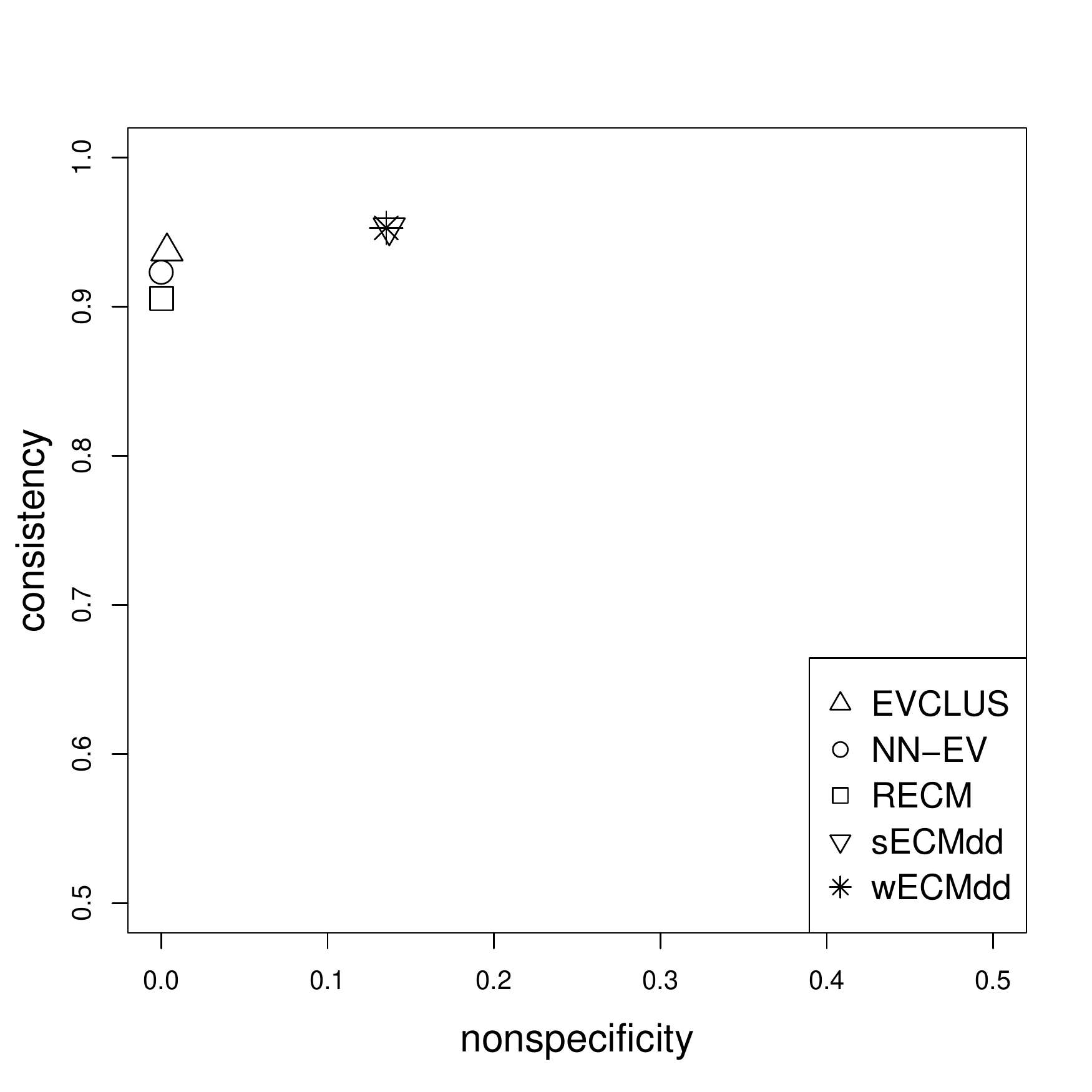}}
\caption{Consistency index (vertical axis) vs. nonspecificity (horizontal axis) for the  \textsf{Protein} (a), \textsf{ChichenPieces} (b),   \textsf{Zongker} (c) and \textsf{Gestures} (d)  datasets. \label{fig:NSCI3}}
\end{figure}

\paragraph{Results} The performances of the five methods in terms of ARI are shown in Table \ref{tab:ARI_dissi_data}. As we can see, EVCLUS and NN-EVCLUS outperform the other methods for the four datasets,  NN-EVCLUS reaching slightly higher values of ARI for the \textsf{ChickenPieces}, \textsf{Zongker} and \textsf{Gestures} datasets. As before, we also display the nonspecificity and consistency indices of the obtained evidential partitions for the four datasets in Figure \ref{fig:NSCI3}. We can see that sECMdd and wECMdd produce less specific evidential partitions, due to the selection of pairs as focal sets. The evidential partitions produced by  EVCLUS and NN-EVCLUS  strictly dominate those obtained by ECMdd and wECMdd  for the \textsf{Protein} dataset (Figure \ref{fig:NSCI_Protein}), and the one produced by RECM for the \textsf{ChichenPieces} dataset (Figure \ref{fig:NSCI_Chicken}). The evidential partitions produced by NN-EVCLUS are always nondominated.

\begin{table}
\caption{ARI values for the five methods on the four dissimilarity datasets. The best value for each dataset is printed in bold, and the values within 5\% of the best value are underlined. \label{tab:ARI_dissi_data}}
\begin{center}
\begin{tabular}{lccccc}
\hline
& EVCLUS & NN-EVCLUS & RECM & sECMdd & wECMdd\\
\hline
\textsf{Protein} & \bf{0.989} &\bf{0.989} &0.863 &0.402 &0.246\\
\textsf{ChickenPieces} & \underline{0.308} &\bf{0.315} &0.251 &0.073& 0.203\\
\textsf{Zongker} & \underline{0.791} & \bf{0.803} &0.217 &0.053 &0.053\\
\textsf{Gestures} & \underline{0.709} & \bf{0.710} & 0.096 & 0.183 &0.095\\
\hline
\end{tabular}
\end{center}
\end{table}

\new{\paragraph{Computing time} The mean and standard deviations of the CPU times (in seconds) over five runs of EVCLUS, NN-EVCLUS, RECM, sECMdd and wECMdd applied to the  \textsf{Protein},  \textsf{ChickenPieces}, \textsf{Zongker} and \textsf{Gestures} datasets are reported in Table \ref{tab:time_rel}. Again, the algorithms were coded in R and executed on a 2019 16"   MacBook Pro with a 2.4 GHz 8-core Intel i9 processor. These times are only indicative because they obviously depend on  implementation. The settings were those described in Table  \ref{tab:rel_param}. The RECM procedure is the fastest of all four algorithms, but it performs poorly on the fours datasets considered in the eperiment. In contrast,  the good performances of NN-EVCLUS come at the price of a higher computing time. However, as already emphasized, the NN-EVCLUS lends itself to parallel implementation, which would speed it up by a potentially large factor.  }

\begin{table}
\caption{\new{Means and standard deviations (in parentheses) of CPU times (in seconds) over five runs of EVCLUS,  NN-EVCLUS,  RECM, sECMdd and wECMdd applied to the \textsf{Protein},  \textsf{ChickenPieces}, \textsf{Zongker} and \textsf{Gestures} datasets (see implementation details in the text). }\label{tab:time_rel}}
\new{
\begin{center}
\begin{tabular}{lccccc}
\hline
& EVCLUS & NN-EVCLUS & RECM & sECMdd & wECMdd\\
\hline
\textsf{Protein} & 2.48 (0.2) & 36.26  (3.4) & 0.11 (0.02) &  0.30  (0.04) & 0.37 (0.07)\\
\textsf{ChickenPieces} & 3.16 (0.3) & 16.72 (0.1) &  0.18 (0.02) &  1.30 (0.2) &  1.07 (0.05)\\
\textsf{Zongker} & 15.53 (4.9) & 251.54 (5.2) & 13.16 (0.3) &47.74 (0.8) &25.06 (0.5)\\
\textsf{Gestures} & 63.61 (21.4) &258.51 (4.7) & 5.12 (0.2) &134.60 (5.5) & 82.95 (1.7)\\
\hline
\end{tabular}
\end{center}
}
\end{table}

\paragraph{Prediction} Whereas EVCLUS and NN-EVCLUS yield similar results on these datasets, a distinctive advantage of NN-EVCLUS is that it makes it possible to predict the cluster membership of of new objects, without recomputing the evidential partition for the extended dataset. To demonstrate this possibility, we randomly split the \textsf{Zongker} dataset into two subsets of 1000 objects. We computed the attribute vectors by PCA for the first set of objects as explained above (with $p=20$), and we trained NN-EVCLUS  using the dissimilarity matrix for this first set. We then computed the attributes for the other 1000 objects and we computed the evidential partition by propagating the attribute values through the NN. The whole process was repeated 10 times. The average training and test ARI values were, respectively, 0.73 (standard deviation: 0.03) and 0.72 (standard deviation: 0.05). These results show that the relation between attributes and mass functions can be successfully learnt by  NN-EVCLUS and generalized to test data, making it possible to predict the cluster membership of new objects.

\subsection{Constrained clustering}
\label{subsec:exper_semi}

Finally, we also compared the performance of NN-EVCLUS for exploiting pairwise constraints (as described in Section \ref{subsec:semi}) to those of alternative constrained evidential clustering methods, namely: CEVCLUS \cite{li18} and CECM \cite{antoine12}. We considered the attribute dataset (\textsf{Glass}) and the dissimilarity dataset (\textsf{ChickenPieces}) with the lowest ARI values in their category (see, respectively,  Tables \ref{tab:ARI_attribute_data} and \ref{tab:ARI_dissi_data}). We also included the \textsf{Iris} dataset in the analysis as it is an example of a dataset with nonsperical clusters, for which pairwise constraints can significantly improve the clustering results. 

Each of the three clustering algorithms was used with and without metric adaptation through PCCA \cite{mignon12}. For the \textsf{Glass} and \textsf{Iris} data, we extracted, respectively,  five and three features using PCCA and we computed the Euclidean matrix in the feature space. The distance matrix was used by CEVCLUS and the features by CECM;  NN-EVCLUS used both. For the \textsf{ChickenPieces}, we first extracted five features from distances using PCA as explained in Section \ref{subsec:rel}, and we computed five new features using PCCA. For CEVCLUS and NN-EVCLUS, parameter $\xi$ in \eqref{eq:stress} was set to 0.5. For CECM, parameter $\xi$ controling the balance between the constraints
and the objective function was also set to 0.5. The other parameters were set as in the previous experiments reported in Sections \ref{subsec:attr}
and \ref{subsec:rel}.

\begin{figure}
\centering  
\subfloat[\label{fig:Glass_semi_50}]{\includegraphics[width=0.33\textwidth]{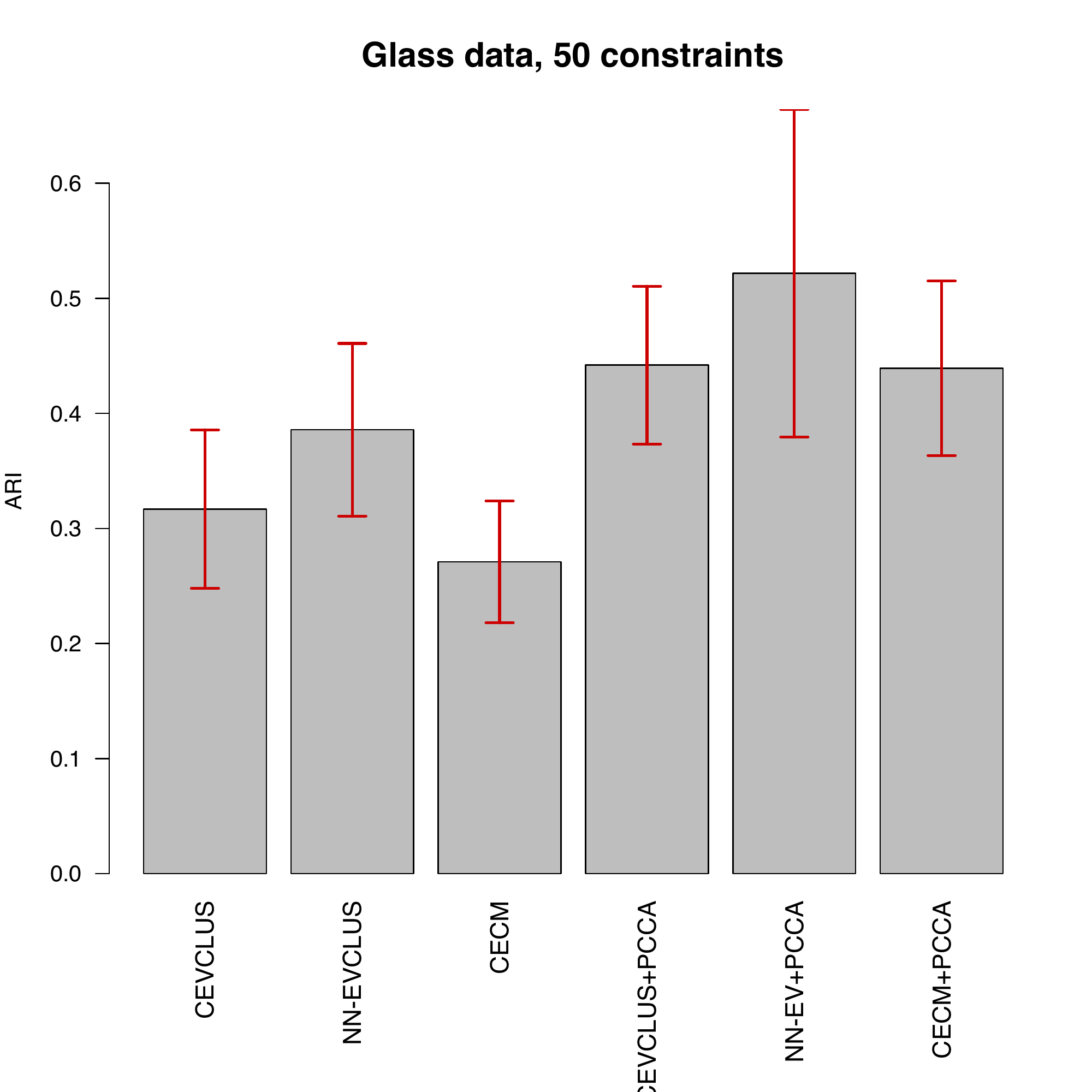}}
\subfloat[\label{fig:Glass_semi_100}]{\includegraphics[width=0.33\textwidth]{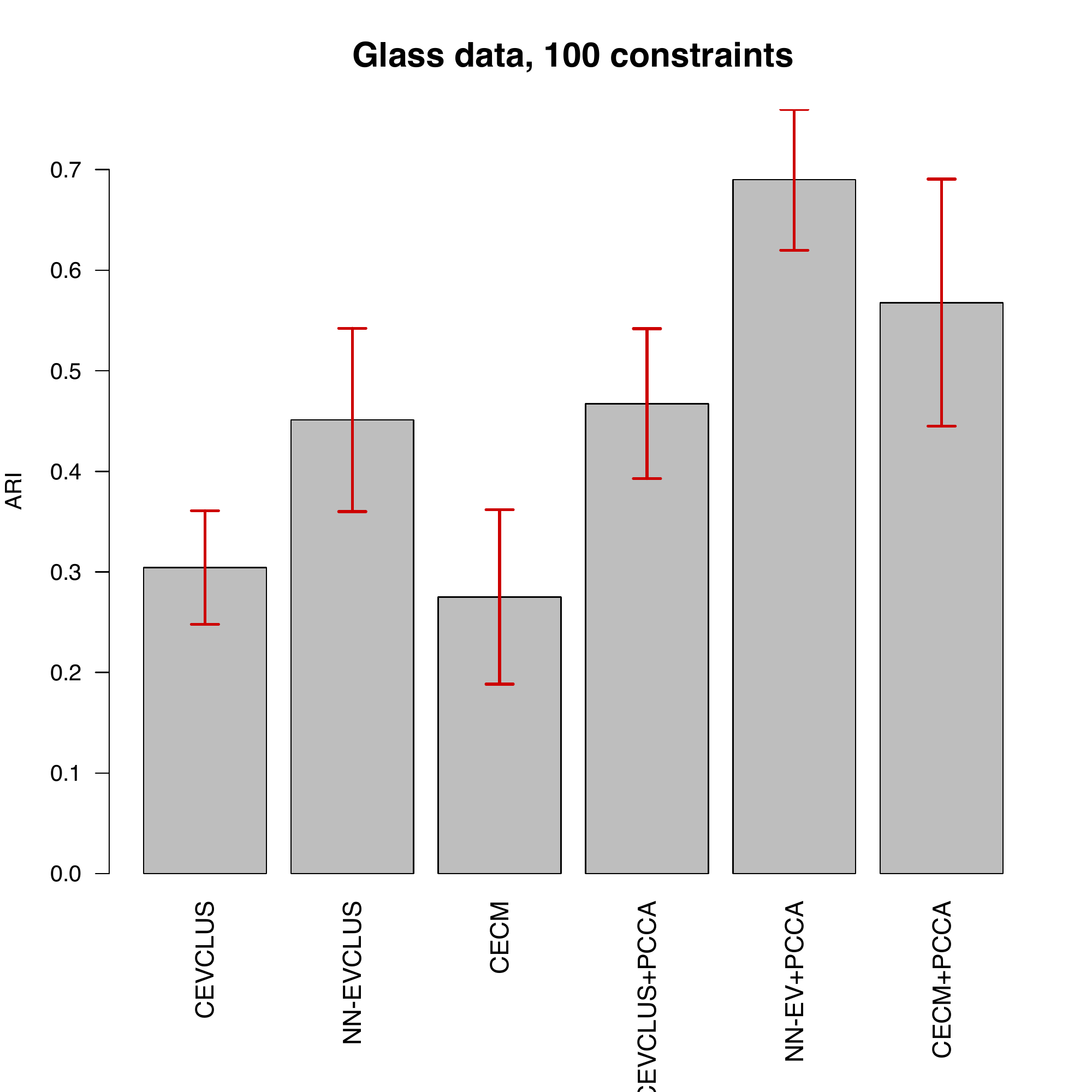}}\\
\subfloat[\label{fig:Glass_semi_150}]{\includegraphics[width=0.33\textwidth]{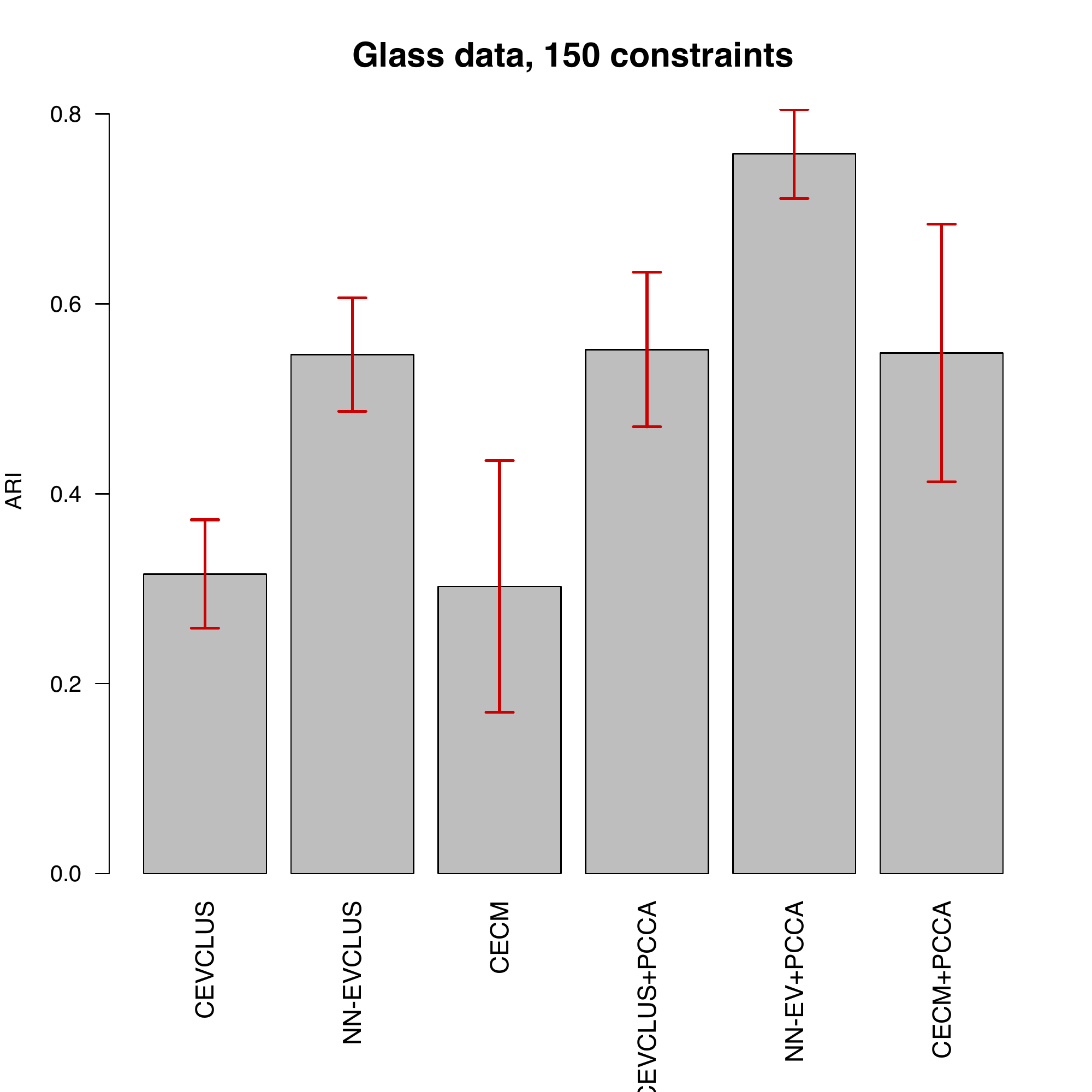}}
\subfloat[\label{fig:Glass_semi_200}]{\includegraphics[width=0.33\textwidth]{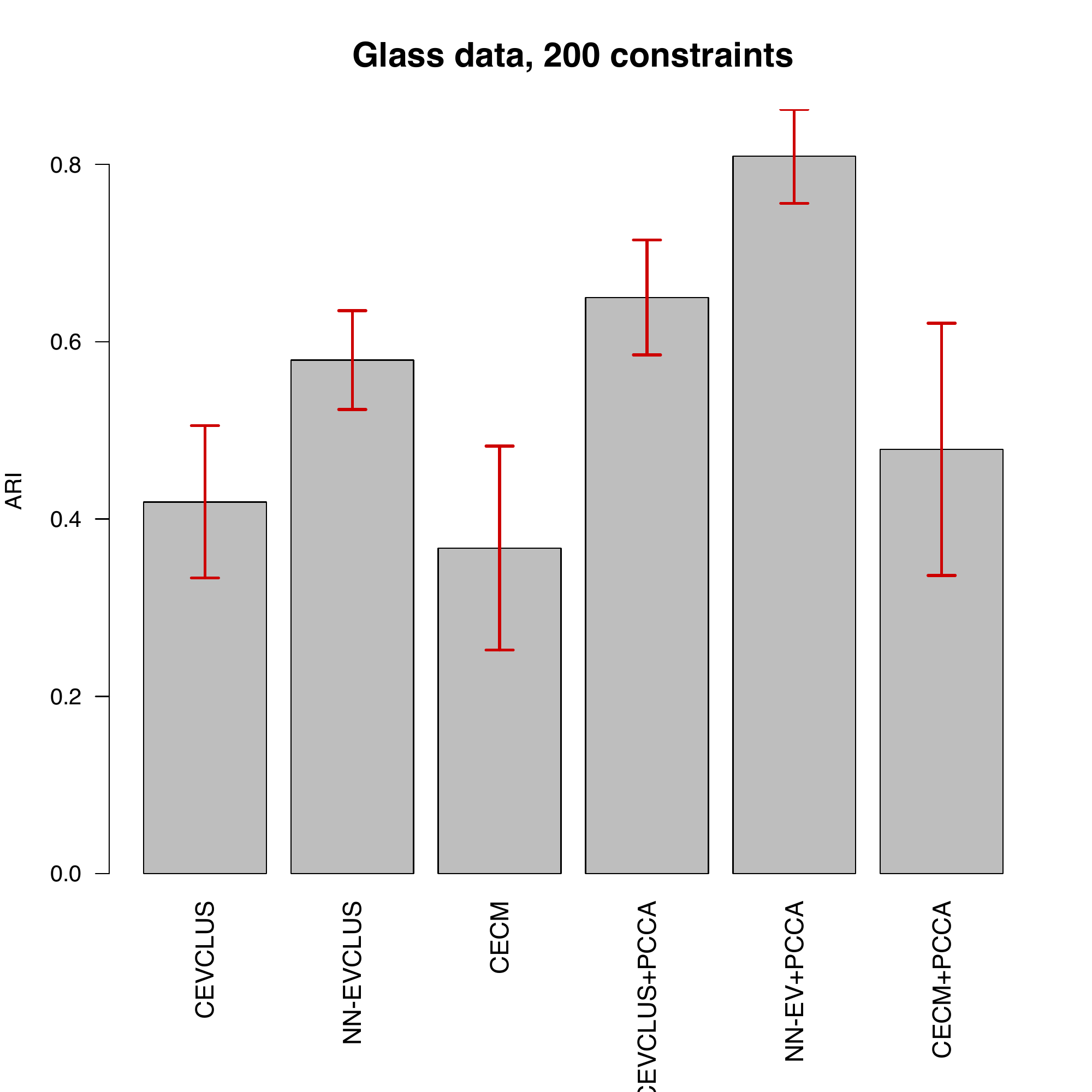}}
\caption{Mean ARI values (over 10 random draws) for the \textsf{Glass} dataset with 50 (a), 100 (b), 150 (c) and 200 (d) constraints. The methods are, from left to right: CEVCLUS, NN-EVCLUS, CECM, and the same methods combined with PCCA.The error bars extend to one standard deviation around the mean.  \label{fig:Glass_semi}}
\end{figure}

\begin{figure}
\centering  
\subfloat[\label{fig:Iris_semi_50}]{\includegraphics[width=0.33\textwidth]{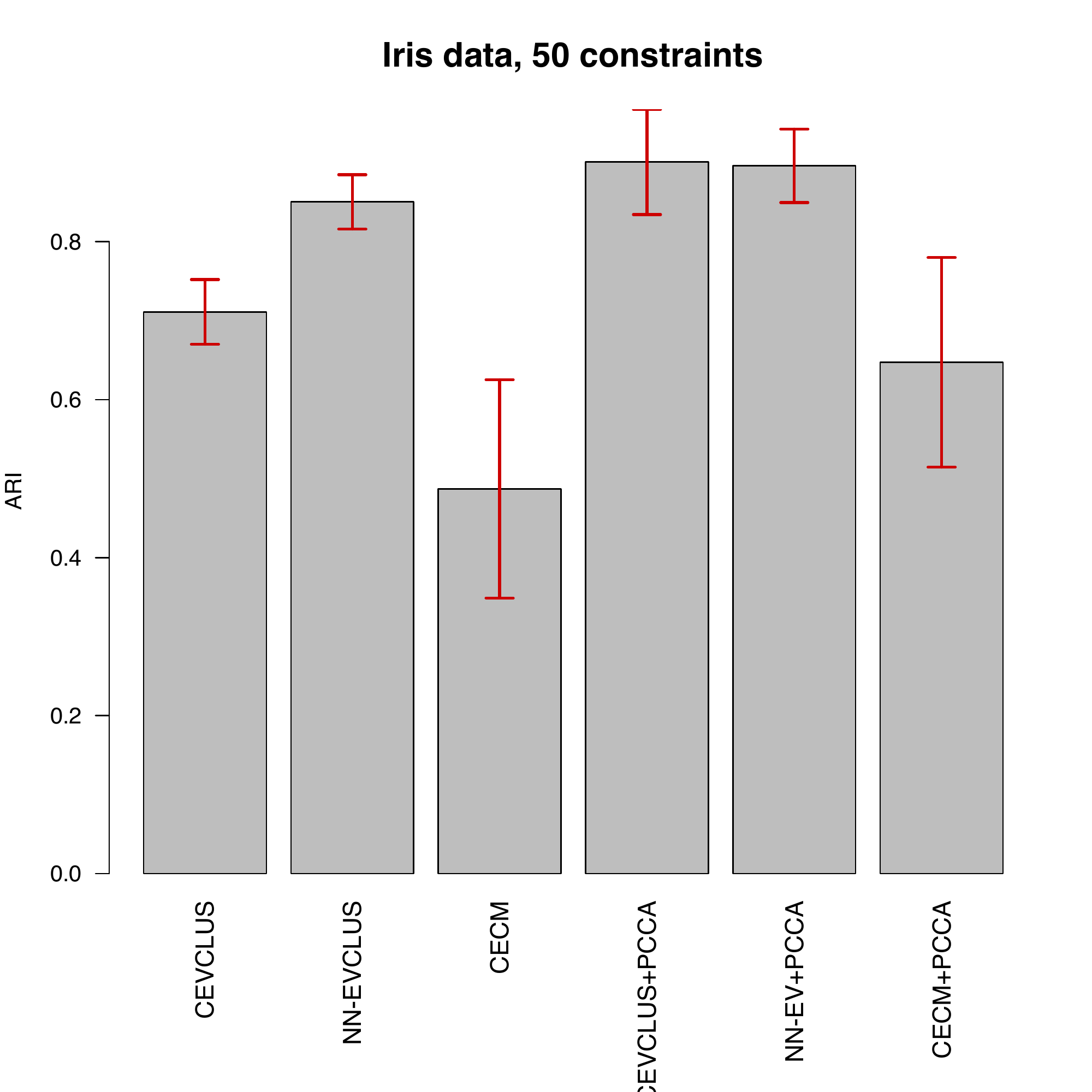}}
\subfloat[\label{fig:Iris_semi_100}]{\includegraphics[width=0.33\textwidth]{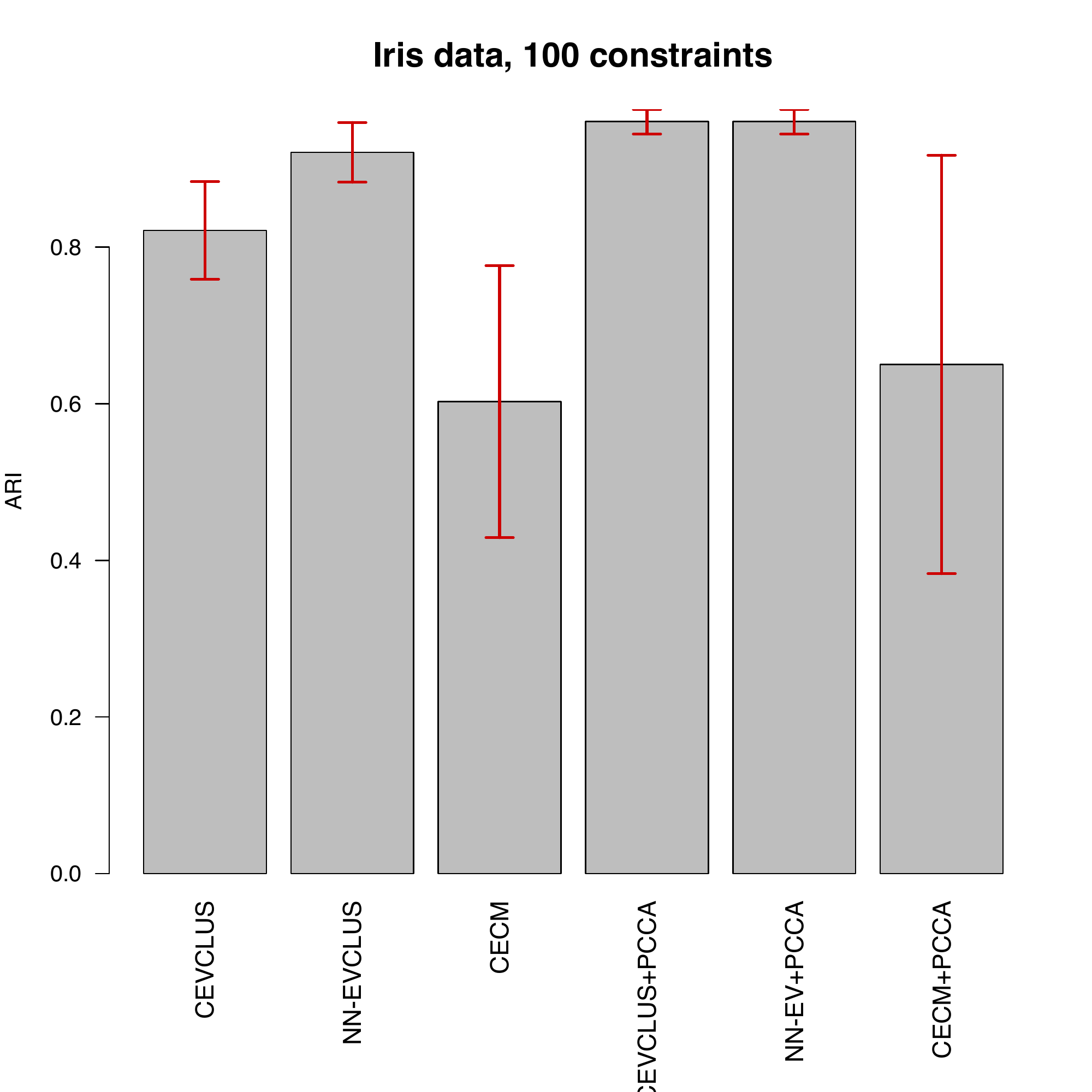}}\\
\subfloat[\label{fig:Iris_semi_150}]{\includegraphics[width=0.33\textwidth]{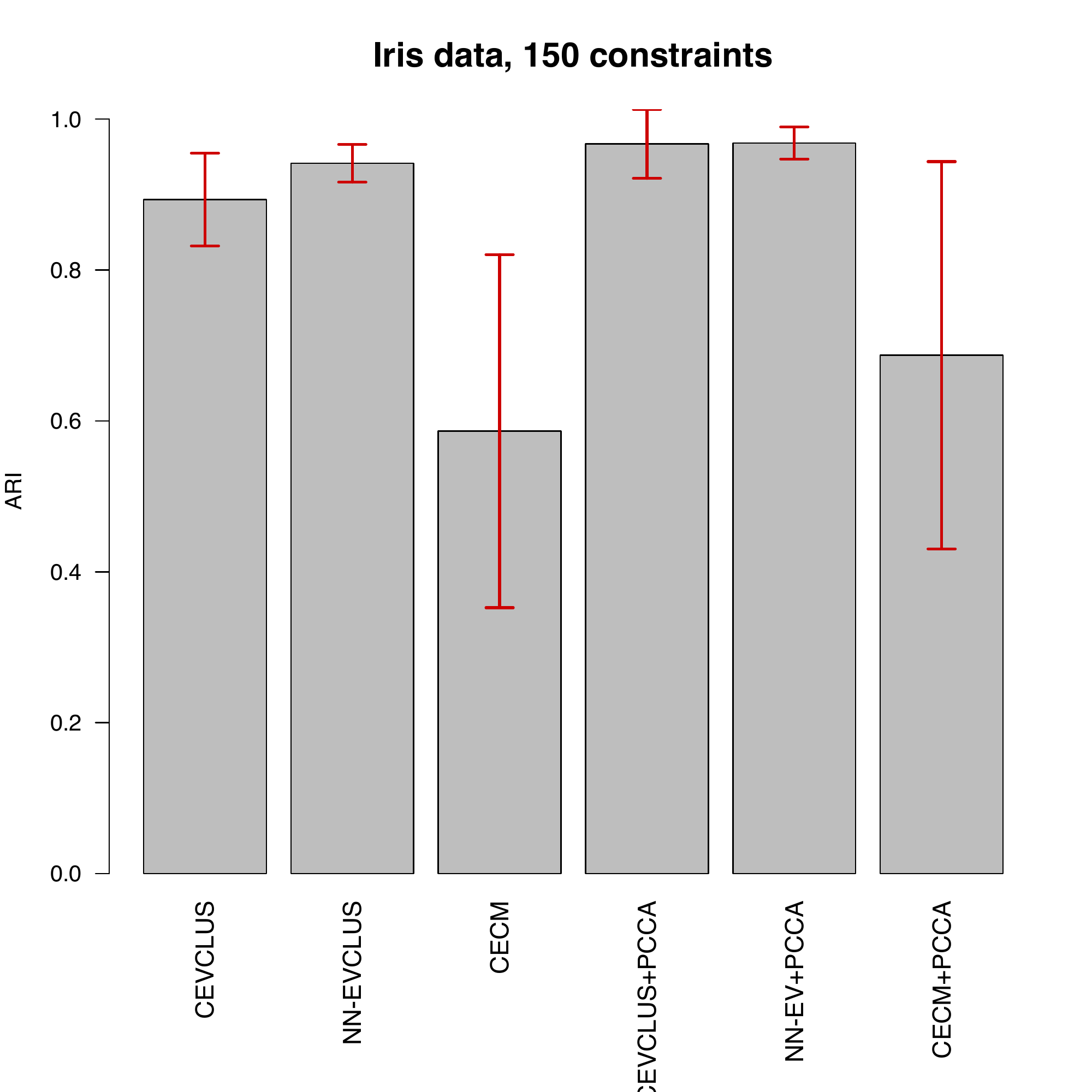}}
\subfloat[\label{fig:Iris_semi_200}]{\includegraphics[width=0.33\textwidth]{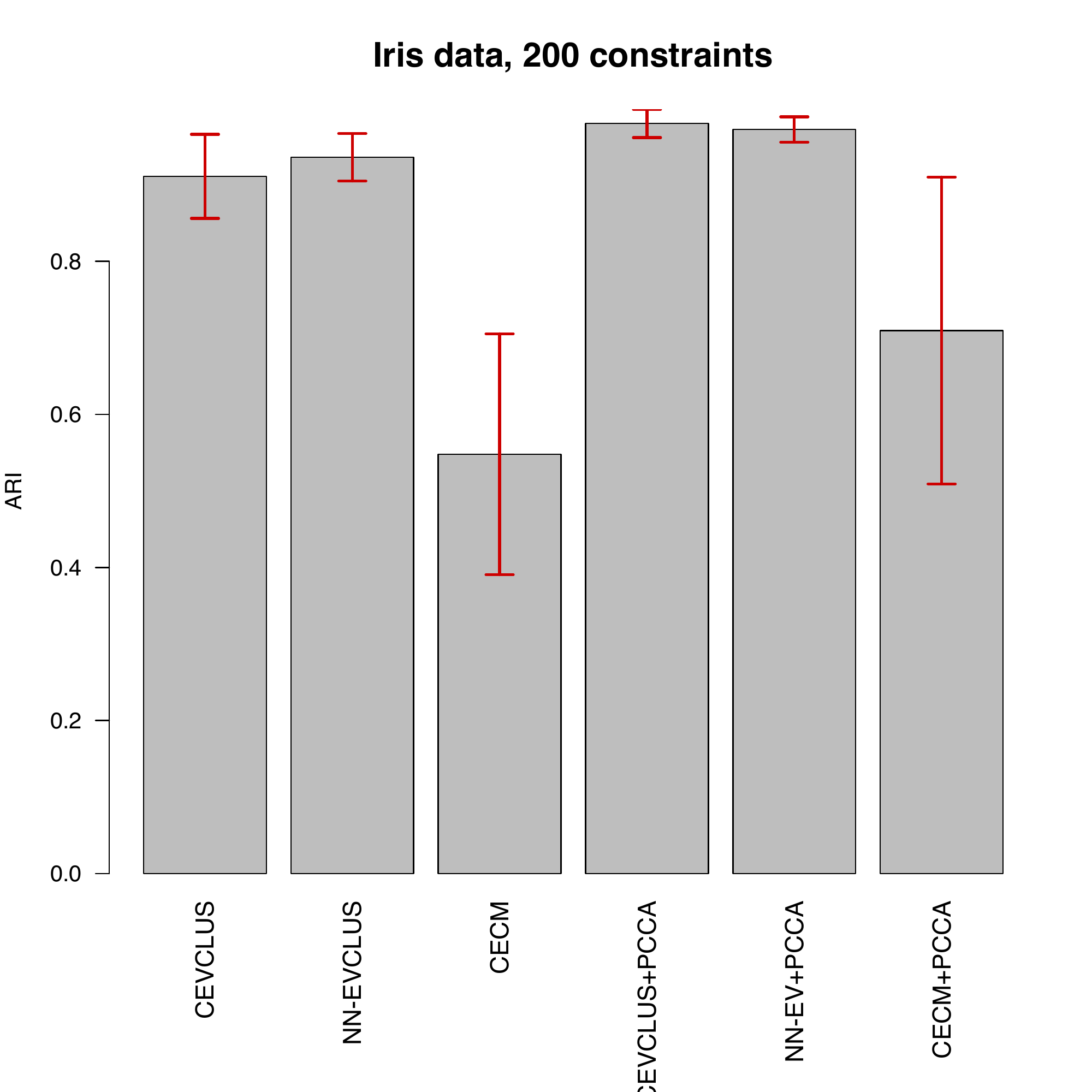}}
\caption{Mean ARI values (over 10 random draws) for the \textsf{Iris} dataset with 50 (a), 100 (b), 150 (c) and 200 (d) constraints. The methods are, from left to right: CEVCLUS, NN-EVCLUS, CECM, and the same methods combined with PCCA. The error bars extend to one standard deviation around the mean. \label{fig:Iris_semi}}
\end{figure}

\begin{figure}
\centering  
\subfloat[\label{fig:Chicken_semi_100}]{\includegraphics[width=0.33\textwidth]{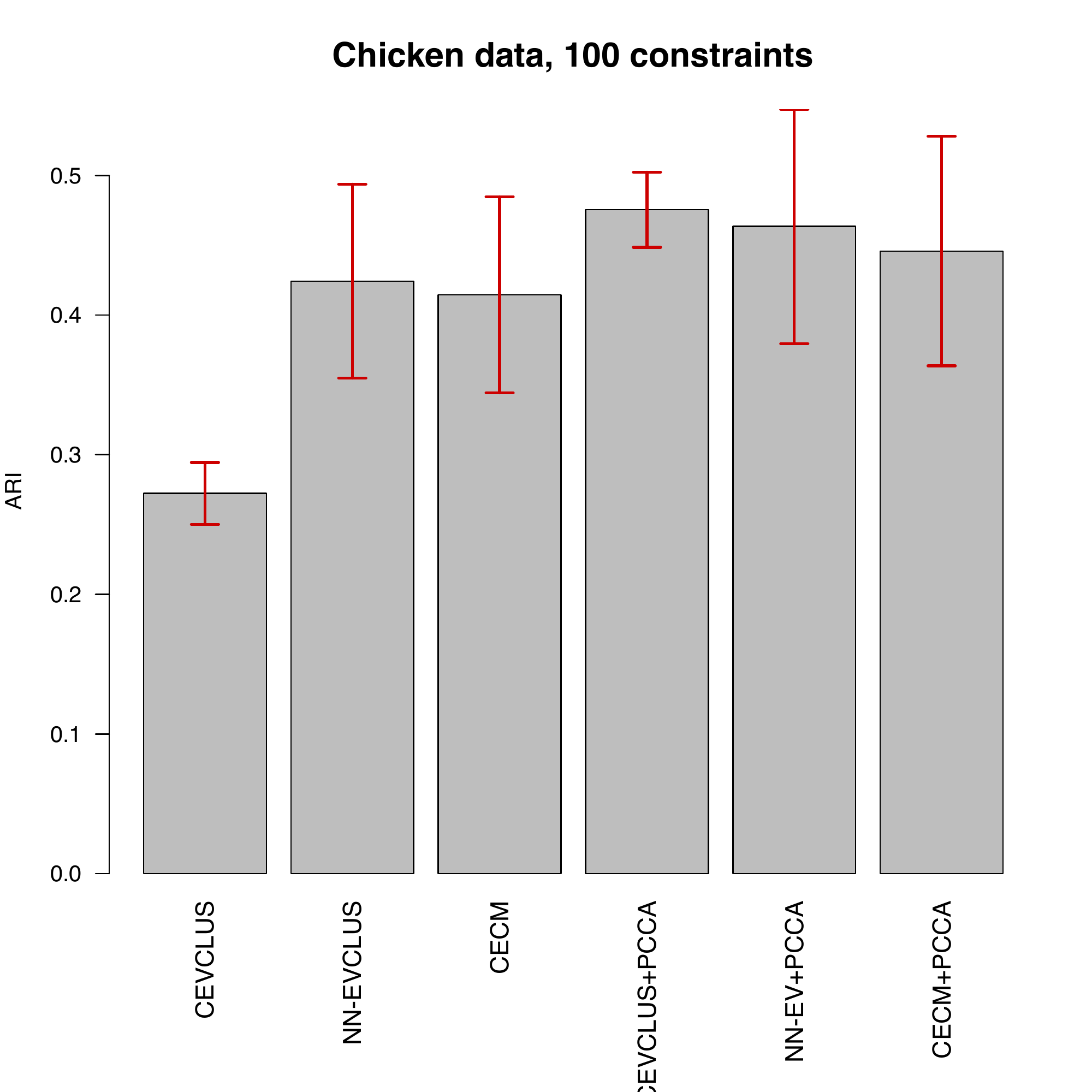}}
\subfloat[\label{fig:Chicken_semi_200}]{\includegraphics[width=0.33\textwidth]{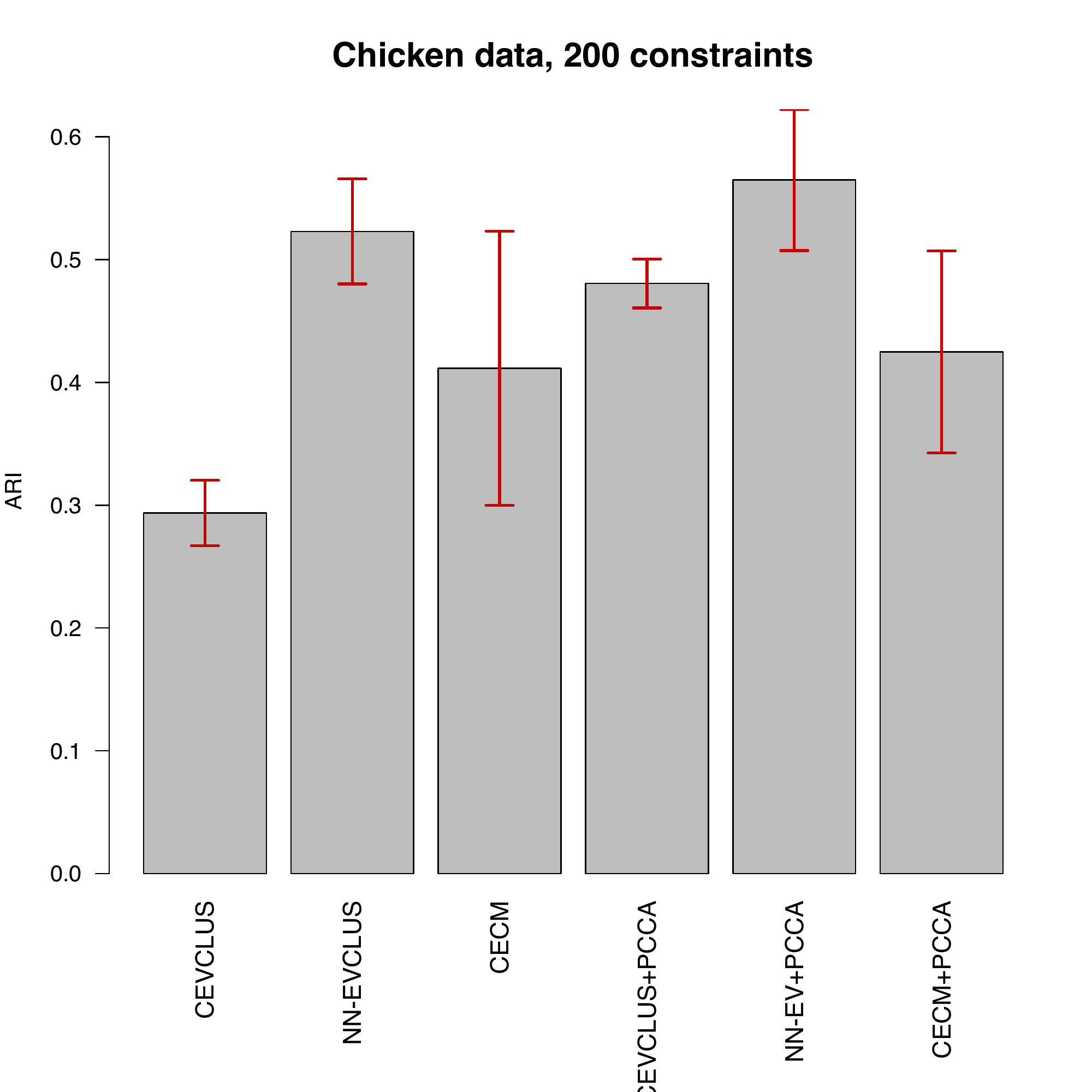}}\\
\subfloat[\label{fig:Chicken_semi_300}]{\includegraphics[width=0.33\textwidth]{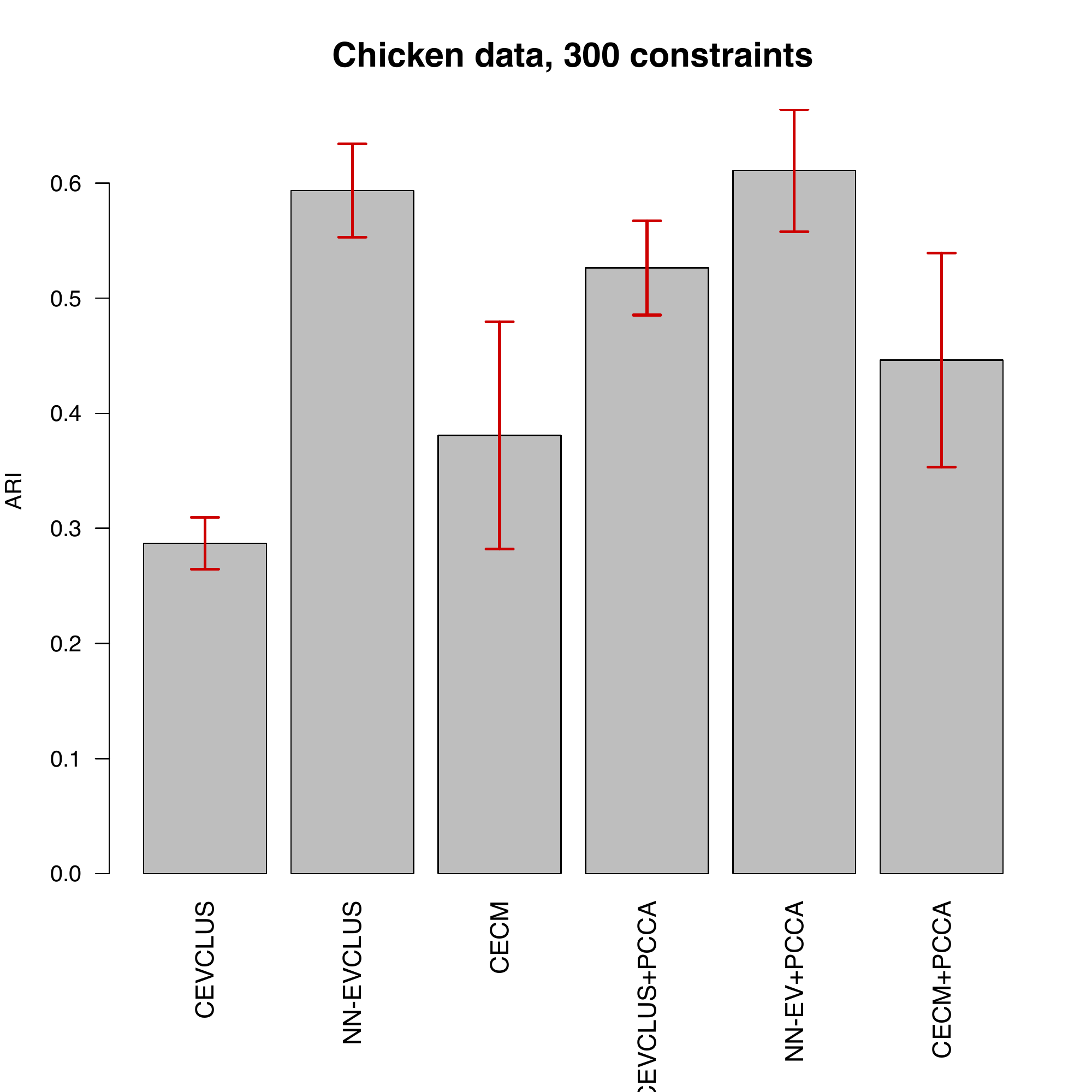}}
\subfloat[\label{fig:Chicken_semi_400}]{\includegraphics[width=0.33\textwidth]{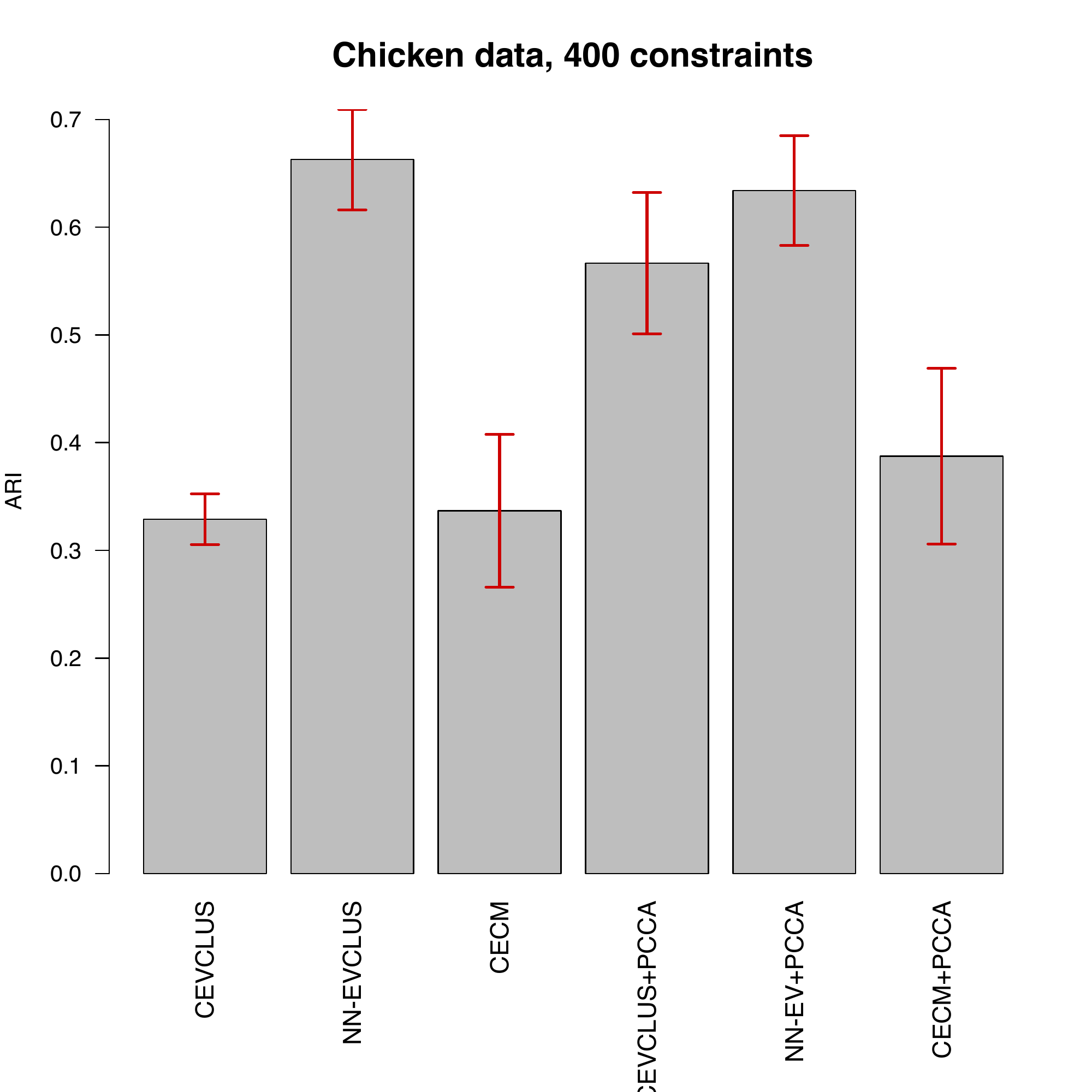}}
\caption{Mean ARI values (over 10 random draws) for the \textsf{ChickenPieces} dataset with 100 (a), 200 (b), 300 (c) and 400 (d) constraints. The methods are, from left to right: CEVCLUS, NN-EVCLUS, CECM, and the same methods combined with PCCA. The error bars extend to one standard deviation around the mean. \label{fig:Chicken_semi}}
\end{figure}

Pairwise constraints were generated randomly from the set of object pairs. For each number of constraints, we drew 10 different sets. The average ARI values for the three datasets are reported with the standard deviations in Figures \ref{fig:Glass_semi}-\ref{fig:Chicken_semi}, \new{and the CPU times with 200 constraints are shown in Table \ref{tab:time_const}.} We can see that, without PCCA, NN-EVCLUS outperformed both CEVCLUS and CECM for the three datasets.  PCCA improved the performances of the three clustering methods. With distances computed in feature space of PCCA, NN-EVCLUS still yielded strictly better results for the \textsf{Glass} and \textsf{ChickenPieces} datasets as shown, respectively, in Figures \ref{fig:Glass_semi} and \ref{fig:Chicken_semi}, and it yielded similar results as CEVCLUS for the \textsf{Iris} dataset (Figure \ref{fig:Iris_semi}). \new{NN-EVCLUS has the highest computing time of the three methods, but it because slightly faster when PCCA is used, as the learning task because simpler.} Overall, the combination of NN-EVCLUS and PCCA consistently provided the best results for the three datasets. 

\begin{table}
\caption{\new{Means and standard deviations (in parentheses) of CPU times (in seconds) over five runs of EVCLUS,  NN-EVCLUS,  CECM, and the same methods combined with PCCA,   applied to the \textsf{Glass},  \textsf{Iris} and \textsf{ChickenPieces} datasets with 200 constraints. }\label{tab:time_const}}
\new{
\begin{center}
\begin{tabular}{lcccccc}
\hline
& EVCLUS & NN-EVCLUS & CECM & EVCLUS & NN-EVCLUS & CECM\\
& & & & +PCCA & +PCCA & +PCCA\\
\hline
\textsf{Glass} & 18.78 (2.1) & 76.22 (12.5)   & 35.27 (24.1) & 31.51 (13.7)& 53.74 (15.9) & 31.07 (12.4)\\
\textsf{Iris} & 1.55 (0.11)& 30.05 (3.17)& 4.78  (2.86)&1.21 (0.07)&23.59 (4.58) &2.08 (1.02)\\
\textsf{Chicken} & 6.80 (0.5) &197.74 (23.6) & 52.85 (44.0) & 3.99 (0.6) &127.18 (21.1) &32.13 (20.5)  \\
\hline
\end{tabular}
\end{center}
}
\end{table}

\clearpage
\section{Conclusions}
\label{sec:concl}

A new neural network-based evidential clustering algorithm, called NN-EVCLUS, has been introduced. This algorithm learns a mapping from  attribute vectors to mass functions on a frame $\Omega$ of $c$ clusters, in such a way that more similar inputs are mapped to mass functions with a lower degree of conflict. It, thus, requires two inputs: a set of attribute vectors and a dissimilarity matrix. In the case of attribute data, dissimilarities are typically  computed as distances in the attribute space. In the case of proximity data, attributes can be computed by performing PCA on the matrix of dissimilarities. When side information is provided in the form of pairwise constraints or labeled data, feature extraction methods such as PCCA or FDA can be used to learn a metric in such a way that objects that are known to belong to different clusters become further apart, while objects in a given cluster are as similar as possible. 

The neural network has a standard multilayer structure but a specific loss function that measures the discrepancy between dissimilarities and degrees of conflict for all or some pairs of objects. Additional error terms  can be added to the loss function to account for pairwise constraints or labeled data. The  network can be trained in batch mode or using minibatch stochastic gradient descent to handle very large datasets. It can be paired with a one-class SVM to make the method robust to outliers and allow for novelty detection. As opposed to EVCLUS, NN-EVCLUS learns a compact representation of the data in the form of connection weights, which makes it able to generalize beyond the learning set and compute an evidential partition for new  data without retraining.

NN-EVCLUS has been compared to alternative evidential clustering algorithms on a range of clustering tasks with both attribute and dissimilarity data. It was shown to outperform other methods for a majority of datasets, by a relatively small margin for EVCLUS and by a larger margin for other algorithms (including ECM, CCM, sECMdd, wECMdd for attribute data, and  RECM, sECMdd, wECMdd for dissimilarity data). \new{NN-EVCLUS was only significantly outperformed by Bootclus on a few attribute datasets with elliptical clusters, for which a Gaussian mixture model is a good fit}. On constrained clustering tasks, NN-EVCLUS was shown to outperform CECM and CEVCLUS.  While metric adaptation using PCCA improved the performances of all methods, the combination of PCCA and EVCLUS yielded the best results overall. 

While we used a standard multilayer perception architecture in this work, more complex architectures such as convolutional neural networks could be used with NN-EVCLUS to cluster  data with a grid-like topology such as time series, images or videos. Also, training time could be drastically reduced by implementing the learning algorithm on GPUs. These ideas are left for further research.

\section*{Acknowledgment}

This research was supported by the Labex MS2T, which was funded through the program ``Investments for the future'' by the National Agency for Research (reference ANR-11-IDEX-0004-02).

\section*{References}

\begin{thebibliography}{10}

\bibitem{antoine12}
V.~Antoine, B.~Quost, M.-H. Masson, and T.~Denoeux.
\newblock {CECM}: Constrained evidential c-means algorithm.
\newblock {\em Computational Statistics \& Data Analysis}, 56(4):894--914,
  2012.

\bibitem{antoine14}
V.~Antoine, B.~Quost, M.-H. Masson, and T.~Denoeux.
\newblock {CEVCLUS}: evidential clustering with instance-level constraints for
  relational data.
\newblock {\em Soft Computing}, 18(7):1321--1335, 2014.

\bibitem{bezdek99}
J.~C. Bezdek, J.~Keller, R.~Krishnapuram, and N.~R. Pal.
\newblock {\em Fuzzy models and algorithms for pattern recognition and image
  processing}.
\newblock Kluwer Academic Publishers, Boston, 1999.

\bibitem{borg97}
I.~Borg and P.~Groenen.
\newblock {\em Modern multidimensional scaling}.
\newblock Springer, New-York, 1997.

\bibitem{bromley93}
J.~Bromley, I.~Guyon, Y.~LeCun, E.~S{\"a}ckinger, and R.~Shah.
\newblock Signature verification using a {``Siamese''} time delay neural
  network.
\newblock In J.~Cowan, G.~Tesauro, and J.~Alspector, editors, {\em Advances in
  Neural Information Processing Systems 6}, pages 737--744. Morgan Kaufmann,
  1994.

\bibitem{chapelle06}
O.~Chapelle, B.~Sch\"olkopf, and A.~Zien, editors.
\newblock {\em Semi-Supervised Learning}.
\newblock MIT Press, Cambridge, Ma, 2006.

\bibitem{denoeux20d}
T.~Den{\oe}ux.
\newblock Calibrated model-based evidential clustering using bootstrapping.
\newblock {\em Information Sciences}, 528:17--45, 2020.

\bibitem{denoeux21}
T.~Den{\oe}ux.
\newblock {\em evclust: Evidential Clustering}, 2021.
\newblock R package version 2.0.0. URL:
  \url{https://CRAN.R-project.org/package=evclust}.

\bibitem{denoeux20b}
T.~Den{\oe}ux, D.~Dubois, and H.~Prade.
\newblock Representations of uncertainty in artificial intelligence: Beyond
  probability and possibility.
\newblock In P.~Marquis, O.~Papini, and H.~Prade, editors, {\em A Guided Tour
  of Artificial Intelligence Research}, volume~1, chapter~4, pages 119--150.
  Springer Verlag, 2020.

\bibitem{denoeux16b}
T.~Denoeux and O.~Kanjanatarakul.
\newblock Beyond fuzzy, possibilistic and rough: An investigation of belief
  functions in clustering.
\newblock In {\em Soft Methods for Data Science (Proc. of the 8th International
  Conference on Soft Methods in Probability and Statistics SMPS 2016)}, volume
  AISC 456 of {\em Advances in Intelligent and Soft Computing}, pages 157--164,
  Rome, Italy, September 2016. Springer-Verlag.

\bibitem{denoeux17a}
T.~Denoeux, S.~Li, and S.~Sriboonchitta.
\newblock Evaluating and comparing soft partitions: an approach based on
  {Dempster-Shafer} theory.
\newblock {\em IEEE Transactions on Fuzzy Systems}, 26(3):1231--1244, 2018.

\bibitem{denoeux04b}
T.~Den{\oe}ux and M.-H. Masson.
\newblock {EVCLUS}: Evidential clustering of proximity data.
\newblock {\em IEEE Trans. on Systems, Man and Cybernetics B}, 34(1):95--109,
  2004.

\bibitem{denoeux16a}
T.~Den{\oe}ux, S.~Sriboonchitta, and O.~Kanjanatarakul.
\newblock Evidential clustering of large dissimilarity data.
\newblock {\em Knowledge-based Systems}, 106:179--195, 2016.

\bibitem{dua19}
D.~Dua and C.~Graff.
\newblock {UCI} machine learning repository, 2017.
\newblock \url{http://archive.ics.uci.edu/ml}.

\bibitem{durso17}
P.~D'Urso.
\newblock Informational paradigm, management of uncertainty and theoretical
  formalisms in the clustering framework: A review.
\newblock {\em Information Sciences}, 400---401:30--62, 2017.

\bibitem{durso19}
P.~D'Urso and R.~Massari.
\newblock Fuzzy clustering of mixed data.
\newblock {\em Information Sciences}, 505:513--534, 2019.

\bibitem{ferone19}
A.~Ferone and A.~Maratea.
\newblock Integrating rough set principles in the graded possibilistic
  clustering.
\newblock {\em Information Sciences}, 477:148--160, 2019.

\bibitem{franti18}
P.~Fr\"anti and S.~Sieranoja.
\newblock K-means properties on six clustering benchmark datasets, 2018.
\newblock http://cs.uef.fi/sipu/datasets/.

\bibitem{goodfellow16}
I.~Goodfellow, Y.~Bengio, and A.~Courville.
\newblock {\em Deep Learning}.
\newblock MIT Press, 2016.
\newblock \url{http://www.deeplearningbook.org}.

\bibitem{graepel99}
T.~Graepel, R.~Herbrich, P.~Bollmann-Sdorra, and K.~Obermayer.
\newblock Classification on pairwise proximity data.
\newblock In {\em Advances in Neural Information Processing Systems 11}, pages
  438--444, Cambridge, MA, 1999. MIT Press.

\bibitem{he21}
Y.~He, Y.~Wu, H.~Qin, J.~Z. Huang, and Y.~Jin.
\newblock Improved i-nice clustering algorithm based on density peaks
  mechanism.
\newblock {\em Information Sciences}, 548:177--190, 2021.

\bibitem{higuera15}
C.~Higuera, K.~J. Gardiner, and K.~J. Cios.
\newblock Self-organizing feature maps identify proteins critical to learning
  in a mouse model of down syndrome.
\newblock {\em PLOS ONE}, 10(6):1--28, 06 2015.

\bibitem{hofmann97}
T.~Hofmann and J.~Buhmann.
\newblock Pairwise data clustering by deterministic annealing.
\newblock {\em IEEE Transactions on Pattern Analysis and Machine Intelligence},
  19(1):1--14, 1997.

\bibitem{hubert85}
L.~Hubert and P.~Arabie.
\newblock Comparing partitions.
\newblock {\em Journal of Classification}, 2(1):193--{\~n}218, 1985.

\bibitem{jain88}
A.~K. Jain and R.~C. Dubes.
\newblock {\em Algorithms for clustering data}.
\newblock Prentice-Hall, Englewood Cliffs, NJ., 1988.

\bibitem{jain97}
A.~K. Jain and D.~Zongker.
\newblock Representation and recognition of handwritten digits using deformable
  templates.
\newblock {\em IEEE Transactions on Pattern Analysis and Machine Intelligence},
  19(12):1386--1391, 1997.

\bibitem{mao95}
{Jianchang Mao} and A.~K. {Jain}.
\newblock Artificial neural networks for feature extraction and multivariate
  data projection.
\newblock {\em IEEE Transactions on Neural Networks}, 6(2):296--317, 1995.

\bibitem{karatzoglou04}
A.~Karatzoglou, A.~Smola, K.~Hornik, and A.~Zeileis.
\newblock kernlab -- an {S4} package for kernel methods in {R}.
\newblock {\em Journal of Statistical Software}, 11(9):1--20, 2004.

\bibitem{kaufman90}
L.~Kaufman and P.~J. Rousseeuw.
\newblock {\em Finding groups in data}.
\newblock Wiley, New-York, 1990.

\bibitem{koklu20}
M.~Koklu and I.~A. Ozkan.
\newblock Multiclass classification of dry beans using computer vision and
  machine learning techniques.
\newblock {\em Computers and Electronics in Agriculture}, 174:105507, 2020.

\bibitem{krishnapuram93}
R.~Krishnapuram and J.~Keller.
\newblock A possibilistic approach to clustering.
\newblock {\em IEEE Trans. on Fuzzy Systems}, 1:98--111, 1993.

\bibitem{li18}
F.~Li, S.~Li, and T.~Den{\oe}ux.
\newblock {k-CEVCLUS}: Constrained evidential clustering of large dissimilarity
  data.
\newblock {\em Knowledge-Based Systems}, 142:29--44, 2018.

\bibitem{lichtenauer08}
J.~Lichtenauer, E.~A. Hendriks, and M.~J.~T. Reinders.
\newblock Sign language recognition by combining statistical {DTW} and
  independent classification.
\newblock {\em IEEE Transactions on Pattern Analysis and Machine Intelligence},
  30:2040--2046, 2008.

\bibitem{liu12}
Z.-G. Liu, J.~Dezert, G.~Mercier, and Q.~Pan.
\newblock Belief c-means: An extension of fuzzy c-means algorithm in belief
  functions framework.
\newblock {\em Pattern Recognition Letters}, 33(3):291--300, 2012.

\bibitem{liu15}
Z.-G. Liu, Q.~Pan, J.~Dezert, and G.~Mercier.
\newblock Credal c-means clustering method based on belief functions.
\newblock {\em Knowledge-Based Systems}, 74(0):119--132, 2015.

\bibitem{masson08}
M.-H. Masson and T.~Denoeux.
\newblock {ECM:} an evidential version of the fuzzy c-means algorithm.
\newblock {\em Pattern Recognition}, 41(4):1384--1397, 2008.

\bibitem{masson09a}
M.-H. Masson and T.~Den{\oe}ux.
\newblock {RECM:} relational evidential c-means algorithm.
\newblock {\em Pattern Recognition Letters}, 30:1015--1026, 2009.

\bibitem{mignon12}
A.~{Mignon} and F.~{Jurie}.
\newblock {PCCA:} a new approach for distance learning from sparse pairwise
  constraints.
\newblock In {\em 2012 IEEE Conference on Computer Vision and Pattern
  Recognition}, pages 2666--2672, 2012.

\bibitem{pal05}
N.~R. {Pal}, K.~{Pal}, J.~M. {Keller}, and J.~C. {Bezdek}.
\newblock A possibilistic fuzzy c-means clustering algorithm.
\newblock {\em IEEE Transactions on Fuzzy Systems}, 13(4):517--530, 2005.

\bibitem{peters06}
G.~Peters.
\newblock Some refinements of rough k-means clustering.
\newblock {\em Pattern Recognition}, 39(8):1481--1491, 2006.

\bibitem{peters14}
G.~Peters.
\newblock Rough clustering utilizing the principle of indifference.
\newblock {\em Information Sciences}, 277:358 -- 374, 2014.

\bibitem{peters13}
G.~Peters, F.~Crespo, P.~Lingras, and R.~Weber.
\newblock Soft clustering: Fuzzy and rough approaches and their extensions and
  derivatives.
\newblock {\em International Journal of Approximate Reasoning}, 54(2):307--322,
  2013.

\bibitem{R20}
{R Core Team}.
\newblock {\em R: A Language and Environment for Statistical Computing}.
\newblock R Foundation for Statistical Computing, Vienna, Austria, 2020.

\bibitem{rodriguez14}
A.~Rodriguez and A.~Laio.
\newblock Clustering by fast search and find of density peaks.
\newblock {\em Science}, 344:1492--1496, 2014.

\bibitem{sholkopf02}
B.~Sch{\"o}lkopf and A.~Smola.
\newblock {\em Learning with kernels}.
\newblock MIT Press, 2002.

\bibitem{shafer76}
G.~Shafer.
\newblock {\em A mathematical theory of evidence}.
\newblock Princeton University Press, Princeton, N.J., 1976.

\bibitem{silva90}
F.~M. Silva and L.~B. Almeida.
\newblock Speeding up backpropagation.
\newblock In R.~Eckmiller, editor, {\em Advances neural computers}, pages
  151--158, New-York, 1990. Elsevier-North-Holland.

\bibitem{smets90a}
P.~Smets.
\newblock The combination of evidence in the {Transferable Belief Model}.
\newblock {\em IEEE Transactions on Pattern Analysis and Machine Intelligence},
  12(5):447--458, 1990.

\bibitem{su19}
Z.~Su and T.~Den{\oe}ux.
\newblock {BPEC}: Belief-peaks evidential clustering.
\newblock {\em IEEE Transactions on Fuzzy Systems}, 27(1):111--123, 2019.

\bibitem{sugiyama06}
M.~Sugiyama.
\newblock Local {Fisher} discriminant analysis for supervised dimensionality
  reduction.
\newblock In {\em Proceedings of 23rd International Conference on Machine
  Learning}, page 905?912. ACM, 2016.

\bibitem{terbraak09}
C.~J. ter Braak, Y.~Kourmpetis, H.~A. Kiers, and M.~C. Bink.
\newblock Approximating a similarity matrix by a latent class model: A
  reappraisal of additive fuzzy clustering.
\newblock {\em Computational Statistics \& Data Analysis}, 53(8):3183--3193,
  2009.

\bibitem{ubukata21}
S.~Ubukata, A.~Notsu, and K.~Honda.
\newblock Objective function-based rough membership c-means clustering.
\newblock {\em Information Sciences}, 548:479--496, 2021.

\bibitem{webb95}
A.~R. Webb.
\newblock Multidimensional scaling by iterative majorization using radial basis
  functions.
\newblock {\em Pattern Recognition}, 28(5):753--759, 1995.

\bibitem{xing03}
E.~P. Xing, M.~I. Jordan, S.~J. Russell, and A.~Y. Ng.
\newblock Distance metric learning with application to clustering with
  side-information.
\newblock In S.~Becker, S.~Thrun, and K.~Obermayer, editors, {\em Advances in
  Neural Information Processing Systems 15}, pages 521--528. MIT Press, 2003.

\bibitem{xu09}
R.~Xu and D.~Wunsch.
\newblock {\em Clustering}.
\newblock Wiley-IEEE Press, Hoboken, New Jersey, 2009.

\bibitem{xu20}
X.~Xu, S.~Ding, Y.~Wang, L.~Wang, and W.~Jia.
\newblock A fast density peaks clustering algorithm with sparse search.
\newblock {\em Information Sciences}, 2020.

\bibitem{yang04}
J.~Yang, Z.~Jin, J.-Y. Yang, D.~Zhang, and A.~F. Frangi.
\newblock Essence of kernel {Fisher} discriminant: {KPCA} plus {LDA}.
\newblock {\em Pattern Recognition}, 37(10):2097--2100, 2004.

\bibitem{yi14}
D.~{Yi}, Z.~{Lei}, S.~{Liao}, and S.~Z. {Li}.
\newblock Deep metric learning for person re-identification.
\newblock In {\em 2014 22nd International Conference on Pattern Recognition},
  pages 34--39, 2014.

\bibitem{ying12}
Y.~Ying and P.~Li.
\newblock Distance metric learning with eigenvalue optimization.
\newblock {\em Journal of Machine Learning Research}, 13(1):1--26, 2012.

\bibitem{zadeh78}
L.~A. Zadeh.
\newblock Fuzzy sets as a basis for a theory of possibility.
\newblock {\em Fuzzy Sets and Systems}, 1:3--28, 1978.

\bibitem{zhou16}
K.~Zhou, A.~Martin, Q.~Pan, and Z.~ga~Liu.
\newblock {ECMdd}: Evidential c-medoids clustering with multiple prototypes.
\newblock {\em Pattern Recognition}, 60:239--257, 2016.

\bibitem{zhou15}
K.~Zhou, A.~Martin, Q.~Pan, and Z.-G. Liu.
\newblock Median evidential c-means algorithm and its application to community
  detection.
\newblock {\em Knowledge-Based Systems}, 74(0):69--88, 2015.

\end{thebibliography}

\appendix

\section{Gradient of $\calL_{ij}$ w.r.t. $\btheta$}
\label{sec:error_grad}

We consider a neural network with one hidden layer as described in Section \ref{subsec:model}. The vector of parameters is  $\btheta=(V,W,\beta_0,\beta_1)$. Let us first compute the derivatives of $\calL_{ij}$ w.r.t the last-layer weights $W=(w_{qh})$. From  \eqref{eq:lossEVCLUS3} and \eqref{eq:propag2}, we have

\begin{equation}
\label{eq:dLijdwqh}
\deriv{\calL_{ij}}{w_{qh}}=\deriv{\calL_{ij}}{\mu_{iq}} \deriv{\mu_{iq}}{w_{qh}} + \deriv{\calL_{ij}}{\mu_{jq}}\deriv{\mu_{jq}}{w_{qh}}=\Delta_{ijq} z_{ih}+\Delta'_{ijq} z_{jh},
\end{equation}
with
\begin{equation}
\Delta_{ijq}=\deriv{\calL_{ij}}{\mu_{iq}}=\deriv{\calL_{ij}}{\kappa_{ij}}\deriv{\kappa_{ij}}{\mu_{iq}}=2(\kappa_{ij}-\delta^*_{ij}) \sum_{r=1}^f \deriv{\kappa_{ij}}{m^*_{ir}} \deriv{m^*_{ir}}{m_{ir}} \deriv{m_{ir}}{\mu_{iq}}
\end{equation}
and
\begin{equation}
\Delta'_{ijq}=\deriv{\calL_{ij}}{\mu_{jq}}=\deriv{\calL_{ij}}{\kappa_{ij}}\deriv{\kappa_{ij}}{\mu_{jq}}=2(\kappa_{ij}-\delta^*_{ij}) \sum_{r=1}^f \deriv{\kappa_{ij}}{m^*_{jr}} \deriv{m^*_{jr}}{m_{jr}}  \deriv{m_{jr}}{\mu_{jq}}.
\end{equation}
From \eqref{eq:conflict1}, we get
\begin{equation}
\label{eq:dkappadm}
\deriv{\kappa_{ij}}{m^*_{ir}}=(\bC \bm^*_j)_r.
\end{equation}
From \eqref{eq:mstar}, 
\[
\deriv{m^*_{ir}}{m_{ir}}=\gamma_i,
\]
and from \eqref{eq:propag2},
\begin{equation}
\label{eq:dmdmu1}
\deriv{m_{ir}}{\mu_{iq}}=\begin{cases}
m_{ir}(1-m_{ir}) & r=q\\
-m_{ir}m_{iq} & r\neq q.
\end{cases}
\end{equation}
Similarly,
\begin{equation}
\label{eq:dkappadmjr}
\deriv{\kappa_{ij}}{m^*_{jr}}=(\bC \bm^*_i)_r,
\quad
\deriv{m^*_{jr}}{m_{jr}}=\gamma_j,
\end{equation}
and
\begin{equation}
\label{eq:dmdmu2}
\deriv{m_{jr}}{\mu_{iq}}=\begin{cases}
m_{jr}(1-m_{jr}) & r=q\\
-m_{jr}m_{jq} & r\neq q.
\end{cases}
\end{equation}

Now, from \eqref{eq:propag1} and \eqref{eq:propag2}, the derivatives  w.r.t the first-layer weights $V=(w_{hk})$ are
\begin{equation}
\label{eq:dLijdvhk}
\deriv{\calL_{ij}}{v_{hk}}=\deriv{\calL_{ij}}{a_{ih}} \deriv{a_{ih}}{v_{hk}} + \deriv{\calL_{ij}}{a_{jh}}\deriv{a_{jh}}{v_{hk}}=\Delta_{ijh}x_{ki}+\Delta'_{ijh}x_{kj}, 
\end{equation}
with
\begin{equation}
\label{eq:Deltaijh}
\Delta_{ijh}=\deriv{\calL_{ij}}{a_{ih}}=\sum_{q=1}^f \deriv{\calL_{ij}}{\mu_{iq}}\deriv{\mu_{iq}}{z_{ih}}\deriv{z_{ih}}{a_{ih}}=I(z_{ih}>0)\sum_{q=1}^f \Delta_{ijq} w_{qh} 
\end{equation}

and
\begin{equation}
\label{eq:Deltapijh}
\Delta'_{ijh}=\deriv{\calL_{ij}}{a_{jh}}=\sum_{q=1}^f \deriv{\calL_{ij}}{\mu_{jq}}\deriv{\mu_{jq}}{z_{jh}}\deriv{z_{jh}}{a_{jh}}=I(z_{ih}>0) \sum_{q=1}^f \Delta'_{ijq} w_{qh}.
\end{equation}

Finally,  we now compute the derivatives of $\calL_{ij}$ w.r.t. $\beta_0$ and $\beta_1$. We have
\begin{equation}
\label{eq:dLdbeta}
\deriv{\calL_{ij}}{\beta_k}=\deriv{\calL_{ij}}{\gamma_{i}} \deriv{\gamma_{i}}{\beta_k} + \deriv{\calL_{ij}}{\gamma_j} \deriv{\gamma_j}{\beta_k}
\end{equation}
for $k\in\{0,1\}$. From \eqref{eq:gamma}, the first term of the sum in the right-hand side of \eqref{eq:dLdbeta} can be computed as
\[
\deriv{\calL_{ij}}{\gamma_{i}}=\deriv{\calL_{ij}}{\kappa_{ij}}\deriv{\kappa_{ij}}{\gamma_i}=2(\kappa_{ij}-\delta^*_{ij})\sum_{r=1}^f \deriv{\kappa_{ij}}{m^*_{ir}}\deriv{m^*_{ir}}{\gamma_i},
\]
with $ \deriv{\kappa_{ij}}{m^*_{ir}}$ given by \eqref{eq:dkappadm},
\begin{equation}
\label{eq:dmdgamma}
\deriv{m^*_{ir}}{\gamma_i}=\begin{cases}
1- m_{ir} & \text{if } r=1\\
-m_{ir} & \text{otherwise, } 
\end{cases}
\end{equation}
and
\begin{subequations}
\label{eq:dgammadbeta}
\begin{align}
\deriv{\gamma_i}{\beta_0}&=\deriv{\gamma_i}{\eta_i}\deriv{\eta_i}{\beta_0}=\frac{1}{(1+\eta_i)^2} \cdot \frac{\exp(\beta_0+\beta_1 f(\bx_i))}{1+\exp(\beta_0+\beta_1 f(\bx_i))}\\
\deriv{\gamma_i}{\beta_1}&= \deriv{\gamma_i}{\beta_0} f(\bx_i).
\end{align}
\end{subequations}
The first term of the sum in the right-hand side of \eqref{eq:dLdbeta} can be computed in the same way, by replacing $i$ with $j$. 

\section{Gradient of $\calP_\mathsf{ML}$ and $\calP_\mathsf{CL}$ w.r.t. $\btheta$}
\label{sec:pairwise_grad}

We have
\[
\calP_\mathsf{ML}=\sum_{(i,j)\in\mathsf{ML}} \calP_{ij}
\quad \text{and} \quad 
\calP_\mathsf{CL}=\sum_{(i,j)\in\mathsf{CL}} \left(2-\calP_{ij}\right),
\]
with
\[
\calP_{ij}=\bm_i^{*T} \bQ \bm_j^*.
\]
We thus only need to compute the gradient of $\calP_{ij}$. The derivatives w.r.t. the hidden-to-output weights are
\[
\deriv{\calP_{ij}}{w_{qh}}=\deriv{\calP_{ij}}{\mu_{iq}} \deriv{\mu_{iq}}{w_{qh}} + \deriv{\calP_{ij}}{\mu_{jq}}\deriv{\mu_{jq}}{w_{qh}}=\nabla_{ijq}z_{ih} +\nabla'_{ijq}z_{jh}
\]
with
\[
\nabla_{ijq}= \deriv{\calP_{ij}}{\mu_{iq}}=\sum_{r=1}^f \deriv{\calP_{ij}}{m^*_{ir}} \underbrace{\deriv{m^*_{ir}}{m_{ir}}}_{1-\gamma_i}\underbrace{\deriv{m_{ir}}{\mu_{iq}}}_{\text{see }\eqref{eq:dmdmu1}}
\]
and
\[
\deriv{\calP_{ij}}{m_{ir}}=(\bQ \bm^*_j)_r.
\]
Similarly,
\[
\nabla'_{ijq}= \deriv{\calP_{ij}}{\mu_{jq}}=\sum_{r=1}^f \deriv{\calP_{ij}}{m^*_{jr}} \underbrace{\deriv{m^*_{jr}}{m_{jr}}}_{1-\gamma_j} \underbrace{\deriv{m_{jr}}{\mu_{jq}}}_{\text{see }\eqref{eq:dmdmu2}}
\]
and
\[
\deriv{\calP_{ij}}{m_{jr}}=(\bQ \bm^*_i)_r.
\]
The derivatives w.r.t. the input-to-hidden weights are,
\[
\deriv{\calP_{ij}}{v_{hk}}=\deriv{\calP_{ij}}{a_{ih}}\deriv{a_{ih}}{v_{hk}} + \deriv{\calP_{ij}}{a_{jh}}\deriv{a_{jh}}{v_{hk}}=\nabla_{ijh}x_{ki}+\nabla'_{ijh}x_{kj}
\]
with
\[
\nabla_{ijh}=\deriv{\calP_{ij}}{a_{ih}}=\sum_{q=1}^f \deriv{\calP_{ij}}{\mu_{iq}}\deriv{\mu_{iq}}{z_{ih}}\deriv{z_{ih}}{a_{ih}}=I(z_{ih}>0)\sum_{q=1}^f \nabla_{ijq} w_{qh} 
\]
and
\[
\nabla'_{ijh}=\deriv{\calP_{ij}}{a_{jh}}=\sum_{q=1}^f \deriv{\calP_{ij}}{\mu_{jq}}\deriv{\mu_{jq}}{z_{jh}}\deriv{z_{jh}}{a_{jh}}=I(z_{ih}>0) \sum_{q=1}^f \nabla'_{ijq} w_{qh}.
\]

Finally, we have
\[
\deriv{\calP_{ij}}{\beta_k}=\deriv{\calP_{ij}}{\gamma_{i}} \deriv{\gamma_{i}}{\beta_k} + \deriv{\calP_{ij}}{\gamma_j} \deriv{\gamma_j}{\beta_k}
\]
for $k\in\{0,1\}$, where the derivatives of $\gamma_i$ and $\gamma_j$ are given by \eqref{eq:dgammadbeta}, and
\[
\deriv{\calP_{ij}}{\gamma_{i}}=\sum_{r=1}^f \deriv{\calP_{ij}}{m^*_{ir}}\deriv{m^*_{ir}}{\gamma_i},
\]
\[
\deriv{\calP_{ij}}{\gamma_{j}}=\sum_{r=1}^f \deriv{\calP_{ij}}{m^*_{jr}}\deriv{m^*_{jr}}{\gamma_j},
\]
with $\deriv{m^*_{ir}}{\gamma_i}$ and $\deriv{m^*_{jr}}{\gamma_j}$ given by \eqref{eq:dmdgamma}.

\section{Gradient of $\calP_s$  w.r.t. $\btheta$}
\label{sec:ns_grad}

We have
\[
\calP_s=\frac1{n_s} \sum_{i\in\calI_s} \calP_i
\]
with
\[
\calP_i=\sum_{l=1}^c (pl^*_{il}-y_{il})^2.
\]
We thus only need to compute the gradient of $\calP_i$. The derivatives w.r.t. to the hidden-to-output weights are
\[
\deriv{\calP_i}{w_{qh}}=\deriv{\calP_{i}}{\mu_{iq}}\deriv{\mu_{iq}}{w_{qh}} =\Delta_{iq}z_{ih},
\]
with
\[
\Delta_{iq}=\deriv{\calP_{i}}{\mu_{iq}}= \sum_{l=1}^c \deriv{\calP_{i}}{pl^*_{il}} \deriv{pl^*_{il}}{\mu_{iq}}=2(pl^*_{il}-y_{il})\sum_{l=1}^c\deriv{pl^*_{il}}{\mu_{iq}}
\]
and 
\[
\deriv{pl^*_{il}}{\mu_{iq}}=\sum_{r=1}^f \underbrace{\deriv{pl^*_{il}}{m^*_{ir}}}_{I(\omega_l \in A_r)} \underbrace{\deriv{m^*_{ir}}{m_{ir}}}_{1-\gamma_i}  \underbrace{\deriv{m_{ir}}{\mu_{iq}}}_{\text{See \eqref{eq:dmdmu1}}}.
\]

The derivatives w.r.t the input-to-hidden weights are given by
\[
\deriv{\calP_{i}}{v_{hk}}=\underbrace{\deriv{\calP_{i}}{a_{ih}}}_{\Delta_{ih}}\underbrace{\deriv{a_{ih}}{v_{hk}}}_{x_{ki}} ,
\]
with
\[
\Delta_{ih}=\sum_{q=1}^f \deriv{\calP_{i}}{\mu_{iq}}\deriv{\mu_{iq}}{z_{ih}}\deriv{z_{ih}}{a_{ih}}=I(z_{ih}>0)\sum_{q=1}^f \Delta_{iq} w_{qh}.
\]

Finally, the derivatives w.r.t $\beta_0$ and $\beta_1$ can be computed as
\[
\deriv{\calP_{i}}{\beta_k}=\deriv{\calP_{i}}{\gamma_{i}} \underbrace{\deriv{\gamma_{i}}{\beta_k}}_{\text{See \eqref{eq:dgammadbeta}}},
\]
with
\[
\deriv{\calP_{i}}{\gamma_{i}}=\sum_{l=1}^c \underbrace{\deriv{\calP_{i}}{pl^*_{il}}}_{2(pl^*_{il}-y_{il})} \deriv{pl^*_{il}}{\gamma_i}
\]
and
\[
\deriv{pl^*_{il}}{\gamma_i}=\sum_{r=1}^f \underbrace{\deriv{pl^*_{il}}{m^*_{ir}}}_{I(\omega_l\in A_r)} \underbrace{\deriv{m^*_{ir}}{\gamma_i}}_{\text{See \eqref{eq:dmdgamma}}}.
\]

\end{document}